\documentclass{article}
\usepackage[utf8]{inputenc}
\usepackage{url}
\usepackage{amsmath,amssymb}
\usepackage[T1]{fontenc}
\usepackage{textcomp} 
\usepackage{lmodern}
\IfFileExists{upquote.sty}{\usepackage{upquote}}{}
\IfFileExists{microtype.sty}{
  \usepackage[]{microtype}
  \UseMicrotypeSet[protrusion]{basicmath} 
}{}
\makeatletter
\@ifundefined{KOMAClassName}{
  \IfFileExists{parskip.sty}{%
    \usepackage{parskip}
  }{
    \setlength{\parindent}{0pt}
    \setlength{\parskip}{6pt plus 2pt minus 1pt}}
}{
  \KOMAoptions{parskip=half}}
\makeatother
\usepackage{xcolor}
\usepackage{soul}
\usepackage{bookmark}

\usepackage{graphicx} 
\usepackage{float}
\newfloat{algorithm}{htbp}{loa}
\floatname{algorithm}{Algorithm}

\IfFileExists{xurl.sty}{\usepackage{xurl}}{} 
\hypersetup{
  pdftitle={Autonomous Artificial Intelligence Agents for 
Clinical Decision Making in Oncology},
  hidelinks,
  pdfcreator={LaTeX via pandoc}}

\usepackage{longtable}
\usepackage{booktabs}
\usepackage{geometry}
\usepackage{amsfonts}
\usepackage{pifont}
\usepackage{array} 
\usepackage{xcolor}
\usepackage{xstring} 
\usepackage{graphicx} 
\usepackage{amsmath}
\usepackage{amssymb}
\usepackage{algorithm}
\usepackage{algpseudocode}
\usepackage{wrapfig} 
\usepackage{caption}
\usepackage{marvosym}
\usepackage{tikz}

\definecolor{myred}{RGB}{255, 0, 0} 
\definecolor{mygreen}{RGB}{0, 255, 0} 

\newcommand{\cmark}{\textcolor{green!80!black}{\ding{51}}}
\newcommand{\xmark}{\textcolor{red}{\ding{55}}}
\newcommand{\qmark}{\textcolor{blue}{\ding{72}}}
\newcommand{\lmark}{\textcolor{orange}{\scalebox{1.3}{\Lightning}}}

\algnewcommand\algorithmicforeach{\textbf{for each}}
\algdef{S}[FOR]{ForEach}[1]{\algorithmicforeach\ #1\ \algorithmicdo}

\newcommand{\customCaption}[1]{%
    \begin{center}
        \Large #1\\[2ex] 
    \end{center}
}

\definecolor{lightblue}{RGB}{205, 175, 255} 
\definecolor{orange}{RGB}{255, 165, 0} 
\definecolor{myred}{RGB}{255, 0, 0}

\title{\phantomsection\label{_fm0tygoa1omj}{}Autonomous Artificial Intelligence Agents \\ for Clinical Decision Making in Oncology
}

\author{}
\date{}

\begin{document}
\maketitle

\begin{center}
Dyke Ferber (1, 2), Omar S. M. El Nahhas (2), Georg Wölflein (3),
Isabella C. Wiest (2, 4), Jan Clusmann (2, 5), Marie-Elisabeth Leßman
(2, 6), Sebastian Foersch (7), Jacqueline Lammert (8, 9, 10, 11),
Maximilian Tschochohei (12), Dirk Jäger (1), Manuel Salto-Tellez (13),
Nikolaus Schultz (14), Daniel Truhn (15), Jakob Nikolas Kather (1, 2, 6,
+)
\end{center}

\begin{enumerate}
\def\labelenumi{\arabic{enumi}.}
\item
  Department of Medical Oncology, National Center for Tumor Diseases
  (NCT), Heidelberg University Hospital, Heidelberg, Germany
\item
  Else Kroener Fresenius Center for Digital Health, Technical University
  Dresden, Dresden, Germany
\item
  School of Computer Science, University of St Andrews, St Andrews,
  United Kingdom
\item
  Department of Medicine II, Medical Faculty Mannheim, Heidelberg
  University, Mannheim, Germany
\item
  Department of Medicine III, University Hospital RWTH Aachen, Aachen,
  Germany
\item
  Department of Medicine I, University Hospital Dresden, Dresden,
  Germany
\item
  Institute of Pathology, University Medical Center Mainz, Mainz,
  Germany
\item
  Department of Gynecology and Center for Hereditary Breast and Ovarian
  Cancer, University Hospital rechts der Isar, Technical University of
  Munich (TUM), Munich, Germany
\item
  Center for Personalized Medicine (ZPM), University Hospital rechts
  der Isar, Technical University of Munich (TUM), Munich, Germany
\item
  German Cancer Consortium (DKTK), Munich partner site, Germany
\item
  EUropean Network for RAre CANcers (EURACAN) Initiative, Munich partner
  site, Germany
\item
  Google Cloud, Munich, Germany
\item
  Integrated Pathology Unit, Institute for Cancer Research and Royal
  Marsden Hospital, London, UK
\item
  Department of Epidemiology and Biostatistics, Memorial Sloan Kettering
  Cancer Center, New York, NY, 10065, USA
\item
  Department of Diagnostic and Interventional Radiology, University
  Hospital Aachen, Germany
\end{enumerate}

+ Corresponding author: jakob-nikolas.kather@alumni.dkfz.de

Jakob Nikolas Kather, MD, MSc

Professor of Clinical Artificial Intelligence

Else Kröner Fresenius Center for Digital Health

Technische Universität Dresden

DE -- 01062 Dresden

Phone: +49 351 458-7558

Fax: +49 351 458 7236

Mail: jakob\_nikolas.kather@tu-dresden.de

ORCID ID: 0000-0002-3730-5348

\section{Abstract}\label{abstract}

Multimodal artificial intelligence (AI) systems have the potential to
enhance clinical decision-making by interpreting various types of
medical data. However, the effectiveness of these models across all
medical fields is uncertain. Each discipline presents unique challenges
that need to be addressed for optimal performance. This complexity is
further increased when attempting to integrate different fields into a
single model.

Here, we introduce an alternative approach to multimodal medical AI that
utilizes the generalist capabilities of a large language model (LLM) as
a central reasoning engine. This engine autonomously coordinates and
deploys a set of specialized medical AI tools. These tools include text,
radiology and histopathology image interpretation, genomic data
processing, web searches, and document retrieval from medical
guidelines.

We validate our system across a series of clinical oncology scenarios
that closely resemble typical patient care workflows. We show that the
system has a high capability in employing appropriate tools (97\%),
drawing correct conclusions (93.6\%), and providing complete (94\%), and
helpful (89.2\%) recommendations for individual patient cases while
consistently referencing relevant literature (82.5\%) upon instruction.

This work provides evidence that LLMs can effectively plan and execute
domain-specific models to retrieve or synthesize new information when
used as autonomous agents. This enables them to function as specialist,
patient-tailored clinical assistants. It also simplifies regulatory
compliance by allowing each component tool to be individually validated
and approved. We believe, that our work can serve as a proof-of-concept
for more advanced LLM-agents in the medical domain.

\clearpage

\section{Main text}\label{main-text}

The future of medical AI is
multimodal\href{https://paperpile.com/c/q3Wf0T/QfTSp}{\textsuperscript{1}}.
Several such AI systems that process a wide scope of data inputs have
been introduced
recently\href{https://paperpile.com/c/q3Wf0T/TvfrY}{\textsuperscript{2}}.
Notable examples include models that analyze radiology images together
with clinical
data\href{https://paperpile.com/c/q3Wf0T/7tuPM}{\textsuperscript{3}}, or
integrate information from histopathology with
genomic\href{https://paperpile.com/c/q3Wf0T/OpwMD}{\textsuperscript{4}}
or text-based
information\href{https://paperpile.com/c/q3Wf0T/VXY2u}{\textsuperscript{5}}.
These advancements have fueled anticipations for the advent of
generalist multimodal AI
systems\href{https://paperpile.com/c/q3Wf0T/nGI1G+52gVw}{\textsuperscript{6,7}},
characterized by their ability to concurrently analyze and reason across
any dimension in medical information.

However, it remains to be investigated whether such generalist
multi-purpose AI models alone are suitable for medical applications. The
distribution of human diseases is wide and complex, which is not
captured in current performance benchmarks, where these models are
predominantly evaluated on a single specific task at a time. In
contrast, real-world clinical decision-making often requires multi-step
reasoning, planning and repeated interactions with data to uncover new
insights in order to make informed and personalized decisions.

Despite advances with models like Med-PaLM
M\href{https://paperpile.com/c/q3Wf0T/52gVw}{\textsuperscript{7}}, the
complexities to develop a generalist foundation LLM that truly performs
on par with precision medicine tools remain a substantial challenge.
Moreover, it would be computationally expensive to frequently retrain
such a model to keep it up to date with the rapidly evolving medical
knowledge. Additionally, at present, regulatory policies in the United
States and the European Union restrict the approval of a universal
multi-purpose AI model under current regulations, given the philosophy
that medical devices should fulfill a singular
purpose\href{https://paperpile.com/c/q3Wf0T/pgkad}{\textsuperscript{8}}.

Previous work has shown that some of these limitations can partially be
overcome by enriching LLMs with domain-specific information. This can be
achieved through either fine-tuning (retraining the model on medical
data)\href{https://paperpile.com/c/q3Wf0T/oWEf6}{\textsuperscript{9}} or
retrieval-augmented generation
(RAG)\href{https://paperpile.com/c/q3Wf0T/lqGVf}{\textsuperscript{10}},
a process that temporarily enhances a LLM's knowledge by incorporating
relevant text excerpts from authoritative sources into the model, such
as medical
guidelines\href{https://paperpile.com/c/q3Wf0T/tkCIH}{\textsuperscript{11}}
or textbooks. Yet, this strategy, concentrating solely on augmenting the
knowledge base of the models, positions LLMs as mere information
extraction tools only, rather than serving as true clinical assistants.
Ideally, such a system would engage in reasoning, strategizing and
performing actions on patient records and retrieve or synthesize new
information to enable customized decision-making. Outside of the medical
field, several such autonomous AI systems - also termed agents - have
been proposed. Equipping a LLM with a suite of tools, like calculators
or web search, has proven superiority in tasks that require multi-step
reasoning and
planning\href{https://paperpile.com/c/q3Wf0T/a53pB+ZS0pc}{\textsuperscript{12,13}}.
Similarly, in biomedical research, Arasteh et al. utilized the
integrated data analysis tools of an LLM to analyze scientific data,
achieving results on par with human
researchers\href{https://paperpile.com/c/q3Wf0T/0lpPi}{\textsuperscript{14}}.
Such an approach would facilitate the opportunity of accessing the
information repositories that currently exist in hospital systems,
allowing for a true model for integrated patient
care\href{https://paperpile.com/c/q3Wf0T/cKaen}{\textsuperscript{15}}.

In this study, we extend the idea of autonomous LLMs that can employ
external tools to solve a given problem to the clinical domain, by
constructing an AI agent tailored to interact with and draw conclusions
from multimodal patient data through separate tools. Contrarily to the
philosophy of an all-encompassing multimodal generalist foundation
model, we see the achievements that specialist unimodal deep learning
models have brought to precision
medicine\href{https://paperpile.com/c/q3Wf0T/RY59p}{\textsuperscript{16}}
as a viable template even in the future by placing a LLM, specifically
GPT-4, at the core of a suite of precision oncology tools. These include
the vision model API dedicated to generating radiology reports from MRI
and CT scans,
MedSAM\href{https://paperpile.com/c/q3Wf0T/xU75x}{\textsuperscript{17}}
for medical image segmentation and in-house developed vision transformer
models trained to detect the presence of genetic alterations directly
from routine histopathology
slides\href{https://paperpile.com/c/q3Wf0T/O5lHG}{\textsuperscript{18}},
in particular, to distinguish between microsatellite instability (MSI)
and microsatellite stable tumors
(MSS)\href{https://paperpile.com/c/q3Wf0T/ad8wV}{\textsuperscript{19}}
and to detect the presence or absence of mutations in \emph{KRAS} and
\emph{BRAF}. Additionally, the system encompasses a basic calculator,
capabilities for conducting web searches via Google and PubMed, as well
as access to the precision oncology database
OncoKB\href{https://paperpile.com/c/q3Wf0T/9sn53}{\textsuperscript{20}}.
To ground the model\textquotesingle s reasoning on medical evidence, we
compile a repository of roughly 6,800 medical documents and clinical
scores from a collection of six different official sources, specifically
tailored to oncology.

To quantitatively test the performance of our proposed system, we devise
a new benchmark strategy. Existing biomedical benchmarks and evaluation
datasets are designed for one or two data
modalities\href{https://paperpile.com/c/q3Wf0T/gZE5n}{\textsuperscript{21}},
and are restricted to closed question-and-answer formats. Recent
advancements have been made with the introduction of new datasets by
Zakka et
al.\href{https://paperpile.com/c/q3Wf0T/tkCIH}{\textsuperscript{11}},
targeting the enhancement of open-ended responses, and
LongHealth\href{https://paperpile.com/c/q3Wf0T/8Iwko}{\textsuperscript{22}},
focusing on patient-related content. Still, these datasets are limited
to text and do not capture multimodal data, such as the combination of
CT or MRI images, microscopic and genetic data, alongside textual
reports. Therefore, in the present study, we develop and assess our
agent using a dataset comprising eleven realistic and multidimensional
patient cases, which we generate with a focus on gastrointestinal
oncology. For each patient case, the agent follows a two-stage process:
Upon receiving the clinical vignette and corresponding question, it
autonomously selects and applies relevant tools to derive supplementary
insights about the patient's condition, which is followed by the
document retrieval step to base its responses on substantiated medical
evidence, duly citing the relevant documents. To evaluate the results,
we designed a blinded manual evaluation by four human experts, focusing
on three areas: the agent's utilization of tools, the quality and
completeness of the textual outputs, and the precision in providing
relevant citations. For effective tool application, the agent must first
recognize the utility of a tool, comprehend the necessary inputs, and
then extract these inputs from the provided patient information. Our
pipeline is summarized in Figure 1, and we exemplify the agents workflow
on a representative patient case in Figure 2A. A detailed description of
our methodology is provided in the Methods section.

We investigated the overall ability of the agent to use appropriate
tools. We found that the agent consistently invoked these tools, on
average three times per patient case with some - expected - variability
(for instance, only one tool was used for patients T and Z, whereas for
patient X, all eight anticipated tool invocations were performed). Among
a total of 33 tool invocations across the 11 patient cases, we note just
one failure caused by calling an unrequired tool, which we detail in
Supplementary Table 1. Moreover, there is only one instance where a tool
deemed necessary by the human experts for answering the question was
inadvertently omitted by the agent (Figure 2B). Specifically, our
observations regarding the use of tools by the agent across the patient
as well as the model's adherence to the clinical state of the art are as
follows:

First, we specifically investigated the use of pathology image
processing tools by the model. Receiving accurate predictions for the
mutation status of cancer driver genes and microsatellite status is
crucial to make appropriate therapeutic recommendations in colorectal
cancer\href{https://paperpile.com/c/q3Wf0T/OyTMG}{\textsuperscript{23}}.
Based on previous
studies\href{https://paperpile.com/c/q3Wf0T/RY59p+O5lHG+qXEpM}{\textsuperscript{16,18,24}},
we provide the agent with AI tools to predict these alterations directly
from routine pathology images. In our evaluation, we find a high
accuracy of these models leading to correct predictions in all seven
patient cases where histopathology data, collected from The Cancer
Genome Atlas (TCGA), was included.

Second, we investigated the use of radiology processing tools by the
agent. This scenario has been previously investigated using the
vision-text AI model GPT-4V for medical image analysis, but has so far
only shown mixed results when using GPT-4V as the sole
model\textsuperscript{25}. We used GPT-4V as our radiology vision model
and tasked the agent with complex problems, including the comparison of
multiple radiology scans over time to evaluate disease progression or
stability. We found that despite occasional omissions, extraneous
details, lack of information or making mistakes (highlighted in red in
Supplementary Figure 4), GPT-4V nonetheless effectively guided clinical
decisions towards the accurate disease trajectory assessments in all
cases.

Third, we sought to assess the completeness of the model's responses,
specifically the model's ability to comprehensively address all
necessary aspects that oncologists would expect within a clinical
workflow (Suppl. Table 6). To facilitate this evaluation, a panel of
medical experts compiled a series of 67 essential statements for all
cases. Completeness was subsequently quantified as the proportion of
statements that were resolved by the model, which resulted in a rate of
94\%, with only 4 out of 67 instances not being covered in the model's
response. Remarkably, the agent is capable of resolving issues, even in
instances where contradictory information was provided in the patient's
description, such as inaccurately reported mutations. In such cases, the
agent pointed out these inconsistencies, recommended further genetic
confirmation and outlined potential treatment options based on the
results (Patient D and X).

Fourth, we assessed the degree of helpfulness of the model by evaluating
the proportion of sub-questions it answers sophistically according to
the human evaluators. Among the aggregate of 37 queries, 33 (89.2\%)
were categorized as having been effectively addressed. Next, a central
point in our analysis was the assessment of response accuracy.
Therefore, we segmented the responses into smaller, evaluable items
either upon the appearance of citations or when a transition in topic is
noted in the subsequent sentence, resulting in a total of 140 assessable
assertions. Our evaluations identify 131 (93.6\%) of these as factually
correct, 6 (4.3\%) as incorrect, and 3 (2.1\%) as potentially
detrimental. Instances of erroneous and harmful responses are
highlighted in Supplementary Table 3 for comprehensive review.

Fifth, aiming to ensure transparency in the decision-making process, we
instigated its adherence to citing relevant sources. Through manual
review, we determine that of the 171 citations provided in the models'
responses, 141 (82.5\%) are accurately aligned with the model's
assertions, while 11 (15.2\%) are found to be unrelated and merely 12
(2.3\%) references are found to be in conflict with the model's
statement. These findings are promising, highlighting that instances of
erroneous extrapolation (so termed hallucinations) by the model are
limited. Supplementary Tables 3 and 4 respectively display the entire
model outputs and the unprocessed complete results from using the tools,
with correct outputs highlighted in green and wrong ones marked in red.
Detailed evaluation results from each human observer are elaborated on
in Supplementary Tables 5, 6 and 7.

In summary, our results demonstrate that combining precision medicine
solutions with an LLM agent enhances its capabilities in problem
solving, aligning with the concept of utilizing LLMs as `reasoning
engines'\textsuperscript{26} rather than merely a repository of medical
knowledge. Integrating three core elements - reasoning engine, a
knowledge database, and tools - enables us to address several
limitations in current concepts: Despite the potential future
development of a generalist medical multimodal foundation model, its
efficacy in addressing very specialized medical queries, such as
predicting rare mutations or measuring disease development on the
millimeter level scale, compared to narrower, domain-specific models
remains uncertain. Moreover, maintaining the alignment of such a
generalist model with the evolving medical knowledge and updates in
treatment guidelines is challenging, as it requires retraining model
components on new data. Our approach addresses all of these issues: It
allows for the rapid update of medical knowledge by simply replacing
pertinent documents in the database or information retrieval from google
search and PubMed, eliminating the need for direct modifications to the
core model itself. Similarly, state-of-the-art medical devices that are
approved by regulatory authorities can be included in our setup, and can
be easily updated.

Our work has several limitations. The agent, though equipped with a
broad array of tools compared to other
frameworks\href{https://paperpile.com/c/q3Wf0T/a53pB}{\textsuperscript{12}},
remains in a premature and experimental stage, thus limiting clinical
applicability. One notable restriction for instance lies in the
provision of only a singular slice of radiology images and the yet
limited capabilities of GPT-4V in interpreting medical images.
Additionally, despite being implemented as a chat-agent, our evaluation
is currently confined to a single interaction without follow-up
questions for the sake of simplicity. Furthermore, we restrict the
setting to oncological use cases; yet it is important to note that the
underlying framework could be adapted to virtually any medical
speciality, given the appropriate tools and data.

Looking ahead, we anticipate more progress in the development of AI
agents - LLMs that act as operating systems - with even improved
capabilities through further scaling\textsuperscript{27}. In the near
future, we envision a framework that embodies characteristics akin to
the GMAI model with the added ability to access precision medicine tools
tailored to answer specialized clinical questions. This approach has
multiple benefits: It enables medical AI models to assist clinicians in
solving real-world patient scenarios using precision medicine tools,
each tailored to specific tasks. Such a strategy facilitates
circumventing data availability constraints inherent in the medical
domain, where data is not uniformly accessible across all disciplines,
preventing a singular entity from developing an all-encompassing
foundational model. Instead, entities can leverage smaller, specialized
models developed by those with direct access to the respective data,
which could greatly improve discovering therapeutic options for
personalized treatments. Moreover, this modular approach allows for the
individual validation, updating, and regulatory compliance of each tool.
In cases where existing tools are unsatisfactory or completely absent,
the agent could rely on its internal strong medical domain knowledge
and, additionally, either refine\textsuperscript{28} or innovate
entirely new tools from scratch. Herein, our study could serve as a
blueprint, providing evidence that agent based generalist medical AI is
within reach.

\section{References}\label{references}

1. \href{http://paperpile.com/b/q3Wf0T/QfTSp}{Acosta, J. N., Falcone, G.
J., Rajpurkar, P. \& Topol, E. J. Multimodal biomedical AI. \emph{Nat.
Med.} \textbf{28}, 1773--1784 (2022).}

2. \href{http://paperpile.com/b/q3Wf0T/TvfrY}{Lipkova, J. \emph{et al.}
Artificial intelligence for multimodal data integration in oncology.
\emph{Cancer Cell} \textbf{40}, 1095--1110 (2022).}

3. \href{http://paperpile.com/b/q3Wf0T/7tuPM}{Khader, F. \emph{et al.}
Multimodal Deep Learning for Integrating Chest Radiographs and Clinical
Parameters: A Case for Transformers. \emph{Radiology} \textbf{309},
e230806 (2023).}

4. \href{http://paperpile.com/b/q3Wf0T/OpwMD}{Chen, R. J. \emph{et al.}
Pan-cancer integrative histology-genomic analysis via multimodal deep
learning. \emph{Cancer Cell} \textbf{40}, 865--878.e6 (2022).}

5. \href{http://paperpile.com/b/q3Wf0T/VXY2u}{Lu, M. Y. \emph{et al.} A
Foundational Multimodal Vision Language AI Assistant for Human
Pathology. \emph{arXiv {[}cs.CV{]}} (2023).}

6. \href{http://paperpile.com/b/q3Wf0T/nGI1G}{Moor, M. \emph{et al.}
Foundation models for generalist medical artificial intelligence.
\emph{Nature} \textbf{616}, 259--265 (2023).}

7. \href{http://paperpile.com/b/q3Wf0T/52gVw}{Tu Tao \emph{et al.}
Towards Generalist Biomedical AI. \emph{NEJM AI} \textbf{1}, AIoa2300138
(2024).}

8. \href{http://paperpile.com/b/q3Wf0T/pgkad}{Derraz, B. \emph{et al.}
New regulatory thinking is needed for AI-based personalised drug and
cell therapies in precision oncology. \emph{NPJ Precis Oncol}
\textbf{8}, 23 (2024).}

9. \href{http://paperpile.com/b/q3Wf0T/oWEf6}{Chen, Z. \emph{et al.}
MEDITRON-70B: Scaling Medical Pretraining for Large Language Models.
\emph{arXiv {[}cs.CL{]}} (2023).}

10. \href{http://paperpile.com/b/q3Wf0T/lqGVf}{Lewis, P. \emph{et al.}
Retrieval-augmented generation for knowledge-intensive NLP tasks.
\emph{Adv. Neural Inf. Process. Syst.} \textbf{abs/2005.11401}, (2020).}

11. \href{http://paperpile.com/b/q3Wf0T/tkCIH}{Zakka, C. \emph{et al.}
Almanac - Retrieval-Augmented Language Models for Clinical Medicine.
\emph{NEJM AI} \textbf{1}, (2024).}

12. \href{http://paperpile.com/b/q3Wf0T/a53pB}{Yao, S. \emph{et al.}
ReAct: Synergizing Reasoning and Acting in Language Models. \emph{arXiv
{[}cs.CL{]}} (2022).}

13. \href{http://paperpile.com/b/q3Wf0T/ZS0pc}{Schick, T. \emph{et al.}
Toolformer: Language models can teach themselves to use tools.
\emph{Adv. Neural Inf. Process. Syst.} \textbf{abs/2302.04761}, (2023).}

14. \href{http://paperpile.com/b/q3Wf0T/0lpPi}{Tayebi Arasteh, S.
\emph{et al.} Large language models streamline automated machine
learning for clinical studies. \emph{Nat. Commun.} \textbf{15}, 1603
(2024).}

15. \href{http://paperpile.com/b/q3Wf0T/cKaen}{Messiou, C., Lee, R. \&
Salto-Tellez, M. Multimodal analysis and the oncology patient: Creating
a hospital system for integrated diagnostics and discovery.
\emph{Comput. Struct. Biotechnol. J.} \textbf{21}, 4536--4539 (2023).}

16. \href{http://paperpile.com/b/q3Wf0T/RY59p}{Kather, J. N. \emph{et
al.} Deep learning can predict microsatellite instability directly from
histology in gastrointestinal cancer. \emph{Nat. Med.} \textbf{25},
1054--1056 (2019).}

17. \href{http://paperpile.com/b/q3Wf0T/xU75x}{Ma, J. \emph{et al.}
Segment anything in medical images. \emph{Nat. Commun.} \textbf{15}, 654
(2024).}

18. \href{http://paperpile.com/b/q3Wf0T/O5lHG}{Wagner, S. J. \emph{et
al.} Transformer-based biomarker prediction from colorectal cancer
histology: A large-scale multicentric study. \emph{Cancer Cell}
\textbf{41}, 1650--1661.e4 (2023).}

19. \href{http://paperpile.com/b/q3Wf0T/ad8wV}{El Nahhas, O. S. M.
\emph{et al.} Joint multi-task learning improves weakly-supervised
biomarker prediction in computational pathology. \emph{arXiv
{[}eess.IV{]}} (2024).}

20. \href{http://paperpile.com/b/q3Wf0T/9sn53}{Chakravarty, D. \emph{et
al.} OncoKB: A Precision Oncology Knowledge Base. \emph{JCO Precis
Oncol} \textbf{2017}, (2017).}

21. \href{http://paperpile.com/b/q3Wf0T/gZE5n}{He, X., Zhang, Y., Mou,
L., Xing, E. \& Xie, P. PathVQA: 30000+ Questions for Medical Visual
Question Answering. \emph{arXiv {[}cs.CL{]}} (2020).}

22. \href{http://paperpile.com/b/q3Wf0T/8Iwko}{Adams, L. \emph{et al.}
LongHealth: A Question Answering Benchmark with Long Clinical Documents.
\emph{arXiv {[}cs.CL{]}} (2024).}

23. \href{http://paperpile.com/b/q3Wf0T/OyTMG}{Cervantes, A. \emph{et
al.} Metastatic colorectal cancer: ESMO Clinical Practice Guideline for
diagnosis, treatment and follow-up. \emph{Ann. Oncol.} \textbf{34},
10--32 (2023).}

24. \href{http://paperpile.com/b/q3Wf0T/qXEpM}{Saillard, C. \emph{et
al.} Validation of MSIntuit as an AI-based pre-screening tool for MSI
detection from colorectal cancer histology slides. \emph{Nat. Commun.}
\textbf{14}, 6695 (2023).}

25. OpenAI. GPT-4V(ision) system card.
\href{https://cdn.openai.com/papers/GPTV_System_Card.pdf}{https://cdn.openai.com/papers/GPTV\_System\_Card.pdf}

26. \href{http://paperpile.com/b/q3Wf0T/scHQM}{Truhn, D., Reis-Filho, J.
S. \& Kather, J. N. Large language models should be used as scientific
reasoning engines, not knowledge databases. \emph{Nat. Med.}
\textbf{29}, 2983--2984 (2023).}

27. \href{http://paperpile.com/b/q3Wf0T/xAcme}{Wei, J. \emph{et al.}
Emergent Abilities of Large Language Models. \emph{arXiv {[}cs.CL{]}}
(2022).}

28. Anthropic. The Claude 3 Model Family: Opus, Sonnet, Haiku.
\href{https://www-cdn.anthropic.com/de8ba9b01c9ab7cbabf5c33b80b7bbc618857627/Model\_Card\_Claude\_3.pdf}
{https://www-cdn.anthropic.com/de8ba9b01c9ab7cbabf5c33b80b7bbc618857627/Model\_Card\_Claude\_3.pdf}

\section{\texorpdfstring{\textbf{Methods}}{Methods}}\label{methods}

\subsection{Dataset composition and data
collection}\label{dataset-composition-and-data-collection}

The pipeline's primary goal is to compile a comprehensive dataset from
high-quality medical sources, ensuring three main components:
correctness, up-to-dateness and contextual relevance, with a particular
emphasis on including knowledge across all medical domains while
additionally encompassing information specifically tailored to oncology:
We restrict our data access to the following six sources:
MDCalc\href{https://paperpile.com/c/biD0Jy/FDHh4}{\textsuperscript{1}}
for clinical scores,
UpToDate\href{https://paperpile.com/c/biD0Jy/05Oi0}{\textsuperscript{2}}
and
MEDITRON\href{https://paperpile.com/c/biD0Jy/0RIXn}{\textsuperscript{3}}
for general purpose medical recommendations and the Clinical Practice
Guidelines from the American Society of Clinical Oncology
(ASCO)\href{https://paperpile.com/c/biD0Jy/he4CN}{\textsuperscript{4}},
the European Society of Medical Oncology
(ESMO)\href{https://paperpile.com/c/biD0Jy/b24LG}{\textsuperscript{5}}
as well as the german and english subset of the onkopedia guidelines
from the German Society for Hematology and Medical Oncology
(DGHO)\href{https://paperpile.com/c/biD0Jy/BIEu8}{\textsuperscript{6}}.
We retrieve and download the relevant documents as either HTML extracted
text or raw PDF files. To reduce the number of documents for the
embedding step, we apply a keyword-based filtering of the documents
contents, targeting terms relevant to our specific use case. Medical
guidelines that were obtained from the MEDITRON project were directly
accessible as preprocessed jsonlines file.

\subsection{Information extraction and data curation from PDF
files}\label{information-extraction-and-data-curation-from-pdf-files}

The critical challenge in text extraction from PDF documents arises from
the inherent nature of PDF files, which are organized primarily for the
user\textquotesingle s ease of reading and display while not adhering to
a consistent hierarchical structure, which complicates the extraction
process. For instance, upon text mining with conventional tools like
PyPDF2\href{https://paperpile.com/c/biD0Jy/jJ2No}{\textsuperscript{7}}
or
PyMuPDF\href{https://paperpile.com/c/biD0Jy/BjFqv}{\textsuperscript{8}},
headers, subheaders and key information from the main text may be
irregularly placed, with titles occasionally embedded within paragraphs
and critical data abruptly interspersed within unrelated text. However,
maintaining the integrity of the original document's structure is
crucial in the medical field to ensure that extracted information
remains contextually coherent, preventing any conflation or
misinterpretation. To overcome these limitations, we utilized
GROBID\href{https://paperpile.com/c/biD0Jy/UpbCO}{\textsuperscript{9}}
(GeneRation Of BIbliographic Data), a Java application and machine
learning library specifically developed for the conversion of
unstructured PDF data into a standardized
TEI\href{https://paperpile.com/c/biD0Jy/dcv9E}{\textsuperscript{10}}
(Text Encoding Initiative) format. Through its particular training on
scientific and technical articles, GROBID enables the effective parsing
of medical documents, preserving text hierarchy and generating essential
metadata such as document and journal titles, authorship, pagination,
publication dates and download URLs.

We next programmatically retrieved the raw document text from the
generated XML fields in the TEI files, concurrently implementing data
cleansing. This process encompasses the removal of extraneous and
irrelevant information such as hyperlinks, graphical elements and
tabular data that was corrupted during the extraction with GROBID as
well as any malformed characters or data like inadvertently extracted IP
addresses. The diversity of source materials presented a further
challenge due to their varied formatting schemas. To address this, we
meticulously reformatted and standardized the text from all sources,
denoting headers with a preceding hash symbol (\#) and inserting blank
lines for the separation of paragraphs. The purified text along with its
corresponding metadata was archived in jsonlines format for subsequent
processing.

\subsection{Agent Composition: Retrieval-Augmented
Generation}\label{agent-composition-retrieval-augmented-generation}

In the following, we delineate the detailed architecture of our agent in
a two-step process, beginning with the creation of our
Retrieval-Augmented Generation
(RAG)\href{https://paperpile.com/c/biD0Jy/JFMee}{\textsuperscript{11}}
database, followed by an overview of the agent's tool utilization and
conclude with an examination of the final retrieval and response
generation modules. Additionally, we highlight the structure of our
model in detail in Algorithm 1, provided in the Supplementary Material.

\subsubsection{Embedding creation and
indexing}\label{embedding-creation-and-indexing}

We leveraged RAG to synergize the generative capabilities of LLMs with
document retrieval to provide domain-specific medical knowledge
(\emph{context}) to a model. The RAG framework has significantly evolved
in complexity recently, so we break down its architecture into three
major components (embeddings, indexing and retrieval) and outline the
implementation details of the first two in the following section. In
RAG, we begin with the conversion of raw text data into numerical
(vector) representations, also termed \emph{embeddings}, which are
consequently stored in a vector database alongside metadata and the
corresponding original text (\emph{indexing}). In more detail, we
compute vector embeddings using OpenAI's `text-embedding-3-large' model
from text segments of varying lengths (512, 256 and 128 tokens), each
featuring a 50-token overlap, from the curated guideline cleaned main
texts in our dataset, alongside their associated metadata for potential
filtering operations. For storage, we employ an instance of a local
vector
database\href{https://paperpile.com/c/biD0Jy/CBoEl}{\textsuperscript{12}}
that also facilitates highly efficient lookup operations via vector
based similarity measures like cosine similarity (dense retrieval). We
store documents from different sources in the same collection.

\subsection{Agent composition: Tools}\label{agent-composition-tools}

To endow the LLM with agentic capabilities, we equipped it with an array
of tools, including the ability to conduct web searches through the
Google Custom Search API and formulate custom PubMed queries.
Information retrieved through Google Search underwent text extraction
and purification and was being integrated directly as context within the
model, while responses from PubMed were processed akin to the above
described RAG procedure in a separate database. For the interpretation
of visual data, such as CT or MRI scans, the LLM-agent has the capacity
to invoke the GPT-4 Vision model which is instructed to provide a
detailed and structured report. In scenarios involving multiple images,
the model first investigates and reports on each image separately prior
to synthesizing a comparative analysis. Due to the stringent adherence
of OpenAI to ethical guidelines, particularly concerning the management
of medical image data, we framed our patient cases as hypothetical
scenarios when presenting them to the model. However, instances of
refusal still arise, prompting us to discard the respective run entirely
and initiate a new one from the beginning. Additionally, tasks that
require a segmentation mask can be completed using
MedSAM\href{https://paperpile.com/c/biD0Jy/2tkEt}{\textsuperscript{13}}.
Moreover, we provided access to a simplified calculation tool that
allows elementary arithmetic operations such as addition, subtraction,
multiplication, and division.

To facilitate addressing queries related to precision oncology, the LLM
leverages the
OncoKB\href{https://paperpile.com/c/biD0Jy/blrPu}{\textsuperscript{14}}
database to access critical information on medical evidence for treating
a vast panel of genetic anomalies, including mutations, copy number
alterations and structural rearrangements. Lastly, GPT-4 is also
equipped to engage specialized vision transformer models for the
histopathological analysis of phenotypic alterations underlying
MSI\href{https://paperpile.com/c/biD0Jy/Y8jQ2}{\textsuperscript{15}} or
\emph{KRAS} and \emph{BRAF}
mutations\href{https://paperpile.com/c/biD0Jy/syenx}{\textsuperscript{16}}.
All necessary information for calling the designated tools are derivable
or producible from the given patient context. Unlike the retrieval
phase, which we manually enforced at each invocation, the decision
regarding the utilization and timing of tools is entrusted entirely to
the agent\textquotesingle s reasoning. However, manual intervention to
prompt tool usage is possible, as demonstrated in patient cases D and X.
The specifications for all tools are delineated in JavaScript Object
Notation (JSON) which is provided to the model and encompasses a brief
textual description of each tool's function along with the required
input parameters. From a procedural point of view, given an input
comprising a variable-length textual patient context and a text query,
the agent generates an initial action plan, followed by a series of
iterative tool applications. The deployment of these tools can be
executed either independently in parallel or sequentially, wherein the
output from one tool serves as the input for another in subsequent
rounds; for instance the size of the segmentation areas obtained from
two images via MedSAM can be utilized to compute a ratio and thus define
disease progression, stability or response, as shown in Figure 1.

\subsection{Agent composition: Combine, retrieve and generate
responses}\label{agent-composition-combine-retrieve-and-generate-responses}

The final retrieval and response generation pipeline is implemented
using
DSPy\href{https://paperpile.com/c/biD0Jy/7JKHw}{\textsuperscript{17}}, a
library that allows for a modular composition of LLM calls. Firstly, the
model receives the original patient context, the posed question and the
outcomes from the tool applications as input. In a method similar to
that described by Xiong et
al\href{https://paperpile.com/c/biD0Jy/Ech6Z}{\textsuperscript{18}}, we
employed Chain-of-Thought
reasoning\href{https://paperpile.com/c/biD0Jy/8ZRLk}{\textsuperscript{19}}
to let the model decompose the initial user query into up to twelve more
granular subqueries derived from both the initial patient context and
the outcomes from tool applications. This facilitates the retrieval of
documents from the vector database that more closely align with each
aspect of a multi-faceted user query. Precisely, for each generated
subquery we extract the top \emph{k} most analogous document passages
from the collection. Subsequently, this data is combined, deduplicated,
reranked\href{https://paperpile.com/c/biD0Jy/4qzoC}{\textsuperscript{20}}
and finally forwarded to the LLM. Prior to generating the final answer,
we instruct the LLM to generate a step-by-step strategy to build a
structured response including identifying missing information that could
help refine and personalize the recommendations. The resulting model
output is then synthesized based on all available information, strictly
following the strategy as a hierarchical blueprint. To enhance the
system's reliability and enable thorough fact-checking - both of which
are fundamental in real-world medical applications - the model was
programmatically configured to incorporate citations for each statement
(as defined as a maximum of two consecutive sentences) using DSPy
suggestions\href{https://paperpile.com/c/biD0Jy/7JKHw}{\textsuperscript{17}}.
On the implementation level, the LLM performs a self-evaluation step,
wherein it compares its own output to the respective context from our
database in a one to two sentences window. We perform a single iteration
over this procedure. All prompts are implemented using DSPy's
signatures.

\subsection{Model Specifications}\label{model-specifications}

In our study, we consistently used the following models through the
official OpenAI Python API for all experiments, performed on March 4 and
March 13 2024. The core framework for the agent and all tools involving
an LLM is the \emph{gpt-4-0125-preview} model\emph{,} for brevity
henceforth referred to as GPT-4. For tasks requiring visual processing,
the \emph{gpt-4-vision-preview} (GPT-4V) model was used via the chat
completions endpoint. The temperature value for both models was
empirically set to 0.1 upon initial experimentation and no further
modifications of model hyperparameters were performed. Additionally, for
generating text embeddings, we utilized the latest version of OpenAI's
embedding models, specifically the \emph{text-embedding-3-large} model,
which produces embeddings with a dimensionality of 3,072.

\subsection{Clinical Case Generation}\label{clinical-case-generation}

To address the limitations in current biomedical benchmarks, we compiled
a collection of eleven distinct multimodal patient cases, primarily
focusing on gastrointestinal oncology, including colorectal, pancreatic,
cholangiocellular and hepatocellular cancers. Each case provides a
comprehensive but entirely fictional patient profile, which includes a
concise medical history overview encompassing diagnoses, significant
medical events, and previous treatments. We pair each patient with
either a single (for 3 out of 11 cases) or two slices of CT or MRI
imaging that serve as either sequential follow-up staging scans of the
liver (six out of eleven) or lungs (in one case) or simultaneous staging
scans of both the liver and lungs at a single point in time (one case).
Images are obtained from the web, the Cancer Imaging
Archive\href{https://paperpile.com/c/biD0Jy/VSUpP+y56Kq}{\textsuperscript{21,22}}
and internally from the Department of Diagnostic and Interventional
Radiology, University Hospital Aachen, Germany. Histology images are
present in seven out of the eleven cases and are obtained from The
Cancer Genome Atlas (TCGA). We include information into genomic
variations (mutations and gene fusions) in four patient descriptions. To
evaluate our model's proficiency in handling complex information, we
decide to not pose a single straightforward question but instead
structure each query with multiple subtasks, subquestions and
instructions, necessitating the model to handle an average of three to
four subtasks in each round.

\subsection{Human results evaluation}\label{human-results-evaluation}

To enhance the assessment of free-text output, we developed a structured
evaluation framework, drawing inspiration from the methodology of
Singhal et
al.\href{https://paperpile.com/c/biD0Jy/ETYfw}{\textsuperscript{23}} Our
evaluation focuses on three primary aspects: the use of tools by the
agent, the quality of the text output produced by the model and the
adherence in providing accurate citations.
In reference to the former, we established a manual baseline for the
expected utilization of tools necessary for generating additional
patient information that is crucial for resolving the patient's task. We
measured this by the ratio of actual versus expected tool uses. The
expectation of tool use was defined as either the model is directly
instructed to use a certain tool, or the output of a tool is essential
to proceed in answering the question - which is the default setting in
almost all situations. Additionally, we assessed the helpfulness of the
model, quantified by the proportion of user instructions and
subquestions directly addressed and resolved by the model.
In assessing the textual outputs, our evaluation first encompassed
factual correctness, defined by the proportion of correct replies
relative to all model outputs. To segment answers into more manageable
units, we split each reply statement-wise (where a statement is
considered a segment that concludes with either a reference to
literature or is followed by a shift in topic in the subsequent
sentence). Correspondingly, we distinguish between incorrectness and
harmfulness in responses. Incorrect responses may include suggestions
for superfluous diagnostic procedures or contain requests for irrelevant
patient information. Conversely, harmful responses, while also
incorrect, are determined by clinical judgment as potentially
deleterious, such as advising suboptimal or contraindicated treatments.
Furthermore, we assess the comprehensiveness of the responses. For this
purpose, we identify on average five to ten specific keywords and terms
for each unique medical scenario. These keywords represent expected
interventions, such as treatments or diagnostic procedures and are
carefully selected for their case relevance and crafted to be as
specific as possible (e.g. precise treatment combinations like `FOLFOX
and bevacizumab' instead of `chemotherapy and antiangiogenic drugs').
This criterion, which we term `completeness', is supposed to measure the
extent to which the agent's response aligns with the essential
information that oncologists would anticipate in a human-generated
answer under similar conditions.
Lastly, we evaluate the alignment of the responses with the original
document segments utilized by the model through RAG. For each reference
in the model's output, we investigate the corresponding reference by its
source ID. Our evaluation encompasses three critical dimensions:
citation correctness, ensuring the model's statements faithfully mirror
the content of the original document; irrelevance, identifying instances
where the model's assertions are not substantiated by the source
material; and incorrect citation, detecting discrepancies where the
information attributed to a source diverges from its actual content.
In cases of a tie, we select the most adverse outcome, adhering to a
hierarchical schema: correct, irrelevant and wrong.
All evaluations described here are performed independently by four
certified clinicians with expertise in oncology.

\section{\texorpdfstring{\textbf{Additional
References}}{Additional References}}\label{additional-references}

1. \href{http://paperpile.com/b/biD0Jy/FDHh4}{MDCalc - Medical
calculators, equations, scores, and guidelines. \emph{MDCalc}}
\url{https://www.mdcalc.com/}\href{http://paperpile.com/b/biD0Jy/FDHh4}{.}

2. \href{http://paperpile.com/b/biD0Jy/05Oi0}{\emph{Three Decades of
UpToDate}. (2023).}

3. \href{http://paperpile.com/b/biD0Jy/0RIXn}{Chen, Z. \emph{et al.}
MEDITRON-70B: Scaling Medical Pretraining for Large Language Models.
\emph{arXiv {[}cs.CL{]}} (2023).}

4. \href{http://paperpile.com/b/biD0Jy/he4CN}{Lyman, G. H. ASCO Clinical
Practice Guidelines and Beyond. \emph{J. Oncol. Pract.} \textbf{3},
330--331 (2007).}

5. \href{http://paperpile.com/b/biD0Jy/b24LG}{ESMO. Guidelines by
topic.}
\url{https://www.esmo.org/guidelines/guidelines-by-topic}\href{http://paperpile.com/b/biD0Jy/b24LG}{.}

6. \href{http://paperpile.com/b/biD0Jy/BIEu8}{Aden, T. Startseite. in
\emph{Google Analytics} 317--370 (Carl Hanser Verlag GmbH \& Co. KG,
München, 2012).}

7. \href{http://paperpile.com/b/biD0Jy/jJ2No}{Welcome to pypdf --- pypdf
4.0.1 documentation.}
\url{https://pypdf.readthedocs.io/en/stable/index.html}\href{http://paperpile.com/b/biD0Jy/jJ2No}{.}

8. \href{http://paperpile.com/b/biD0Jy/BjFqv}{Artifex. Module fitz -
PyMuPDF 1.23.21 documentation.}
\url{https://pymupdf.readthedocs.io/en/latest/module.html}\href{http://paperpile.com/b/biD0Jy/BjFqv}{.}

9. \href{http://paperpile.com/b/biD0Jy/UpbCO}{GROBID Documentation.}
\url{https://grobid.readthedocs.io/en/latest/}\href{http://paperpile.com/b/biD0Jy/UpbCO}{.}

10. \href{http://paperpile.com/b/biD0Jy/dcv9E}{Ide, N. \& Véronis, J.
Text Encoding Initiative. \emph{Springer Neth} (1995)
doi:}\href{http://dx.doi.org/10.1007/978-94-011-0325-1}{10.1007/978-94-011-0325-1}\href{http://paperpile.com/b/biD0Jy/dcv9E}{.}

11. \href{http://paperpile.com/b/biD0Jy/JFMee}{Lewis, P. \emph{et al.}
Retrieval-augmented generation for knowledge-intensive NLP tasks.
\emph{Adv. Neural Inf. Process. Syst.} \textbf{abs/2005.11401}, (2020).}

12. \href{http://paperpile.com/b/biD0Jy/CBoEl}{the AI-native open-source
embedding database.}
\url{https://www.trychroma.com/}\href{http://paperpile.com/b/biD0Jy/CBoEl}{.}

13. \href{http://paperpile.com/b/biD0Jy/2tkEt}{Ma, J. \emph{et al.}
Segment anything in medical images. \emph{Nat. Commun.} \textbf{15}, 654
(2024).}

14. \href{http://paperpile.com/b/biD0Jy/blrPu}{Chakravarty, D. \emph{et
al.} OncoKB: A Precision Oncology Knowledge Base. \emph{JCO Precis
Oncol} \textbf{2017}, (2017).}

15. \href{http://paperpile.com/b/biD0Jy/Y8jQ2}{El Nahhas, O. S. M.
\emph{et al.} Joint multi-task learning improves weakly-supervised
biomarker prediction in computational pathology. \emph{arXiv
{[}eess.IV{]}} (2024).}

16. \href{http://paperpile.com/b/biD0Jy/syenx}{Wagner, S. J. \emph{et
al.} Transformer-based biomarker prediction from colorectal cancer
histology: A large-scale multicentric study. \emph{Cancer Cell}
\textbf{41}, 1650--1661.e4 (2023).}

17. \href{http://paperpile.com/b/biD0Jy/7JKHw}{Khattab, O. \emph{et al.}
DSPy: Compiling Declarative Language Model Calls into Self-Improving
Pipelines. \emph{arXiv {[}cs.CL{]}} (2023).}

18. \href{http://paperpile.com/b/biD0Jy/Ech6Z}{Xiong, W. \emph{et al.}
Answering Complex Open-Domain Questions with Multi-Hop Dense Retrieval.
\emph{arXiv {[}cs.CL{]}} (2020).}

19. \href{http://paperpile.com/b/biD0Jy/8ZRLk}{Wei, J. \emph{et al.}
Chain-of-Thought Prompting Elicits Reasoning in Large Language Models.
\emph{arXiv {[}cs.CL{]}} (2022).}

20. \href{http://paperpile.com/b/biD0Jy/4qzoC}{Reimers, N. Say Goodbye
to Irrelevant Search Results: Cohere Rerank Is Here. \emph{Context by
Cohere}} \url{https://txt.cohere.com/rerank/}
\href{http://paperpile.com/b/biD0Jy/4qzoC}{(2023).}

21.
\href{http://paperpile.com/b/biD0Jy/VSUpP}{COLORECTAL-LIVER-METASTASES -
the cancer imaging archive (TCIA). \emph{The Cancer Imaging Archive
(TCIA)}}
\url{https://www.cancerimagingarchive.net/collection/colorectal-liver-metastases/}
\href{http://paperpile.com/b/biD0Jy/VSUpP}{(2023).}

22. \href{http://paperpile.com/b/biD0Jy/y56Kq}{Clark, K. \emph{et al.}
The Cancer Imaging Archive (TCIA): maintaining and operating a public
information repository. \emph{J. Digit. Imaging} \textbf{26}, 1045--1057
(2013).}

23. \href{http://paperpile.com/b/biD0Jy/ETYfw}{Singhal, K. \emph{et al.}
Large language models encode clinical knowledge. \emph{Nature}
\textbf{620}, 172--180 (2023).}

\section{\texorpdfstring{\textbf{Additional
Information}}{Additional Information}}\label{additional-information}

\subsection{\texorpdfstring{\textbf{Data availability
statement}}{Data availability statement}}\label{data-availability-statement}

We plan to release our source codes for researchers to extend on our
work upon publication in a scientific journal.

\subsection{\texorpdfstring{\textbf{Ethics
statement}}{Ethics statement}}\label{ethics-statement}

This study does not include confidential information. All research
procedures were conducted exclusively on publicly accessible, anonymized
patient data and in accordance with the Declaration of Helsinki,
maintaining all relevant ethical standards. The overall analysis was
approved by the Ethics commission of the Medical Faculty of the
Technical University Dresden (BO-EK-444102022).

\subsection{\texorpdfstring{\textbf{Acknowledgements}}{Acknowledgements}}\label{acknowledgements}

The results generated in our study are in part based upon data generated
by the TCGA Research Network: https://www.cancer.gov/tcga.

\subsection{\texorpdfstring{\textbf{Author
Contributions}}{Author Contributions}}\label{author-contributions}

DF designed and performed the experiments, evaluated and interpreted the
results and wrote the initial draft of the manuscript. OSMEN and GW
provided scientific support for running the experiments and contributed
to writing the manuscript. ICW contributed to writing the manuscript. DJ
supervised the study. DT and JNK designed and supervised the experiments
and wrote the manuscript. All authors contributed scientific advice and
approved the final version of the manuscript.

\subsection{\texorpdfstring{\textbf{Funding}}{Funding}}\label{funding}

JNK is supported by the German Federal Ministry of Health (DEEP LIVER,
ZMVI1-2520DAT111; SWAG, 01KD2215B), the Max-Eder-Programme of the German
Cancer Aid (grant \#70113864), the German Federal Ministry of Education
and Research (PEARL, 01KD2104C; CAMINO, 01EO2101; SWAG, 01KD2215A;
TRANSFORM LIVER, 031L0312A; TANGERINE, 01KT2302 through ERA-NET
Transcan), the German Academic Exchange Service (SECAI, 57616814), the
German Federal Joint Committee (Transplant.KI, 01VSF21048) the European
Union's Horizon Europe and innovation programme (ODELIA, 101057091;
GENIAL, 101096312) and the National Institute for Health and Care
Research (NIHR, NIHR213331) Leeds Biomedical Research Centre. DT is
funded by the German Federal Ministry of Education and Research
(TRANSFORM LIVER, 031L0312A), the European Union's Horizon Europe and
innovation programme (ODELIA, 101057091), and the German Federal
Ministry of Health (SWAG, 01KD2215B). GW is supported by Lothian NHS. JC
is supported by the Mildred-Scheel-Postdoktorandenprogramm of the German
Cancer Aid (grant \#70115730). JL is supported by the TUM School of
Medicine and Health Clinician Scientist Program (project no. H-08). JL
receives intellectual and financial support through the DKTK School of
Oncology Fellowship. SF is supported by the German Federal Ministry of
Education and Research (SWAG, 01KD2215A), the German Cancer Aid (DECADE,
70115166) and the German Research Foundation (504101714). The views
expressed are those of the author(s) and not necessarily those of the
NHS, the NIHR or the Department of Health and Social Care. No other
funding is disclosed by any of the authors.

\subsection{\texorpdfstring{\textbf{Competing
Interests}}{Competing Interests}}\label{competing-interests}

OSMEN holds shares in StratifAI GmbH. JNK declares consulting services
for Owkin, France; DoMore Diagnostics, Norway; Panakeia, UK, and
Scailyte, Basel, Switzerland; furthermore JNK holds shares in Kather
Consulting, Dresden, Germany; and StratifAI GmbH, Dresden, Germany, and
has received honoraria for lectures and advisory board participation by
AstraZeneca, Bayer, Eisai, MSD, BMS, Roche, Pfizer and Fresenius. DT
received honoraria for lectures by Bayer and holds shares in StratifAI
GmbH, Germany. The authors have no additional financial or non-financial
conflicts of interest to disclose. MST is a scientific advisor to
Mindpeak and Sonrai Analytics and has received honoraria recently from
BMS, MSD, Roche, Sanofi, and Incyte. SF has received honoraria from MSD
and BMS.

\section{\texorpdfstring{\textbf{Figures}}{Figures}}\label{figures}

\begin{figure}[htbp]
  \centering
  \includegraphics[width=\linewidth]{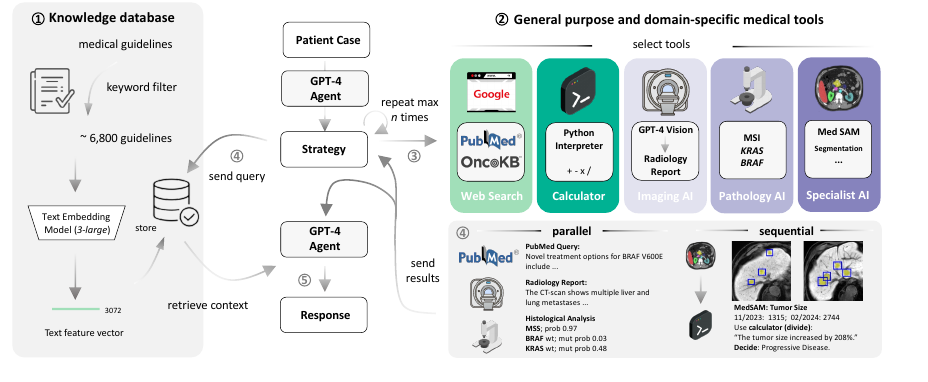}
  \caption{High-level overview of the RAG-Agents framework. \\
  At its core, the system accesses a curated knowledge database comprising
medical documents, clinical guidelines, and scoring tools. This database
is refined from a broader collection through keyword-based search, with
the selected documents undergoing text embeddings for efficient storage
and retrieval (1). The framework is further augmented with a suite of
medical tools, including specialized web search capabilities through
platforms such as Google, PubMed, and OncoKB\textquotesingle s
open-access version. The agent\textquotesingle s capabilities are
expanded through the integration of a vision model tailored for
generating detailed reports from CT and MRI scans, alongside Med-SAM, a
state-of-the-art medical image segmentation model and access to a simple
calculator. Additionally, the system employs vision transformers
specifically developed for the prediction of microsatellite instability
(MSI vs. MSS) and the detection of \emph{KRAS} and \emph{BRAF} mutations
in microscopic tumor samples (2). All tools are selected autonomously by
the agent (3) with a maximum of ten per invocation and can be utilized
either in parallel or in a sequential chain (4). This way, the agent can
generate relevant patient information on demand and utilize this
knowledge to query relevant documents within its database (4). This
enables it to generate a highly specific and patient-focused response
that integrates the initial clinical data with newly acquired insights,
all while being substantiated by authoritative medical documentation
(5).}
  \label{fig:Figure 1}
\end{figure}

\begin{figure}[htbp]
  \centering
  \includegraphics[width=\linewidth]{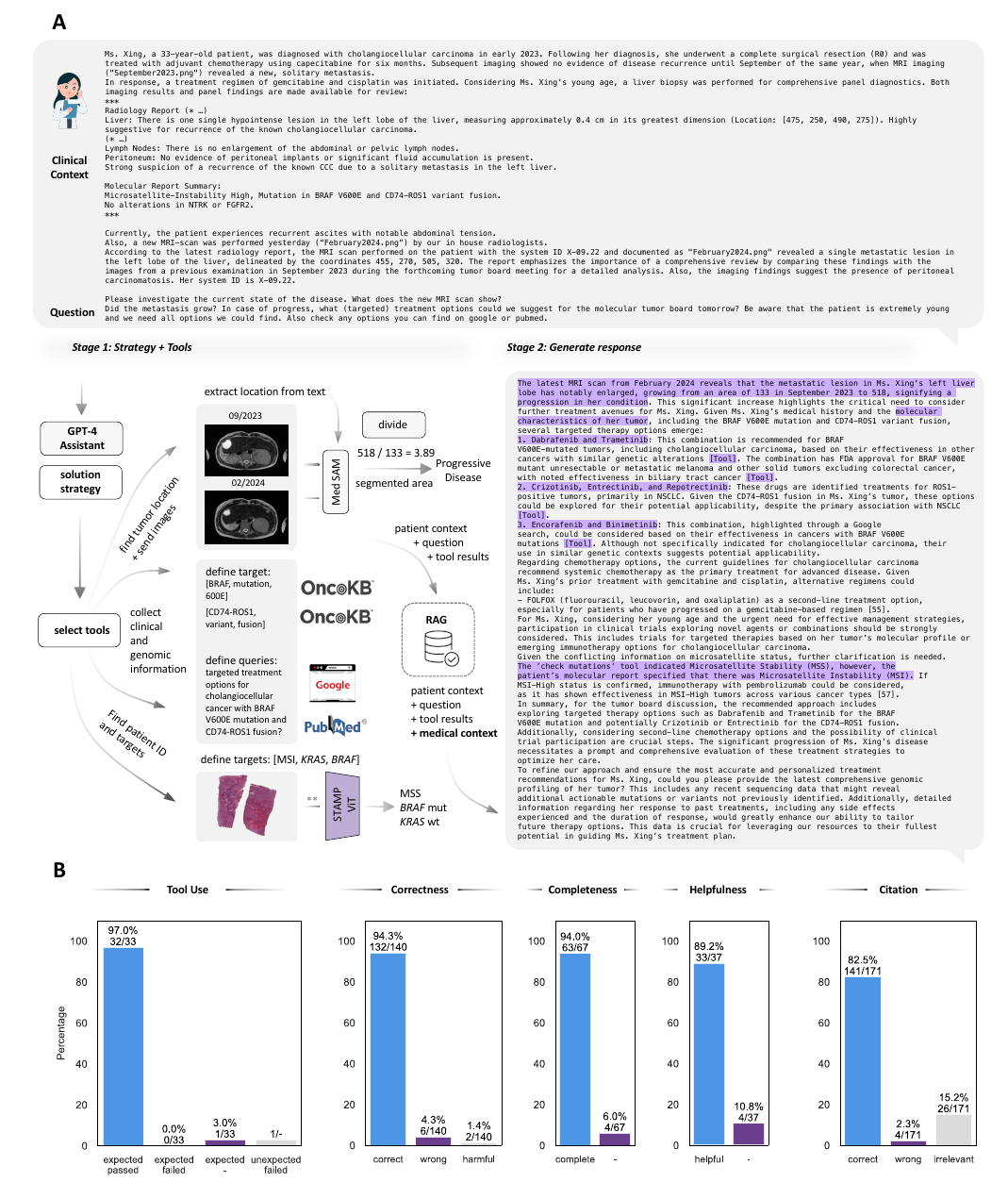}
  \caption{Details and efficacy of the agent's pipeline in
patient case evaluation.\\
We highlight the full agent's pipeline for patient X, showcasing the
complete input process and the collection of tools deployed by the
agent. In the initial 'tools' phase, the model identifies tumor
}
  \label{fig:Figure2}
\end{figure}
\clearpage

\noindent localization from patient data (* abridged for readability. The complete
text is available in Supplementary Table 3.) and utilizes MedSAM for the
generation of segmentation masks. Measuring the area of the segmented
region enables the calculation of tumor progression over time as the
end model calculates an increase by a factor of 3.89. The agent also
references the OncoKB database for mutation information from the
patient's context (BRAF V600E and CD74-ROS1) and performs literature
searches via PubMed and Google. For histological modeling, we must note
here that we have streamlined the processing: The original STAMP
pipeline consists of two steps, where the first is the timely and
computationally intensive calculation of feature vectors, which we have
performed beforehand for convenience. The second step is performed by
the agent by selecting targets of interest and the location of the
patient's data and executing the respective vision transformer (**). The
subsequent phase involves data retrieval via RAG and the production of
the final response.
Panel B shows the results from a manual evaluation conducted by a panel
of four medical experts. The metric \textquotesingle Tool
Use\textquotesingle{} reflects the ratio of tools employed by the agent
versus the number anticipated (32/33), whereas
\textquotesingle Completeness\textquotesingle{} (63/67) measures the
proportion of experts\textquotesingle{} expected answers, as
predetermined by keywords, that the model accurately identifies or
proposes. \textquotesingle Helpfulness\textquotesingle{} quantifies the
ratio of sub questions the model actually answers out of all questions
or instructions given by the user (33/37).
\textquotesingle Correctness\textquotesingle{} (131/140),
\textquotesingle Wrongness\textquotesingle{} (6/140), and
\textquotesingle Harmfulness\textquotesingle{} (3/140) represent the
respective ratios of accurate, incorrect (yet not detrimental), and
damaging responses relative to the total number of responses. Here, a
response is constituted by individual paragraphs per answer. Lastly, we
measure whether a provided reference is correct (141/171), irrelevant
(11/171, the reference's content does not mirror the model's statement)
or wrong (3/171). Results shown here are obtained from the majority vote
across all observers, with selecting the in cases of a tie.

\clearpage
\section{\texorpdfstring{\textbf{Supplementary
Material}}{Supplementary Material}}\label{supplementary-material}

\begin{figure}[H]
  \centering
  \includegraphics[width=0.8\linewidth]{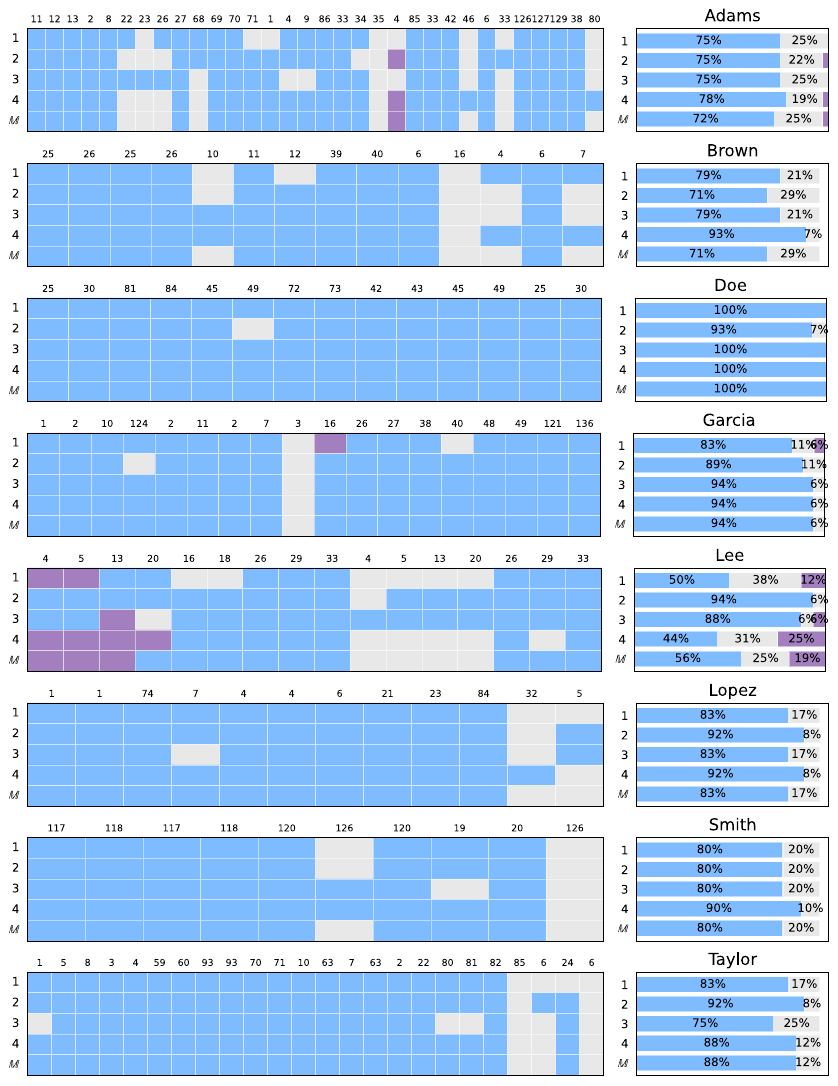}
  \vspace{-3em}
  \caption*{}
  \label{fig:SupplFig1}
\end{figure}

\begin{figure}[H]
  \centering
  \includegraphics[width=0.8\linewidth]{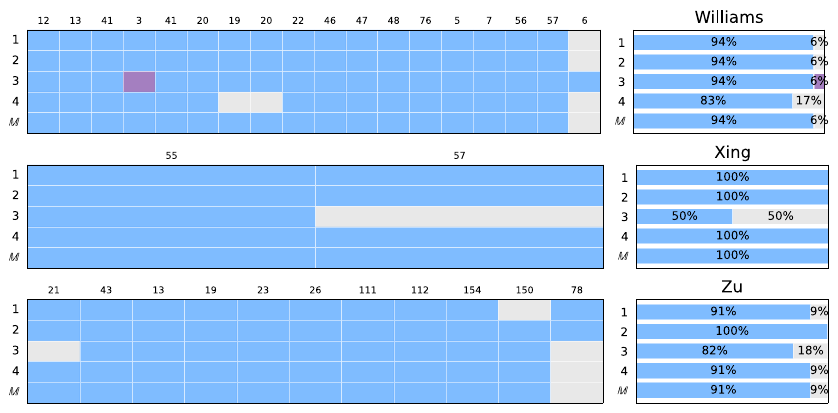}
  \vspace{-3em}
  \caption*{}
  \label{fig:SupplFig1cont}
\end{figure}

\textbf{Supplementary Figure 1. Citation evaluation by human observers
per patient.}
The figure presents an assessment of the citations evaluated for each
patient, illustrated as individual subplots, with each row representing
a distinct reviewer. Citations deemed accurate are marked in blue,
irrelevant citations in gray, and incorrect citations are highlighted in
purple. The notation \textquotesingle M\textquotesingle{} signifies the
consensus achieved through the majority vote of the four observers. In
instances of equal votes, the more conservative rating is adopted,
adhering to a predefined hierarchy of evaluation: correct, irrelevant,
and incorrect.

\begin{longtable}{@{}l|cc|cc|cc|c@{}}
\caption*{{\Large Supplementary Table 1: Tool Use}}\\
\toprule
\multicolumn{1}{c|}{} & \multicolumn{2}{c|}{\textbf{Radiology}} & \multicolumn{2}{c|}{\textbf{Genetics}} & \multicolumn{2}{c|}{\textbf{Web Search}} & \multicolumn{1}{c}{\textbf{-}} \\

No & GPT-4V & MedSAM & Kras/Braf/MSI & OncoKB & Google & PubMed & Calculator \\
\midrule
A & \cmark & - & \cmark & - & - & - & - \\
\midrule
B & \cmark & - & \cmark & - & - & - & - \\
\midrule
D & \cmark & - & \cmark & \cmark & \cmark & - & - \\
\midrule
G & - & \cmark \cmark & - & - & - & - & \cmark \\
\midrule
Le & \cmark & - & - & - & \cmark & \qmark & - \\
\midrule
Lo & \cmark & - & \cmark & - & - & - & - \\
\midrule
S & \cmark & - & \cmark & - & - & - & - \\
\midrule
T & \cmark & - & - & - & - & - & - \\
\midrule
W & - & \cmark \cmark & \cmark & \cmark & - & - & \cmark \\
\midrule
X & - & \cmark \cmark & \cmark & \cmark \cmark & \cmark & \cmark & \cmark \\
\midrule
Z & \cmark & - & \xmark & - & - & - & - \\

\bottomrule
\end{longtable}

\setcounter{footnote}{0}
  
\textbf{Supplementary Table 1. Distribution of required and actually
invoked tools per patient case.}
Table 1 shows the alignment between anticipated and actual tool
utilization by the agent. A checkmark (\cmark) signifies a successful and
expected tool application, whereas a cross (\xmark) denotes an error during
execution. Notably, in the isolated instance involving patient Z
(patient names are abbreviated by surname start letters in this table),
the attempt to use the \emph{KRAS}, \emph{BRAF} and MSI vision
transformer models was compromised due to an erroneous application
of the patient\textquotesingle s name as an invalid identifier, coupled
with the absence of requisite images. A dash (-) indicates the
non-necessity for tool deployment in a specific context for a given
task, while a star (\texttt{*}) marks an anticipated yet unexecuted tool
application. The repetition of symbols quantitatively reflects the
frequency of tool engagement.

\customCaption{Supplementary Table 2: Pseudo-Algorithm}

\begin{algorithm}[H]

\caption{%
    Response generation algorithm.
    This algorithm generates a the model's final response after the agent has invoked the necessary tools in response to the question.
    It involves several calls to the LLM which are denoted as functions of the form \texttt{Generate*($\cdot$)}; these functions invoke the LLM with a templated prompt into which the input arguments are inserted. CoT = Chain-of-Thought, P = Predict (akin to dspy.Predict).
}
\begin{algorithmic}[1]
    \Require
        \Statex Patient context $C_\textit{patient}$ \Comment{Contains clinical information about the patient}
        \Statex Question $Q$ \Comment{Question(s) or instruction(s) related to the patient}
        \Statex Agent tools $A_\textit{tools}$ \Comment{List of all available tools}
        \Statex Tool outputs $T_\textit{out}$ \Comment{Free-text summary of the results of the tools used by the agent}
        \Statex Documents $D$ \Comment{Collection of medical documents (guidelines, textbooks, etc.)}
    \Ensure Comprehensive and accurate response $R$
    \State $C_Q \gets \text{empty list}$ \Comment{Question context (will contain passages relevant to $Q$)}
    \State $\textit{Subqueries} \gets \texttt{GenerateSubqueries}_{\text{CoT}}(C, Q, T_{\text{out}})$ \Comment{Generate 10-14 sub-questions for RAG}
    \ForEach{subquery $q$ \textbf{in} $\textit{Subqueries}$}
        \State $P \gets \text{Retrieve}_n(D, q)$ \Comment{Retrieve $n=40$ passages relevant to the subquery $q$}
        \State $P \gets \text{Rerank}(P, q)$ \Comment{Re-rank the retrieved passages based on relevance to $q$}
        \State $P \gets \text{Top}_k(P)$ \Comment{Keep only the top $k=10$ passages}
        \State $C_Q \gets C_Q \cup P$ \Comment{Add the passages to the question context}
    \EndFor
    \State $C_Q \gets \text{Deduplicate}(C_Q)$ \Comment{Remove duplicate passages}
    \For{$i$ \textbf{in} $1, \dots, |C_Q|$}  \Comment{Add numbered source information to each passage}
        \State $C_Q[i] \gets \text{Concat}(\text{"Source"}, i, C_Q[i])$
    \EndFor
    \State $\textit{Strategy} \gets \texttt{GenerateAnswerStrategy}_{\text{CoT}}(C, Q, A_\textit{tools}, T_\textit{out}, C_Q)$
    \State $\textit{CitedResponse} \gets \texttt{GenerateCitedResponse}_{\text{P}}(C, Q, A_\textit{tools}, T_\textit{out}, C_Q, \textit{Strategy})$
    \State $\textit{Suggestions} \gets \texttt{GenerateSuggestions}_{\text{CoT}}(\textit{CitedResponse}, A_\textit{tools}, T_\textit{out})$
    \State $R \gets \text{Concat}(\textit{CitedResponse}, \textit{Suggestions})$ \Comment{Final output}
    \State \Return $R$
\end{algorithmic}
\end{algorithm}

\clearpage
\textbf{Supplementary Table 2. Pseudocode for RAG and response
generation.}
The algorithm involves several calls to a language model (GPT-4) to
generate its final response. Therefore it requires the patient's
clinical case narrative, the question submitted by the user, an array of
agent tools, free-text summaries of the tool's outputs and a collection
of context-relevant medical documents that will be retrieved through
RAG. For each query, the algorithm formulates a set of subsidiary questions,
conducts a search for \emph{n} relevant text passages, re-ranks them and
keeps only the top \emph{k} passages while removing duplicates. Next, it
uses several Chain-of-Thought steps to generate a detailed and
structured strategy on how to answer the questions and identifies
potential areas where additional input from the user could be
beneficial. Finally, it executes the strategy, incorporating the
utilization of the documents retrieved earlier in the process.
\clearpage

\setlength{\fboxsep}{1pt} 

\begin{longtable}{@{} l >{\raggedright\arraybackslash}p{14cm} @{}}
\caption*{{\Large Supplementary Table 3: Model Outputs}}\\
\\ 

\toprule 
\textbf{No.} & \textbf{Patient Case Description} \\
\midrule 
\endfirsthead 

\toprule
\textbf{No.} & \textbf{Patient Case Description} \\
\midrule
\endhead 

A & Mrs. Adams (DOB: 03/05/1969, ID: A-CC-38), a 55-year-old female, has been battling left-sided colorectal cancer for an extended period. Four years after her initial diagnosis, in November 2023, a relapse was detected, characterized by multiple metastases to the liver, as evidenced by the imaging study dated 11/11/2023 (CTA\_11\_11\_2023.png). Subsequently, she commenced a chemotherapy regimen comprising FOLFOX and bevacizumab at an external hospital in Berlin. The latest imaging, captured on 02/24/2024 (CTA\_02\_24\_2024.png), is stored in the patient's folder within the Images path of the system. To date, there are no additional metastatic sites identified beyond the liver. Mrs. Adams has generally tolerated the chemotherapy well. However, she has recently reported an increase in blood pressure, with readings consistently in the range of 170-190 mmHg. Apart from elevated blood pressure, she has not experienced any other symptomatic manifestations of her condition.
\\
\\
& \textbf{Doctor:} How did the disease develop? What are the next steps in her cancer treatment plan? \\
\\

& \textbf{Appendix\footnotemark:} \\

& \begin{minipage}{4cm}
    \centering
    CT-A, 11/11/2023 \\
    \includegraphics[width=\linewidth]{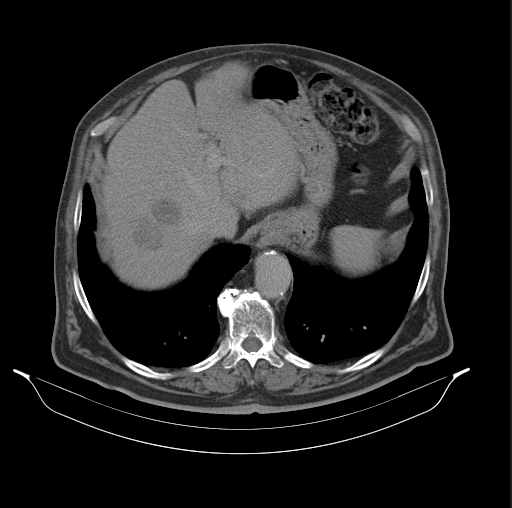} 
  \end{minipage}%
  \hspace{1mm} 
  \begin{minipage}{4cm}
    \centering
    CT-A, 02/24/2024 \\
    \includegraphics[width=\linewidth]{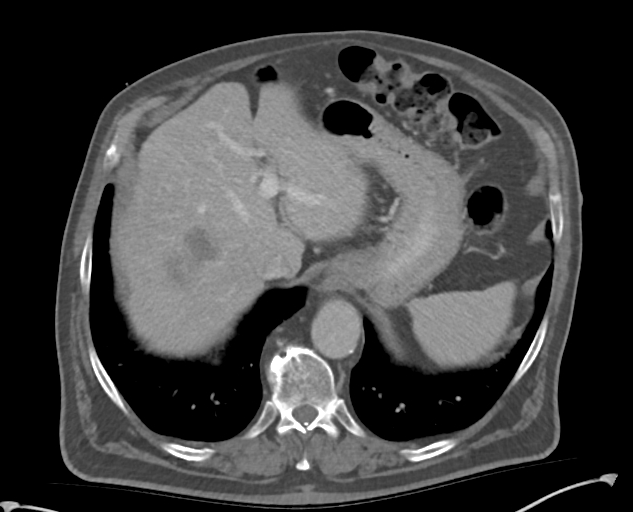} 
  \end{minipage} \\
& \begin{minipage}{6cm}
    \centering
    Histology \\
    \includegraphics[width=\linewidth]{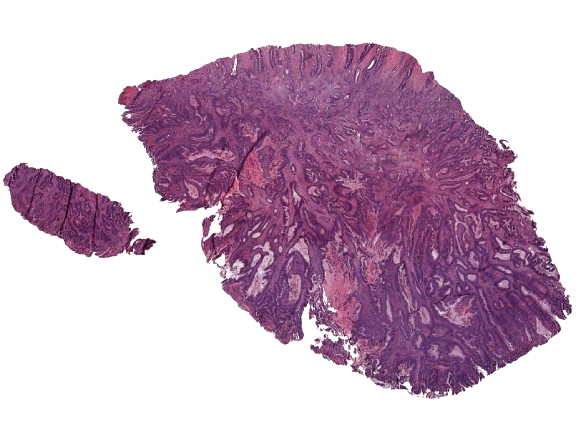} 
  \end{minipage} \\
\\
\\
& \textbf{Model Response\footnotemark:} \\

& Given Mrs. Adams' \colorbox{lightblue}{stable liver metastases as evidenced by the comparison} \colorbox{lightblue}{of CTA\_11\_11\_2023.png and CTA\_02\_24\_2024.png [Tool]}, alongside her genetic profile indicating \colorbox{lightblue}{MSS, and wild-type status for both BRAF and KRAS [Tool]}, her disease appears to be in a controlled state under the current regimen of FOLFOX and bevacizumab. The stability of liver lesions suggests that the chemotherapy, in conjunction with bevacizumab, is effective in managing her metastatic colorectal cancer (mCRC) \colorbox{lightblue}{[Tool]}. \\

& The development of hypertension, a known side effect of bevacizumab [11][12][13], necessitates a multidisciplinary approach to manage this condition without compromising her cancer treatment efficacy. Given the \colorbox{lightblue}{stable disease and the absence of BRAF and} \colorbox{lightblue}{KRAS mutations [Tool]}, Mrs. Adams remains a candidate for continued anti-EGFR therapy, considering her left-sided mCRC, which is typically more responsive to such treatments [2][8]. \\
\\
& \textbf{Next Steps in Treatment Plan:} \\

& \textbf{1. Management of Hypertension:} Immediate consultation with a cardiologist is recommended to manage her newly developed hypertension, potentially related to bevacizumab [22][23][26][27]. Adjustments to her antihypertensive therapy may be necessary, with continuous monitoring of blood pressure to prevent cardiovascular complications [68][69][70][71]. \\

& \textbf{2. Ongoing Cancer Treatment:} Given the stable disease, continuation of the current regimen is advisable. However, the patient's tolerance, especially concerning the development of hypertension, should be closely monitored [1][4]. If hypertension cannot be controlled, a dose adjustment or a switch to an alternative antiangiogenic agent with a \colorbox{orange}{potentially different side effect profile, such as aflibercept or ramucirumab,} \colorbox{orange}{could be considered} [9][86]. \\

& \textbf{3. Re-evaluation for Surgical Intervention:} Despite the stable metastatic disease, re-evaluation for the possibility of liver metastasectomy or ablation should be considered, as these could offer a chance for prolonged survival or even cure in selected cases [33][34][35]. \\

& \textbf{4. Maintenance Therapy:} If surgery is not feasible and once a maximum response to the current regimen is achieved, transitioning to maintenance therapy with a fluoropyrimidine (e.g., capecitabine) with or without bevacizumab could be considered to reduce toxicity while maintaining disease control [4][85]. \\

& \textbf{5. Genetic and Molecular Marker Re-evaluation:} Although initial genetic predictions indicate MSS, and wild-type BRAF and KRAS, a broader panel re-evaluation might uncover other actionable mutations or markers that could open up additional targeted therapy options [33][42][46]. \\

& \textbf{6. Clinical Trials and Novel Therapies:} Exploration of clinical trials focusing on novel therapies or combinations that could be beneficial for her specific genetic and molecular profile is recommended [6][33]. \\

& \textbf{7. Palliative Care Integration:} Early integration of palliative care to manage symptoms, improve quality of life, and support Mrs. Adams and her family through treatment decisions is crucial [126][127][129]. \\

& \textbf{8. Regular Monitoring:} Continuous assessment of her disease with imaging studies and tumor markers is essential to evaluate the response to treatment and adjust the treatment plan as necessary [38][80]. \\
& This comprehensive approach, tailored to Mrs. Adams' specific medical history, current condition, and potential treatment pathways, aims to optimize her quality of life while effectively managing her mCRC. Regular reassessment of her condition and treatment tolerance will guide any necessary adjustments to her treatment plan. \\
& To further refine Mrs. Adams' treatment plan, it would be helpful to know the specifics of her blood pressure readings over time, to assess the severity and progression of her hypertension. Utilizing my array of resources, I can further assist by identifying optimal blood pressure management strategies that align with her ongoing cancer treatment. Additionally, detailed information regarding her response to previous treatments and any side effects experienced would be invaluable. This would enable a more personalized approach to considering the integration of anti-EGFR therapy, ensuring it complements her current regimen while minimizing potential adverse effects. \\

\bottomrule 


B & Mrs. Brown, identified in our system as X11-01A, is a 49-year-old female patient who has been navigating her journey with metastatic colorectal cancer since 2019. Her treatment history includes chemotherapy followed by surgical interventions, specifically a total mesorectal excision and a right-sided hemi-hepatectomy. In the previous year, a follow-up CT scan revealed the presence of a single metastasis in the liver, prompting the initiation of a chemotherapy regimen that combined FOLFOX with Panitumumab. Currently, Mrs. Brown returns with a new CT scan dated February 10, 2024, ordered by her oncologist. Additionally, she provides microscopic images from a recent biopsy of the tumor. All relevant imaging data have been uploaded to our system.
\\
\\
& \textbf{Doctor:} How do we best proceed with the oncologic treatment of the patient? Is a continuation of chemotherapy plus Panitumumab recommended? What other options do we have? 
\\
\\
& \textbf{Appendix:} \\

& \begin{minipage}{4cm}
    \centering
    CT-A, 11/01/2023 \\
    \includegraphics[width=\linewidth]{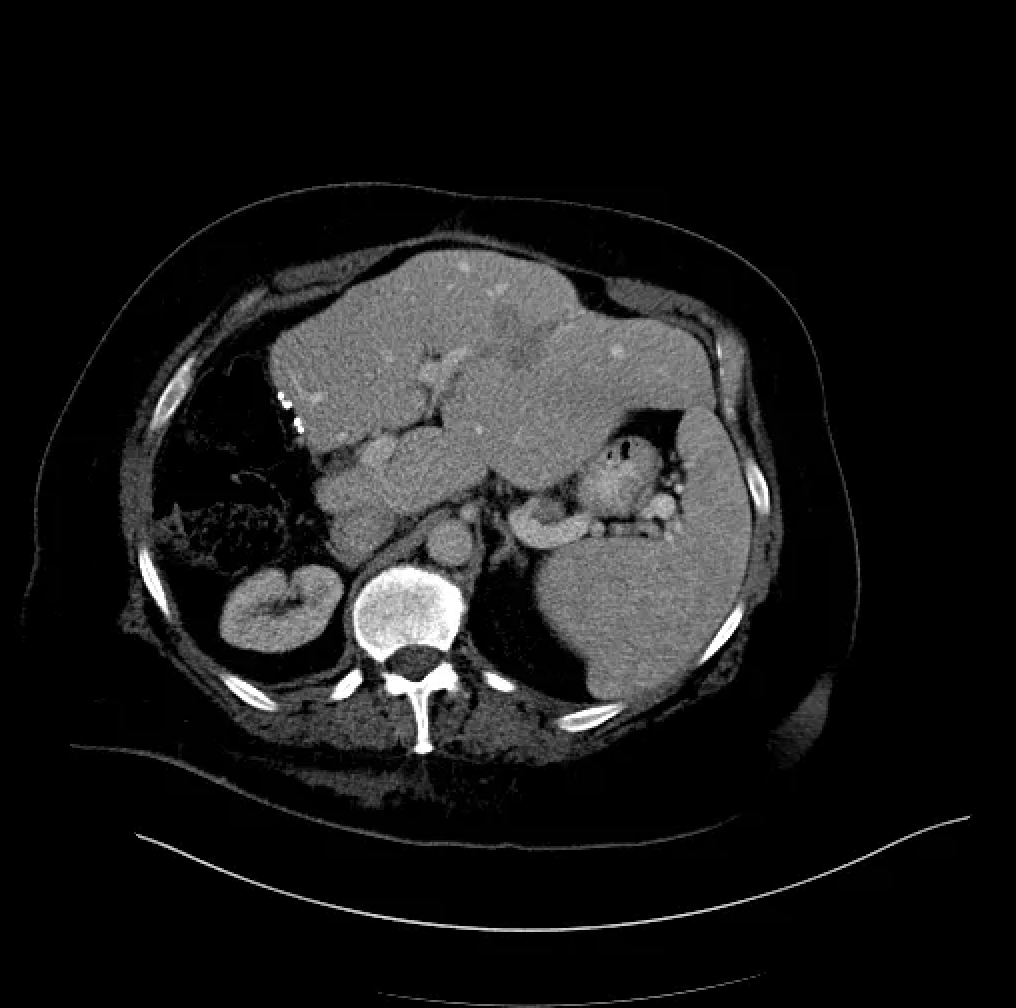} 
  \end{minipage}%
  \hspace{1mm} 
  \begin{minipage}{4cm}
    \centering
    CT-A, 02/10/2024 \\
    \includegraphics[width=\linewidth]{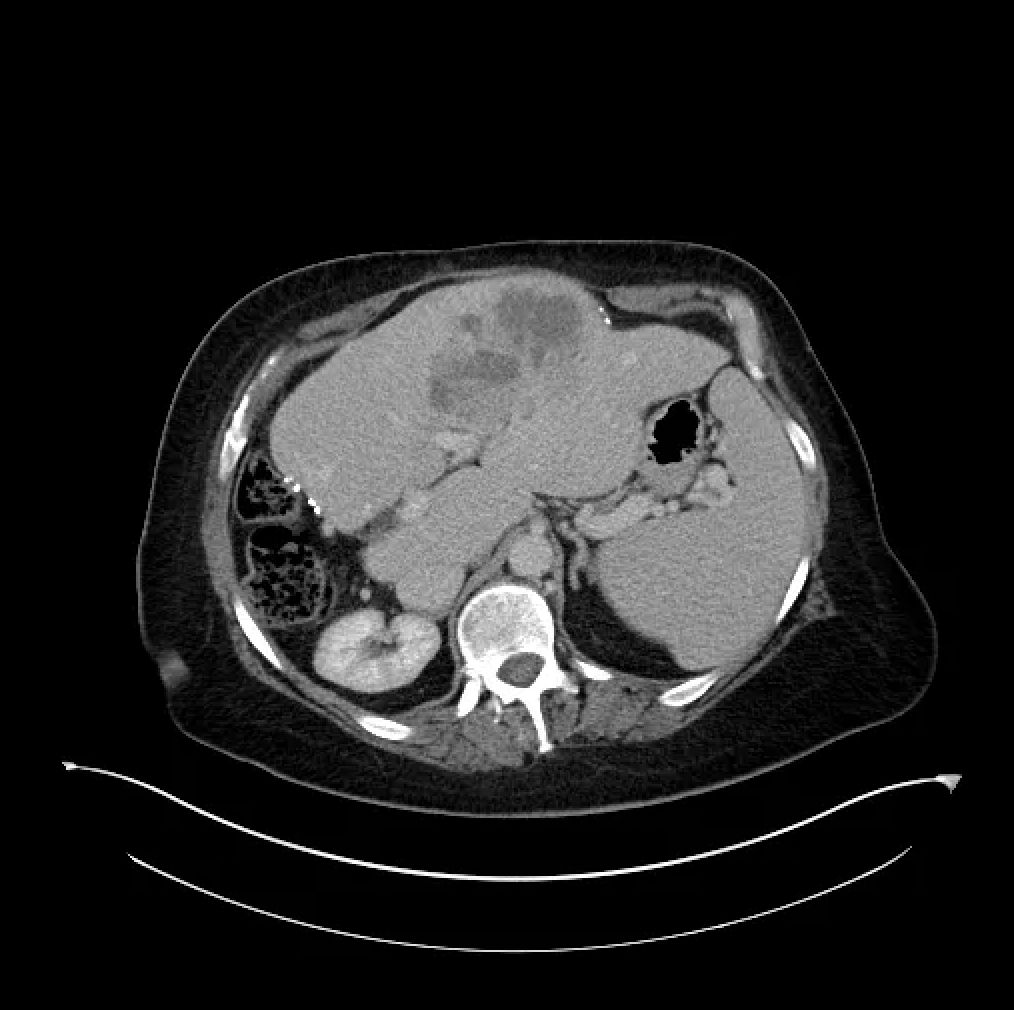} 
  \end{minipage} \\
& \begin{minipage}{6cm}
    \centering
    Histology \\
    \includegraphics[width=\linewidth]{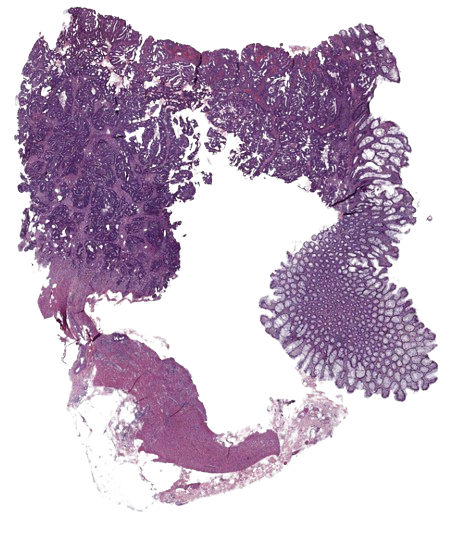} 
  \end{minipage} \\

& \textbf{Model Response:} \\

& Given Mrs. Brown's current clinical scenario, including the \colorbox{lightblue}{recent CT findings and genetic} \colorbox{lightblue}{profile of her tumor}, a strategic re-evaluation of her treatment plan is warranted. The \colorbox{lightblue}{presence of a new liver lesion} suggests disease progression, and the \colorbox{lightblue}{genetic analysis} \colorbox{lightblue}{revealing a KRAS mutation} significantly impacts the choice of targeted therapy. Specifically, the efficacy of Panitumumab, an EGFR inhibitor, is compromised in the context of KRAS mutations [25], [26]. Therefore, continuing chemotherapy plus Panitumumab is not recommended based on the current evidence. \\

& \textbf{1. Discontinue Panitumumab:} Given the KRAS mutation, discontinuing Panitumumab is advised. KRAS mutations are known to confer resistance to EGFR inhibitors, rendering treatments like Panitumumab ineffective [25], [26]. \\

& \textbf{2. Adjust Chemotherapy Regimen:} Considering the disease progression, adjusting the chemotherapy regimen might be necessary. Options include switching to a different chemotherapy backbone such as FOLFIRI (folinic acid, fluorouracil, and irinotecan) if not previously used, or exploring the addition of bevacizumab, an anti-VEGF therapy, which is not affected by KRAS mutation status and could offer a benefit in this setting [10], [11], [12]. \\

& \textbf{3. Explore Additional Targeted Therapies:} Given the KRAS mutation, targeted therapies that are effective against KRAS-mutated tumors should be considered. While direct KRAS inhibitors like sotorasib are currently approved for lung cancer with KRAS G12C mutations, ongoing trials are exploring their efficacy in colorectal cancer. They may offer potential benefits, although their applicability depends on the specific KRAS mutation present. Participation in a clinical trial targeting KRAS mutations could be an option [39], [40]. \\

& \textbf{4. Consider Local Treatments for Liver Metastasis:} Depending on the size, location, and number of liver metastases, local treatments such as radiofrequency ablation (RFA) or stereotactic body radiation therapy (SBRT) could be considered to manage the liver disease [6], [16]. \\

& \textbf{5. Clinical Trials:} \colorbox{red}{Enrolling Mrs. Brown in clinical trials} focusing on novel therapies for KRAS-mutated mCRC or new combinations of existing drugs could provide access to cutting-edge treatments and potentially improve outcomes [4]. \\

& \textbf{6. Supportive Care and Monitoring:} Regardless of the treatment adjustments, ensuring supportive care to manage symptoms and side effects, along with regular monitoring of disease status through imaging and biomarkers, is crucial [6]. \\

& \textbf{7. Patient Education and Involvement:} Discussing the rationale behind changing the treatment plan, potential benefits and side effects of the new regimen, and the importance of close monitoring with Mrs. Brown is essential. Her preferences and concerns should be taken into account in shared decision-making [7]. \\
& In summary, the treatment plan for Mrs. Brown should involve discontinuing Panitumumab due to the KRAS mutation, considering an alternative chemotherapy regimen possibly combined with bevacizumab, exploring eligibility for clinical trials, and evaluating local treatment options for liver metastasis. Regular follow-up and supportive care remain integral components of her care plan. \\
& To refine our approach based on Mrs. Brown's unique genetic profile and the progression of her disease, it would be beneficial to gather more detailed information on any additional genetic alterations that have not yet been explored. Utilizing my array of resources, I can further assist by identifying the most promising treatment avenues tailored to her specific genetic landscape. This includes pinpointing novel or underutilized targeted therapies that may bypass the resistance mechanisms posed by the KRAS mutation. Additionally, understanding the full scope of her disease through comprehensive genetic profiling could unveil new opportunities for clinical trial participation, potentially offering access to innovative treatments not yet available in standard care. \\

\bottomrule 

D & Mr. Doe, a 53-year-old male patient with the identification X22-70C, has presented to the oncology department with a recent diagnosis of colorectal cancer. He has provided CT images, which have been stored in the 'Imaging' directory under a subfolder bearing his family name. Unfortunately, a corresponding radiology report is not available at this time.
Additionally, Mr. Doe mentions a biopsy conducted last week, the slides of which have been uploaded by the pathology department into the 'Histology' folder. His medical history and physical examination have not revealed any cancer-specific findings.
During his consultation, Mr. Doe repeatedly expresses his refusal to undergo chemotherapy. He also mentions undergoing molecular testing at another facility, which, to the best of his recollection, identified a BRAF V600E mutation.
\\
\\
& \textbf{Doctor:} Is the patient rather suitable for surgery or systemic therapy? Are there options for targeted therapies? Check available drugs for the known mutation in databases and online. Are there other mutations? \\
\\

& \textbf{Appendix:} \\
& \begin{minipage}{4cm}
    \centering
    CT Liver \\
    \includegraphics[width=\linewidth]{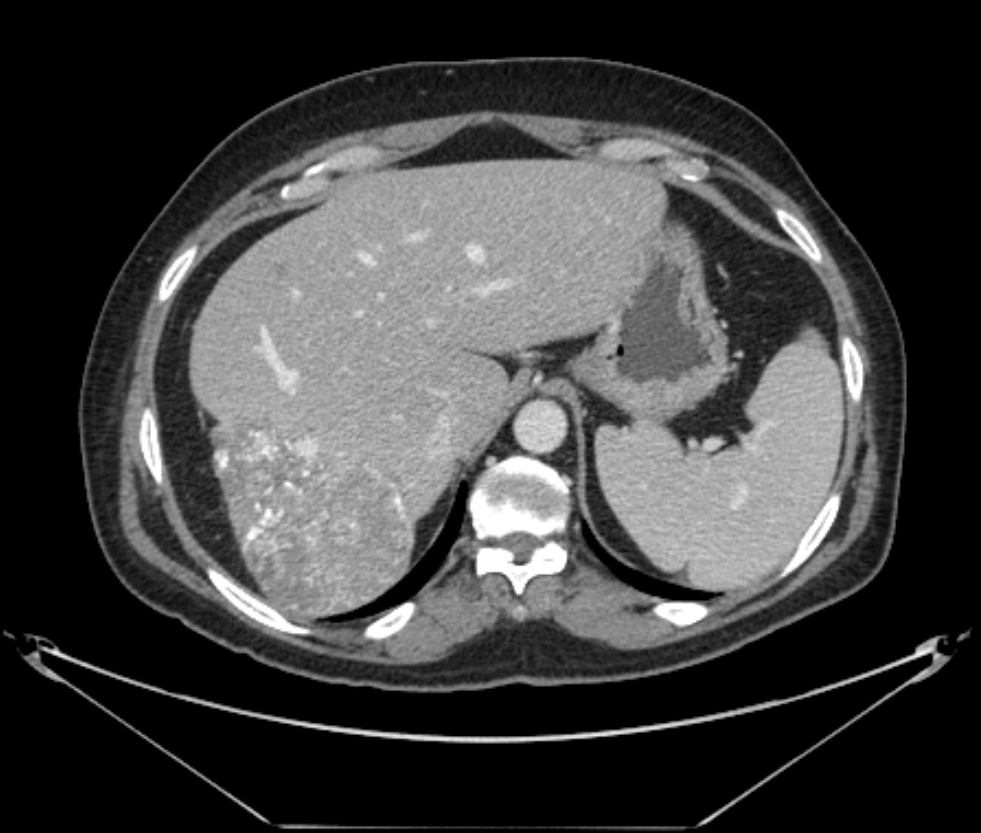} 
  \end{minipage}%
  \hspace{1mm} 
  \begin{minipage}{3.5cm}
    \centering
    CT Lung \\
    \includegraphics[width=\linewidth]{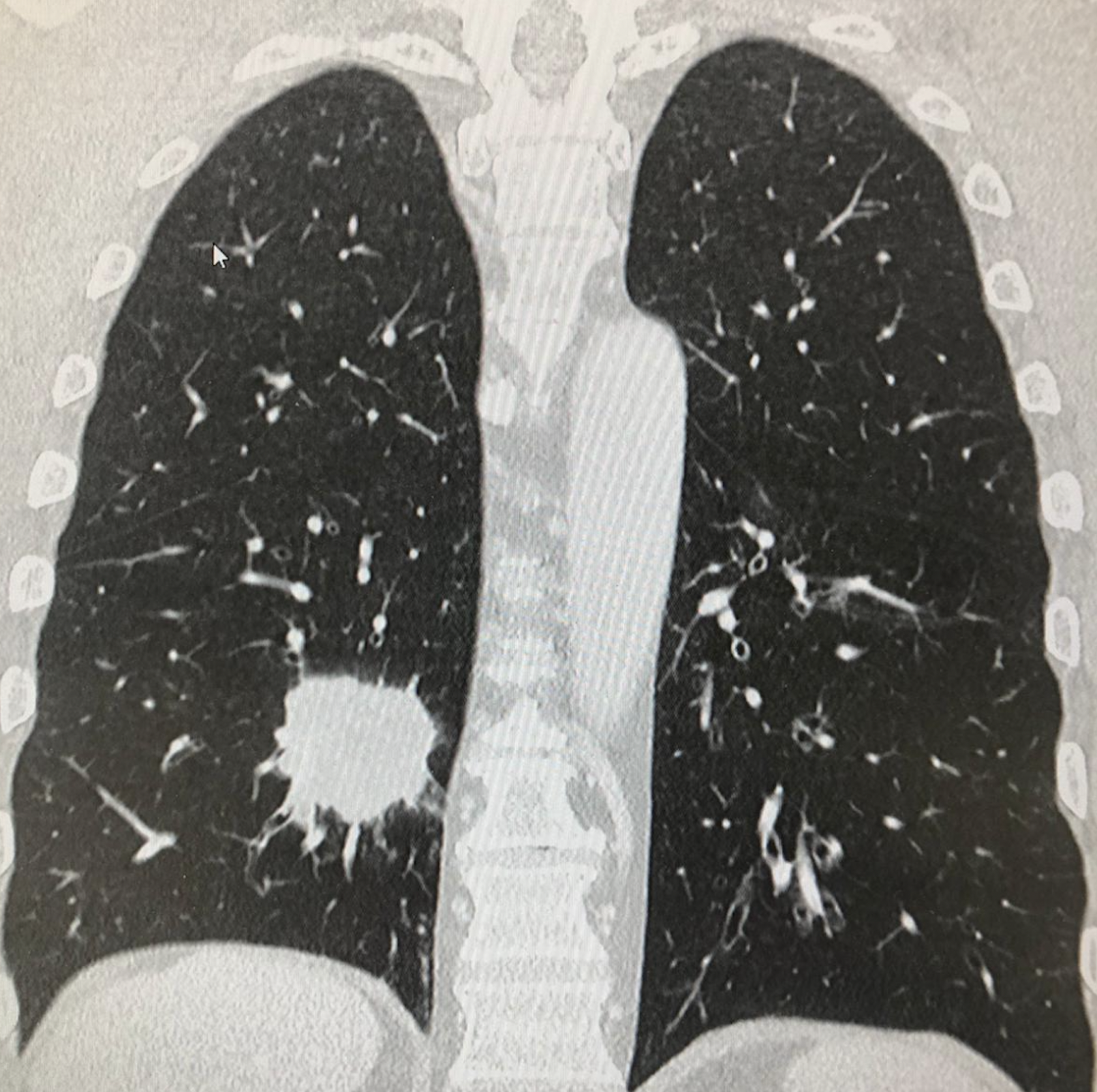} 
  \end{minipage} \\
& \begin{minipage}{6cm}
    \centering
    Histology \\
    \includegraphics[width=\linewidth]{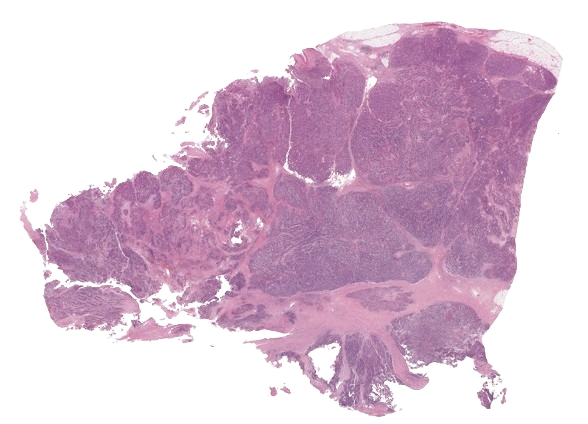} 
  \end{minipage} \\

\\
& \textbf{Model Response:} \\
& Given Mr. Doe's advanced metastatic colorectal cancer with \colorbox{lightblue}{evidence of metastases in both the lungs and liver}, as indicated by the \colorbox{lightblue}{\textbf{gen\_radiology\_report}} tool analysis of \colorbox{lightblue}{\textbf{CT\_Lung.png} and \textbf{CT\_Liver.png} [Tool]}, he is more suitable for systemic therapy rather than surgery. The \colorbox{red}{presence of multiple well} \colorbox{red}{-defined nodules in both lung fields} and multiple hypodense lesions in both lobes of the liver strongly suggests a widespread metastatic disease, making surgical intervention less feasible and unlikely to provide curative outcomes \colorbox{lightblue}{[Tool]}.\\

& The \colorbox{lightblue}{\textbf{check\_mutations} tool's prediction of MSI-High status} with a probability of 0.95 \colorbox{lightblue}{[Tool]} opens up possibilities for immunotherapy, particularly given Mr. Doe's refusal to undergo chemotherapy. Pembrolizumab, an immune checkpoint inhibitor, is recommended for patients with MSI-High or dMMR metastatic colorectal cancer (mCRC) as a first-line treatment [25][30]. This aligns with Mr. Doe's treatment preferences and the current medical guidelines suggesting pembrolizumab for MSI-High mCRC [81][84].\\

& The discrepancy regarding the BRAF V600E mutation status, with Mr. Doe's recollection of a positive result and the \colorbox{lightblue}{\textbf{check\_mutations} tool predicting wild-type BRAF} with a probability of 0.48 \colorbox{lightblue}{[Tool]}, necessitates a confirmatory molecular test. If the BRAF V600E mutation is confirmed, targeted therapy with a combination of encorafenib and cetuximab is recommended in the second line, based on the BEACON study results and FDA approval for mCRC with BRAF V600E mutation after prior therapy [45][49]. This combination has shown efficacy in increasing overall survival compared with standard chemotherapy [72][73].\\

& Furthermore, the \colorbox{lightblue}{slight likelihood of a KRAS mutation, as indicated by the } \colorbox{lightblue}{\textbf{check\_mutations} tool} with a probability of 0.52 \colorbox{lightblue}{[Tool]}, suggests that anti-EGFR therapies such as cetuximab or panitumumab, typically used in RAS wild-type mCRC, may not be as effective [42][43]. This further emphasizes the importance of confirming the BRAF mutation status, as the presence of a KRAS mutation limits the use of certain targeted therapies.\\

& Given the advanced nature of Mr. Doe's disease and his preference against chemotherapy, exploring clinical trials and newer treatment modalities is also a prudent step. This could include trials focusing on novel immunotherapies, targeted therapies based on other actionable mutations, or combination therapies that have not yet been standardly approved but show promise in early-phase trials.\\

& In summary, Mr. Doe's treatment strategy should involve:\\
& 1. Confirming the BRAF mutation status through a reliable molecular testing method. If BRAF V600E mutation is confirmed, proceed with targeted therapy using encorafenib plus cetuximab in future lines of treatment [45][49].\\
& 2. Considering immunotherapy with pembrolizumab due to the MSI-High status, which has shown efficacy in MSI-High or dMMR mCRC as a first-line treatment [25][30].\\
& 3. Exploring clinical trials and newer treatment modalities, given the advanced nature of Mr. Doe's disease and his preference against chemotherapy.\\
& 4. Involving a multidisciplinary team including medical oncologists, genetic counselors, and possibly palliative care specialists to address Mr. Doe's treatment preferences, manage symptoms, and ensure the best possible quality of life.\\
& Further investigations should include a repeat or confirmatory genetic test to resolve the discrepancy regarding the BRAF mutation status and comprehensive molecular profiling to identify any other actionable mutations or biomarkers that could guide therapy selection. This approach ensures a comprehensive and patient-centered care plan that aligns with the latest clinical evidence and Mr. Doe's treatment preferences.\\
& To further refine our analysis and recommendations, could you please provide the coordinates of the nodules and lesions highlighted in the CT scans? This information is crucial for a detailed segmentation analysis, which will help in accurately assessing the extent of metastatic spread. Additionally, if you could supply the measurements of these lesions over time, we can perform a progression analysis to better understand the disease's trajectory. Lastly, sharing specific search terms or questions you have regarding colorectal cancer, particularly those focusing on KRAS mutations and targeted therapy options, would enable a targeted literature review, ensuring we gather the most relevant and up-to-date information to guide treatment planning.\\

\bottomrule 


G & Mr. Garcia, a 61-year-old male, has been under treatment for metastatic colorectal cancer. He previously underwent six cycles of FOLFOX therapy at an external hospital. Mr. Garcia has provided a CT scan of the abdomen from November 8, 2023, taken before the initiation of his treatment, now uploaded under the filename '08\_11\_23.png.' We also have an corresponding old radiology report: \\

& \*****\\
& \textbf{Radiology Report from November, 8 2023:}\\
& Abdominal MRI-scan.\\
& Clinical Information: Biospy-confirmed colorctal cancer with potential liver metastases according to ultra-sound. \\
& Technique: Multi-phase imaging of the abdomen with intravenous contrast.\\
& Findings:\\
& Primary Tumor: The primary colorectal tumor is identified within the upper 1/3 of the rectum without evidence of intestinal occlusion\\
& Liver: There are multiple hepatic lesions consistent with metastases. We highlight three target lesions: The largest lesion is located in position 190, 195, 225, 225. Two more lesions cand be found at [230, 100, 260, 130] and [140, 245, 170, 270]. No evidence of biliary dilation.\\
& Kidneys, Adrenals, Pancreas, Spleen: No abnormalities detected. Organs appear normal in size, shape, and attenuation without evidence of lesions or masses.\\
& Lymph Nodes: No enlarged abdominal or pelvic lymph nodes are identified.\\
& Vasculature: The abdominal aorta and its major branches are patent without evidence of aneurysm or significant stenosis.\\
& Bowel and Mesentery: No evidence of bowel obstruction, focal bowel wall thickening, or mesenteric masses.\\
& Peritoneum: No ascites or peritoneal nodules are evident.\\
& Other Findings: There are no other significant abnormalities detected in the abdomen.\\
& \*****\\

& A recent control imaging conducted by our in-house radiologists is saved as '02\_10\_24.png'. All related imaging data are accessible on the server under 'Imaging/{family\_name}.' The preliminary report states "... multiple hepatic lesions consistent with metastases. The largest lesion is located in position: [120, 260, 165, 305], more are found in positions [240, 220, 280, 255], [195, 185, 235, 220] and 145, 235, 180, 275. No comparison to previous imaging available." \\
\\
\\
& \textbf{Doctor:} How do we best proceed with the treatment of this patient? \\
\\

& \textbf{Appendix:} \\

& \begin{minipage}{4cm}
    \centering
    MRI-A, 11/01/2023 \\
    \includegraphics[width=\linewidth]{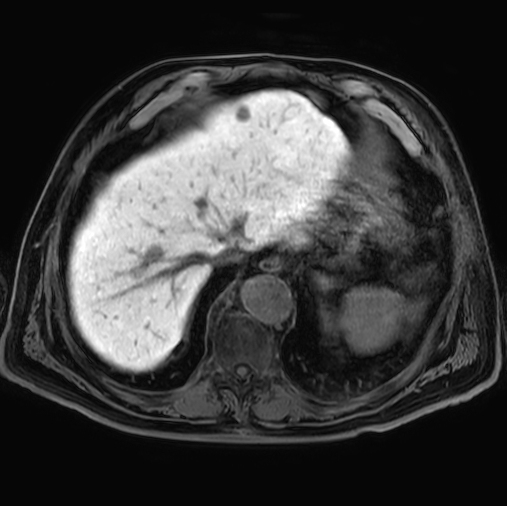} 
  \end{minipage}%
  \hspace{1mm} 
  \begin{minipage}{4cm}
    \centering
    MRI-A, 02/10/2024 \\
    \includegraphics[width=\linewidth]{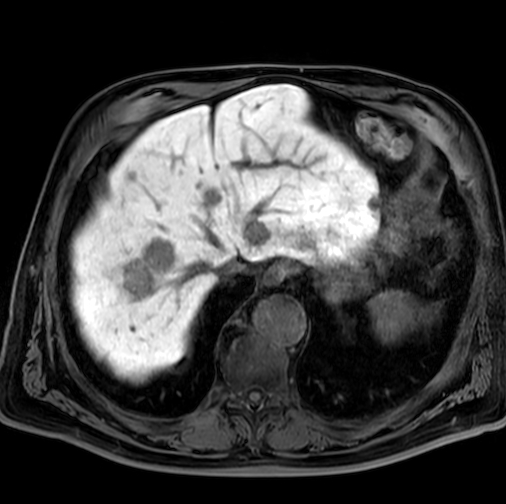} 
  \end{minipage} \\

\\
& \textbf{Model Response:} \\

& Given Mr. Garcia's \colorbox{lightblue}{progression of metastatic colorectal cancer}, as evidenced by the \colorbox{lightblue}{significant increase in the overall area of hepatic lesions} from 1,315 square units to 2,809 square units \colorbox{lightblue}{[Tool]}, which indicates a \colorbox{lightblue}{growth by a factor of 2.14 [Tool]}, a comprehensive and multidisciplinary approach is warranted. The strategy for Mr. Garcia should include the following steps, aligned with the structured approach and medical guidelines: \\

& \textbf{1. Re-evaluation of the Molecular Profile:} Given the progression of disease, it's crucial to reassess the molecular characteristics of Mr. Garcia's tumor. This includes testing for RAS, BRAF mutations, and MSI status, as these can significantly influence the choice of targeted therapies [1], [2]. If not previously done, testing for HER2 amplification and NTRK fusions should also be considered due to their potential actionability [10], [124]. \\

& \textbf{2. Assessment for Second-Line Chemotherapy Options:} Based on the molecular profile and prior treatment with FOLFOX, if Mr. Garcia's tumor is RAS wild-type and without BRAF V600E mutation, options may include anti-EGFR therapy (cetuximab or panitumumab) combined with irinotecan-based chemotherapy [2], [11]. For RAS or BRAF mutant tumors, or if anti-EGFR therapy is not suitable, consideration of FOLFIRI with bevacizumab or aflibercept could be appropriate [2], [7]. \\

& \textbf{3. Consideration of Immunotherapy:} For tumors that are MSI-H or dMMR, immunotherapy with \colorbox{red}{pembrolizumab} or nivolumab, with or without ipilimumab, should be considered [3], [16], [26], [27]. This is based on the significant efficacy of immune checkpoint inhibitors in this subset of patients. \\

& \textbf{4. Liver-Directed Therapies:} Given the liver-dominant nature of Mr. Garcia's metastatic disease, exploring liver-directed therapies such as radioembolization (Y-90), transarterial chemoembolization (TACE), or ablative therapies could provide local control and potentially palliate symptoms [38], [40], [48], [49]. These options should be discussed in a multidisciplinary team (MDT) setting, considering Mr. Garcia's overall health status and liver function. \\

& \textbf{5. MDT Discussion and Supportive Care Optimization:} An MDT discussion involving medical oncology, surgical oncology, radiology, and palliative care is essential to integrate the above considerations into a personalized treatment plan for Mr. Garcia [Strategy]. Concurrently, optimization of supportive care measures to manage symptoms and maintain quality of life is paramount [121]. \\

& \textbf{6. Clinical Trial Enrollment:} Finally, given the refractory nature of Mr. Garcia's disease, exploration of eligibility for clinical trials testing novel agents or combinations should be a priority. This could offer access to emerging therapies and contribute to the advancement of treatment for metastatic colorectal cancer [136]. \\
& In summary, Mr. Garcia's treatment plan should be personalized based on the latest molecular profiling, consider second-line systemic therapy options tailored to his tumor's characteristics, evaluate the feasibility of liver-directed therapies, and ensure supportive care needs are met, all within the context of an MDT discussion. Participation in clinical trials should also be actively pursued. \\
& To further refine our approach in managing Mr. Garcia's condition, I would need to analyze histologic images for genetic mutations that could guide targeted therapy selection. Could you provide the path to the folder containing Mr. Garcia's histologic images? This will enable me to perform genetic modeling to check for MSI, BRAF, and KRAS mutations. Additionally, knowing any specific genetic alterations identified would allow me to consult relevant databases for targeted treatment options, enhancing the precision of Mr. Garcia's treatment plan. Lastly, detailed information regarding Mr. Garcia's eligibility criteria for clinical trials would allow for a targeted search for novel therapeutic opportunities, potentially offering access to groundbreaking treatments. \\

\bottomrule 


Le & Mr. Lee, a 72-year-old individual, was initially diagnosed with pancreatic cancer in December 2022. Following his diagnosis, he underwent a pancreatoduodenectomy (Whipple procedure), and completed five cycles of gemcitabine chemotherapy. Mr. Lee chose to conclude the chemotherapy sessions based on personal preference, although he was tolerating the treatment well. He has been under regular observation since then. In November of the previous year, a CT scan identified a small lung nodule, raising concerns about a potential recurrence of the cancer (The CT slides are in the ‘Imaging’ directory under his family subdirectory for evaluation). After consulting with Mr. Lee, a decision was made to adopt a watchful waiting approach. Recently, another CT scan was conducted and the results have been uploaded to the server. Despite these medical concerns, Mr. Lee reports feeling well and maintains an active lifestyle, engaging in sports activities twice weekly. He currently exhibits no clinical symptoms indicative of cancer progression. Molecular analysis of the tumor has revealed the presence of a KRAS G12C mutation.

\\
& \textbf{Doctor:} Please provide an overview of the standard chemotherapy protocols recommended for situations similar to Mr. Lee's, in alignment with current medical guidelines. Additionally, investigate and summarize the most recent online information regarding treatment options for KRAS G12C mutations, specifically for Mr. Lee's case. \\
\\

& \textbf{Appendix:} \\

& \begin{minipage}{4cm}
    \centering
    CT Lung, November 2023 \\
    \includegraphics[width=\linewidth]{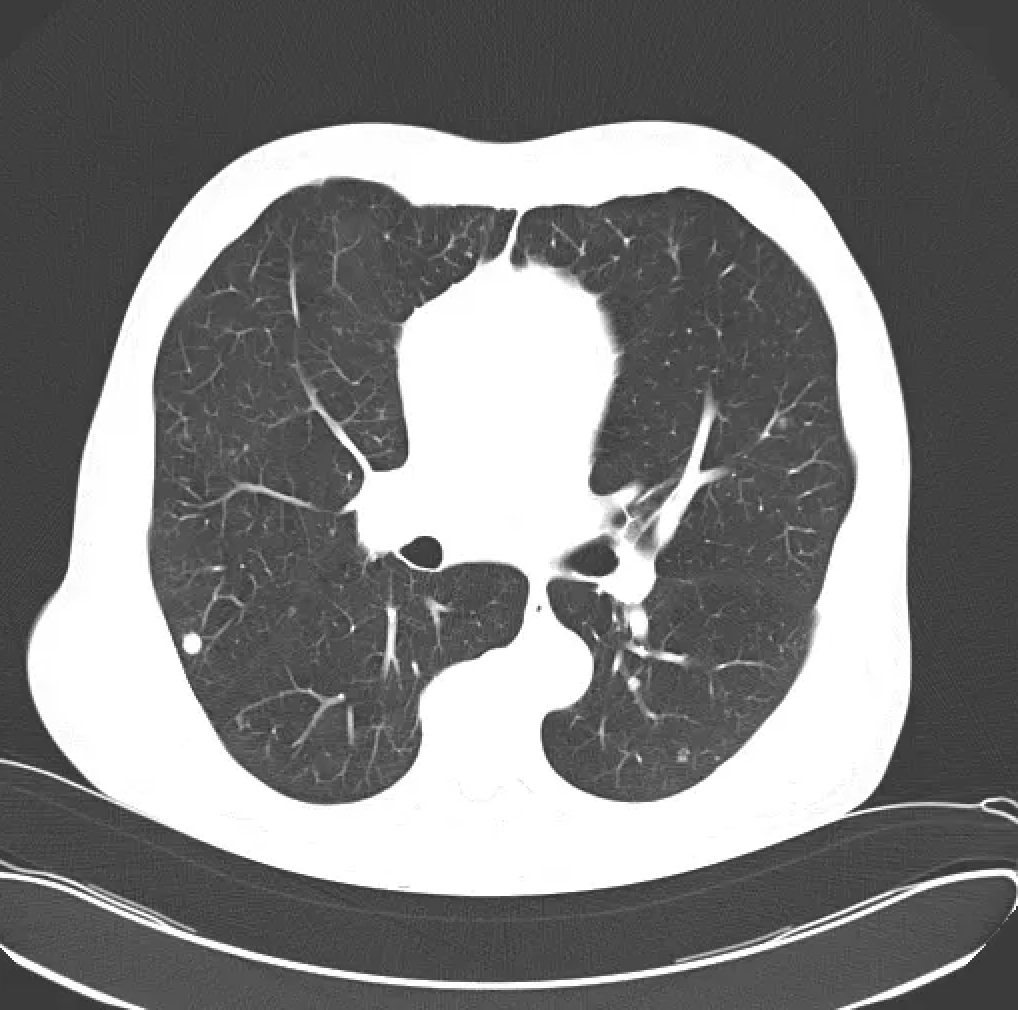} 
  \end{minipage}%
  \hspace{1mm} 
  \begin{minipage}{4cm}
    \centering
    CT Lung, February 2024 \\
    \includegraphics[width=\linewidth]{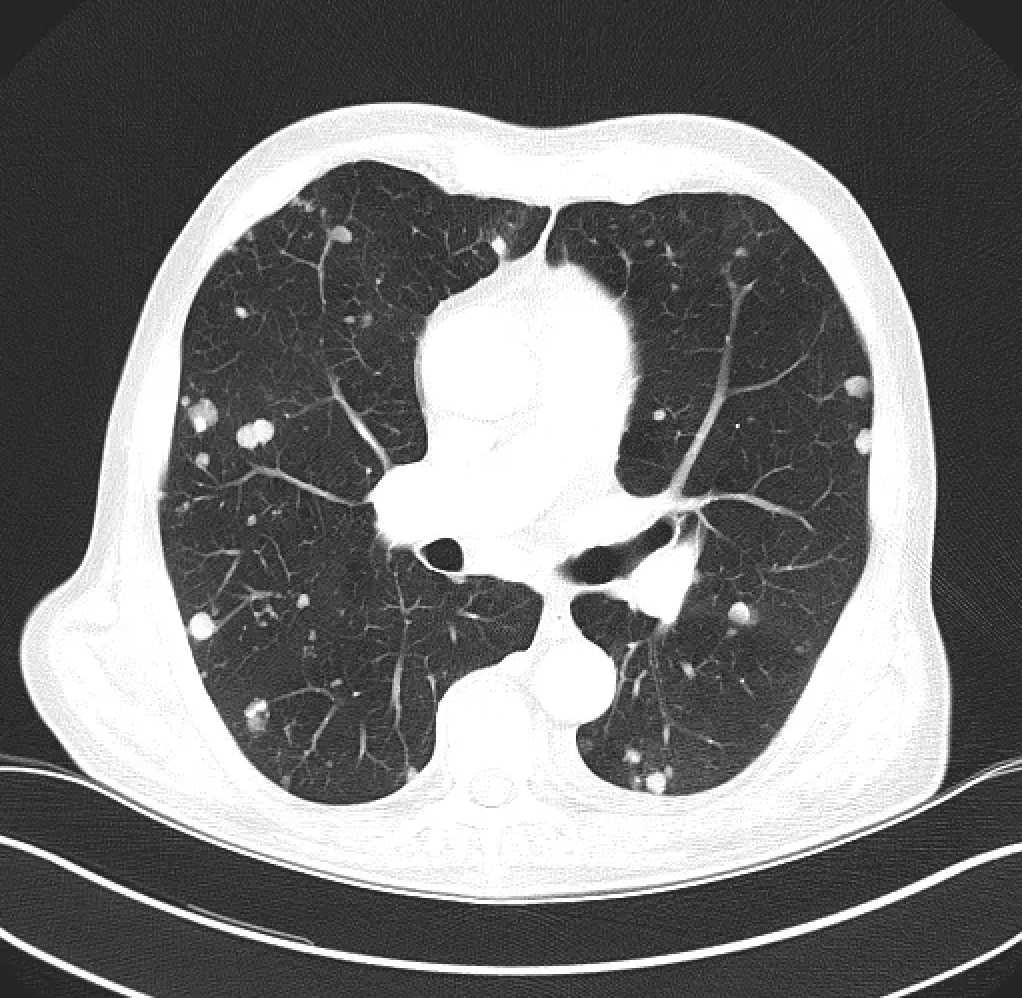} 
  \end{minipage} \\

\\
\\
\\
& \textbf{Model Response:} \\

& Given Mr. Lee's history of pancreatic cancer, post-Whipple procedure status, and completion of five cycles of gemcitabine chemotherapy, along with the \colorbox{lightblue}{recent CT scan} \colorbox{lightblue}{findings indicating progression of lung nodules}, a comprehensive approach is warranted. The presence of a KRAS G12C mutation in his tumor adds a significant layer to his treatment planning, as this \colorbox{lightblue}{mutation opens the door to targeted therapy options} \colorbox{lightblue}{that have shown promise in recent studies [Tool]}. \\
\\

& \textbf{Standard Chemotherapy Protocols} \\
& For patients like Mr. Lee, who have undergone surgical resection followed by adjuvant chemotherapy but show signs of disease recurrence or progression, the choice of further treatment depends on several factors including performance status, extent of disease, and molecular tumor profile. In the absence of specific contraindications, and considering Mr. Lee's good performance status and active lifestyle, a more aggressive approach could be considered. \\

& \textbf{1. \colorbox{orange}{Modified FOLFIRINOX}:} This regimen, consisting of oxaliplatin, irinotecan, fluorouracil (5-FU), and leucovorin, has been recommended for patients with a good performance status in the metastatic setting and could be considered here, despite its intensive nature [4][5]. The regimen's efficacy in the adjuvant setting post-surgery suggests its potential utility in cases of recurrence [13][20]. \\

& \textbf{2. Gemcitabine and Nab-Paclitaxel:} This combination has shown efficacy in the metastatic pancreatic cancer setting and could be an option for Mr. Lee, given his previous tolerance of gemcitabine [16][18]. This regimen might offer a balance between efficacy and tolerability. \\
\\

& \textbf{Targeted Therapy for KRAS G12C Mutation} \\
& The discovery of the KRAS G12C mutation in Mr. Lee's tumor is pivotal. Recent advancements have led to the development of targeted therapies for this specific mutation, which is traditionally considered challenging to treat \colorbox{lightblue}{[Tool]}. \\

& \textbf{1. Sotorasib:} As the first targeted agent with regulatory approval for KRAS G12C-mutated NSCLC, sotorasib represents a novel treatment option. Although primarily studied in lung cancer, its efficacy in KRAS G12C-positive tumors suggests potential applicability in Mr. Lee's case, given the lung nodules' progression [26][29][33]. The reported durable clinical responses and manageable safety profile make it a compelling option \colorbox{lightblue}{[Tool]}. \\

& \textbf{2. Divarasib (GDC-6036):} This agent has shown durable responses across KRAS G12C-positive tumors, including low-grade adverse events, which could be advantageous for maintaining Mr. Lee's quality of life \colorbox{lightblue}{[Tool]}. \\

& \textbf{3. Adagrasib:} Similar to sotorasib, adagrasib has shown efficacy in KRAS G12C-mutated tumors. Its use, alone or in combination with cetuximab, has been evaluated in colorectal cancer patients with this mutation, showing promising efficacy \colorbox{lightblue}{[Tool]}. Given the cross-tumor activity of KRAS G12C inhibitors, adagrasib could be considered for Mr. Lee, especially in light of the lung nodules' progression. \\
\\

& \textbf{Conclusion and Next Steps} \\
& For Mr. Lee, a dual approach incorporating systemic chemotherapy with consideration for targeted therapy against the KRAS G12C mutation appears warranted. Given the progression in lung nodules and the presence of the KRAS G12C mutation, exploring the use of sotorasib or adagrasib, in consultation with a multidisciplinary team, could provide a targeted approach to his treatment. Additionally, enrollment in clinical trials investigating these or new agents for KRAS G12C mutations should be considered, offering Mr. Lee access to cutting-edge treatments and contributing to the growing body of knowledge on managing this genetic alteration \colorbox{lightblue}{[Tool]}. \\
& Regular monitoring and supportive care measures are essential to maintain Mr. Lee's quality of life during treatment, including managing any side effects from the targeted therapy and providing psychological support as needed. This approach aligns with current medical guidelines and reflects a personalized treatment strategy based on Mr. Lee's unique medical history, current condition, and molecular tumor profile [4][5][13][20][26][29][33]. \\
& To further refine our approach and ensure the most accurate and personalized treatment recommendations for Mr. Lee, I would need to delve deeper into specific aspects of his condition. If you could provide more details on the lung nodules, this would help in assessing the dynamics of the disease more accurately. Additionally, if histopathology images are available, providing the unique identifier for Mr. Lee along with specifying interest in the KRAS mutation would enable a precise genetic modeling to confirm the KRAS G12C mutation status. This detailed information is crucial for tailoring the most effective treatment strategy. \\

\bottomrule 


Lo & Mr. Lopez, aged 54 (DOB: 03/01/1970, Patient ID: AA-21-C), visited our oncology department for a consultation following his recent diagnosis of left-sided colorectal cancer, which was discovered during a routine colonoscopy where a biopsy was taken. He had an appointment at the radiology department earlier today for a CT scan; the images have been uploaded for our review, although we are currently awaiting the radiologist's report. Mr. Lopez reports feeling generally well, with an ECOG performance status of 1, and he exhibits no specific symptoms related to the tumor. His medical history includes hypertension and a past humerus fracture in 2013. He is currently not on any medication.

\\
& \textbf{Doctor:} What is the optimal management approach for this patient? Develop a detailed treatment plan. Should we consider local treatment options such as surgical intervention or liver-directed therapies? \\
\\
\\
\\
\\
\\
\\
\\

& \textbf{Appendix:} \\

& \begin{minipage}{4cm}
    \centering
    CT Abdomen \\
    \vspace{2mm}
    \includegraphics[width=\linewidth]{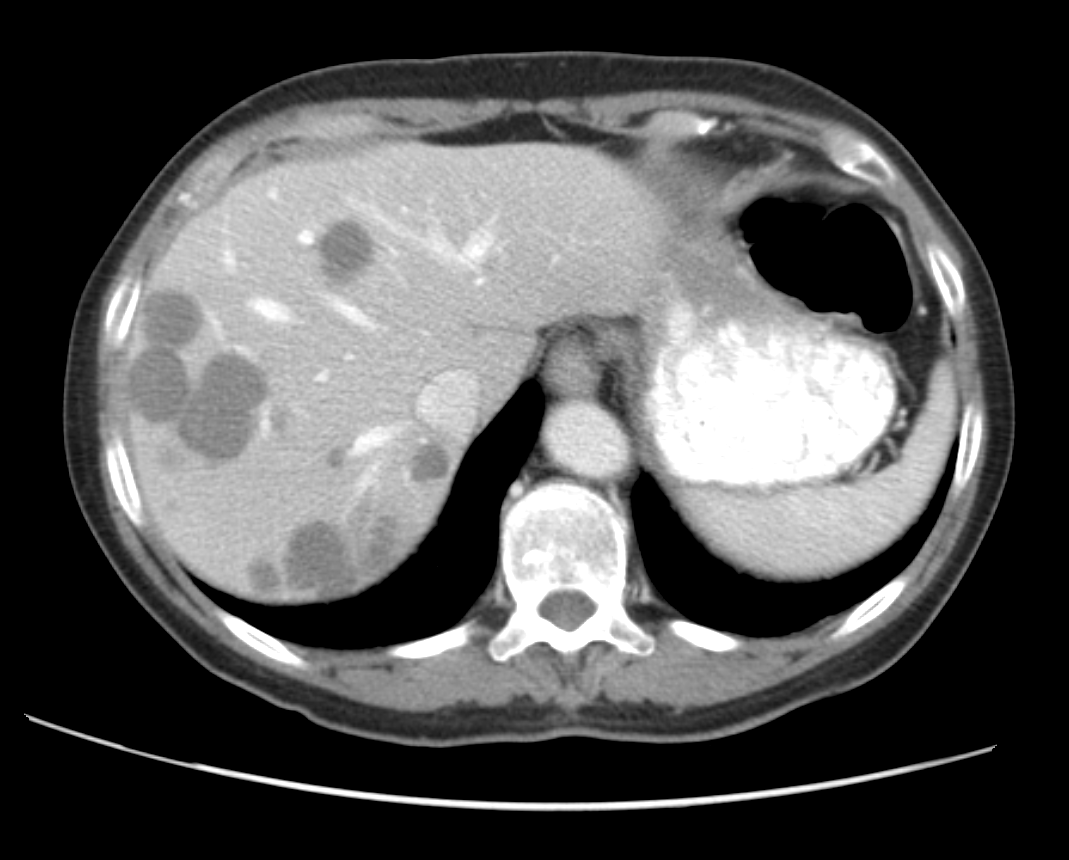} 
    \end{minipage}%
    \hspace{1mm}
    \begin{minipage}{6cm}
    \centering
    Histology \\
    \includegraphics[width=\linewidth]{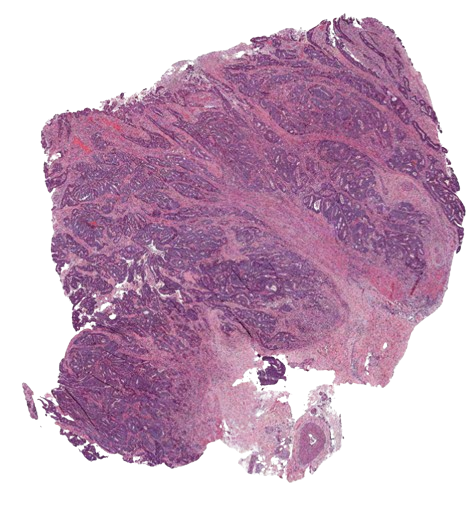} 
  \end{minipage} \\

& \noindent \textbf{Model Response:} \\ 

& Given Mr. Lopez's diagnosis of left-sided colorectal cancer with \colorbox{lightblue}{potential liver metastases,} \colorbox{lightblue}{MSS status, and wild-type BRAF and KRAS genes [Patient], [Tool]}, the optimal management approach should be multi-pronged, incorporating systemic chemotherapy, evaluation for surgical resection or liver-directed therapies, and supportive care measures.  \\

& \textbf{1. Systemic Chemotherapy Initiation:} Based on the NCCN guidelines [1], for Mr. Lopez, who has MSS, left-sided colorectal cancer without BRAF or KRAS mutations, a combination chemotherapy regimen such as FOLFOX (oxaliplatin plus fluorouracil and leucovorin) or FOLFIRI (irinotecan plus fluorouracil and leucovorin) should be considered as the first line of treatment. Given the absence of RAS mutations, the addition of an anti-EGFR agent, specifically cetuximab or panitumumab, is recommended to enhance treatment efficacy [1], [74]. \\

& \textbf{2. Further Imaging and Assessment for Liver Metastases:} Before deciding on local treatments, further characterization of the liver lesions with MRI or PET-CT is advised to confirm metastasis and assess resectability [7]. This step is crucial as it influences the decision regarding local versus systemic therapy and the potential for curative intervention [4]. \\

& \textbf{3. Evaluation for Surgical Resection or Liver-Directed Therapies:} If the liver metastases are confirmed and deemed limited and potentially resectable, Mr. Lopez should be evaluated by a multidisciplinary team including a surgical oncologist [4], [6]. Options such as radiofrequency ablation (RFA), microwave ablation (MWA), or stereotactic body radiotherapy (SBRT) could be considered if surgical resection is not feasible due to the location or number of lesions, or if Mr. Lopez's health status does not permit surgery [21], [23], [84]. \\

& \textbf{4. Maintenance Therapy and Regular Monitoring:} Following the initial treatment phase, transitioning Mr. Lopez to maintenance therapy with a less intensive regimen, such as fluoropyrimidine alone or in combination with bevacizumab, could be considered to manage the disease while minimizing side effects [32]. Regular monitoring through imaging studies and tumor markers will be crucial to adjust the treatment plan based on the disease's response and progression. \\

& \textbf{5. Comprehensive Supportive Care:} Throughout the treatment process, it is essential to provide supportive care measures aimed at managing symptoms, improving quality of life, and addressing any treatment-related side effects. This includes nutritional support, pain management, psychological support, and management of chemotherapy or targeted therapy-related toxicities [5]. \\
& In summary, Mr. Lopez's treatment plan should start with systemic chemotherapy combined with an anti-EGFR agent, followed by a detailed assessment of liver metastases for potential resectability or suitability for liver-directed therapies. The treatment strategy should be flexible, with adjustments based on response to therapy and Mr. Lopez's overall health status. A multidisciplinary approach involving oncologists, radiologists, surgical oncologists, and supportive care specialists is essential for optimal management. \\
& To further refine Mr. Lopez's treatment strategy with precision and up-to-date research findings, could you provide the exact genetic mutations beyond BRAF and KRAS wild-type status? This will enable a detailed OncoKB analysis to identify any additional targeted therapy options. Additionally, for a comprehensive assessment of the liver metastases, please supply the file path to the most recent CT scan images along with the coordinates for each visible lesion. This will assist in accurately determining the extent of liver involvement and the feasibility of local treatments. \\

\bottomrule 


S & Mrs. Smith, a 57-year-old patient with the System ID B13-22A, is under the care of our oncology department due to her diagnosis of metastatic colorectal cancer in September 2023. Today marks her initial visit to our department, and she has provided a brief summary of her medical journey thus far: \\
\\
& \*****\\
& To our esteemed colleague, \\
\\
& This correspondence pertains to Mrs. Smith, born on March 31, 1977, who is receiving oncological treatment in our facility. \\
\\
& \textbf{Primary Diagnosis:} \\
& Colorectal cancer, staged as T3, N2, M+ (HEP), identified in May 2023. \\
& Genetic profiling of the liver biopsy from May 2023 confirmed KRAS wild type, BRAF wild type, and MSS status. \\
& \textbf{Secondary Diagnoses:} \\ 
& Hypothyreosis \\ 
& arterial Hypertension \\ 
& Chemotherapy-induced polyneuropathy \\
& \textbf{Timeline:} \\ 
& 05/2023: Diagnosis established for colorectal cancer with hepatic metastasis, substantiated by liver biopsy revealing KRAS wt, BRAF wt, MSS. \\ 
& 05-08/2023: Administered 6 cycles of FOLFOX + Panitumumab \\ 
& 08/2023: Imaging studies of the thorax and abdomen indicated a partial remission in the known hepatic metastases without the emergence of new metastases. The primary tumor remained stable. \\ 
& 08/2023-11/2023: Continued treatment with six cycles of FOLFOX and Panitumumab, with a reduction in Oxaliplatin dosage to 80\% from the third cycle onwards (totaling nine cycles) due to the onset of polyneuropathy. \\ 
& 11/23 - 02 / 2024: 6 cycles 5-fluorouracil with Panitumumab \\ 

& \*****\\
& Mrs. Smith has also brought along an external CT scan\footnotemark~of the chest and abdomen dated February 2024 for our evaluation. Additionally, a recent tumor biopsy was conducted, though the reports are yet to be furnished.

\\
& \textbf{Doctor:} What is the current state of the patient's health, particularly with respect to the liver metastases previously identified? Are these metastases showing signs of stability? Based on the current condition, is it advisable to persist with the regimen of 5-Fluorouracil and Panitumumab? \\
\\

& \textbf{Appendix:} \\

& \begin{minipage}{4cm}
    \centering
    MRI-A, 08/11/2023 \\
    \includegraphics[width=\linewidth]{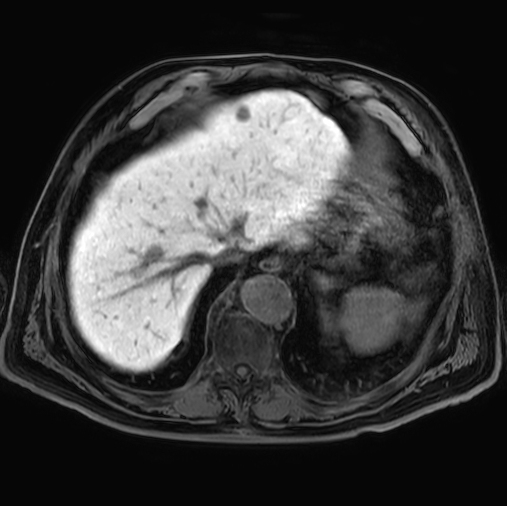} 
  \end{minipage}%
  \hspace{1mm} 
  \begin{minipage}{4cm}
    \centering
    MRI-A, 02/10/2024 \\
    \includegraphics[width=\linewidth]{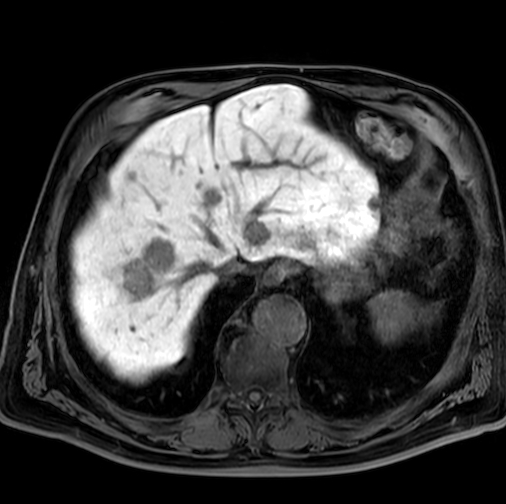} 
  \end{minipage} \\
& \begin{minipage}{6cm}
    \centering
    Histology \\
    \includegraphics[width=\linewidth]{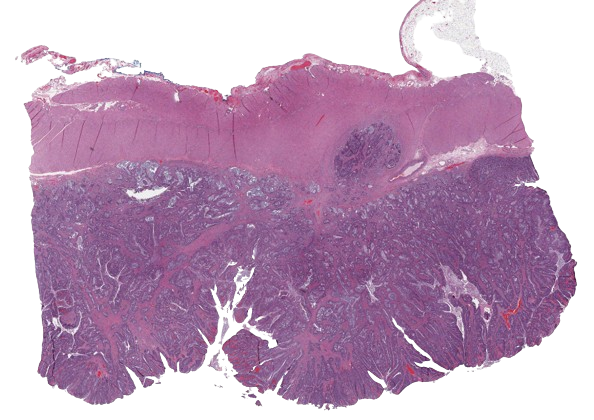} 
  \end{minipage} \\

& \textbf{Model Response:} \\

& Given the comprehensive details provided about Mrs. Smith's medical history and the recent diagnostic findings, a thorough analysis reveals critical insights into her current health status, particularly concerning her \colorbox{lightblue}{liver metastases} and the implications of the \colorbox{lightblue}{newly detected KRAS mutation} on her treatment regimen. \\ 

& \textbf{Current Health Status and Liver Metastases:} \\ 
& The radiology report analysis comparing MRI scans from February 2024 and November 2023 indicates a notable change in the liver metastases. The \colorbox{red}{increased brightness} in the February 2024 scan compared to the November 2023 scan suggests a \colorbox{lightblue}{potential progression or alteration in the liver metastases [Tool]}. However, \colorbox{red}{without} \colorbox{red}{additional clinical context, such as liver function tests or symptoms, the clinical} \colorbox{red}{significance of this radiological change remains uncertain.} It's crucial to correlate these imaging findings with clinical assessments and laboratory results including tumor markers to accurately determine the current state of the liver metastases. \\ 
\\ 

& \textbf{Implications of KRAS Mutation:} \\ 
& The \colorbox{lightblue}{detection of a KRAS mutation} (with a probability of 0.52) represents a significant change from the initial genetic profiling, which identified Mrs. Smith as KRAS wild type \colorbox{lightblue}{[Tool]}. This finding is crucial as it impacts the efficacy of the current treatment regimen. Anti-EGFR therapies, such as Panitumumab, are known to be ineffective in the presence of KRAS mutations [117][118]. The presence of a KRAS mutation suggests that Mrs. Smith's cancer has acquired resistance to the anti-EGFR monoclonal antibody, thereby reducing the effectiveness of her current treatment regimen of 5-Fluorouracil and Panitumumab. \\ 

& \textbf{Treatment Recommendations:} \\ 
& Given the new genetic mutation analysis revealing a KRAS mutation, continuing the regimen of 5-Fluorouracil and Panitumumab is not advisable [117][118]. The detection of a KRAS mutation necessitates a reevaluation of Mrs. Smith's treatment plan. Alternative treatment options should be considered, focusing on regimens that do not rely on the efficacy of anti-EGFR therapies. Options may include chemotherapy combinations without anti-EGFR monoclonal antibodies or considering enrollment in clinical trials exploring novel agents targeting KRAS mutations or other pathways relevant to her tumor's genetic profile [120][126]. \\ 
\\
& \textbf{Further Actions:} \\ 
& \textbf{1. Clinical Correlation:} A multidisciplinary team discussion is recommended to integrate the new radiological and genetic findings with Mrs. Smith's clinical status, including any new symptoms or changes in her overall health. \\ 
& \textbf{2. Additional Testing:} Further genetic testing to confirm the KRAS mutation subtype could provide insights into targeted therapy options, including KRAS G12C inhibitors if applicable [120]. \\ 
& \textbf{3. Treatment Adjustment:} Based on the KRAS mutation, switching to a chemotherapy regimen not reliant on EGFR inhibition, such as FOLFIRI (irinotecan, fluorouracil, and leucovorin), possibly in combination with bevacizumab if not contraindicated, should be considered [19][20]. \\ 
& \textbf{4. Clinical Trials:} Exploration of clinical trials targeting KRAS mutations or other novel therapies suitable for Mrs. Smith's genetic profile and disease state is highly recommended [126]. \\ 
& In conclusion, the new findings necessitate a significant adjustment in Mrs. Smith's treatment strategy, moving away from anti-EGFR therapies due to the detected KRAS mutation and considering alternative systemic therapies or clinical trial enrollment for novel treatments. \\ 
& To enhance our understanding of the progression of Mrs. Smith's liver metastases and refine her treatment options, could you please supply the exact dates of the CT scans you're referring to? This will allow for a precise comparison of changes over time. Additionally, knowing the precise KRAS mutation subtype (e.g., G12C, G12D) will enable a more targeted search for applicable therapies and clinical trials. Lastly, for a comprehensive literature review on the efficacy of her current treatment in the context of her specific KRAS mutation, please provide any keywords or phrases that you believe are crucial for this search, such as "KRAS mutation colorectal cancer resistance." These details are vital for leveraging our resources to their fullest potential, ensuring a tailored approach to Mrs. Smith's care. \\

\bottomrule 

T & Mr Taylor, a 71-year-old male with a significant medical history of Hepatitis C Virus-induced liver cirrhosis, initially presented with small esophageal varices and has been under maintenance treatment with omeprazole and amlodipine for arterial hypertension. In July 2018, he was diagnosed with multifocal hepatocellular carcinoma (HCC), for which he received four sessions of Transarterial Chemoembolization (TACE). Despite these interventions, his HCC advanced, complicated by portal vein thrombosis as of February 2023. His liver function, as per the Child-Pugh classification, remains at A5, his ECOG PS is 0. AFP 90.4; CEA 19.9; Ca19-9 16.7. \\

\\
& \textbf{Doctor:} The progression of Mr. Taylors's HCC in the context of PVT requires a reassessment of therapeutic strategies to address the current complexity of his oncologic and hepatic status. Given his clinical information on the patient and a new CT image that was performed and uploaded, can you suggest the next steps in the management of his disease? \\
\\
\\

& \textbf{Appendix:} \\

& \begin{minipage}{4cm}
    \centering
    CT Abdomen, March 7th, 2024 \\
    \includegraphics[width=\linewidth]{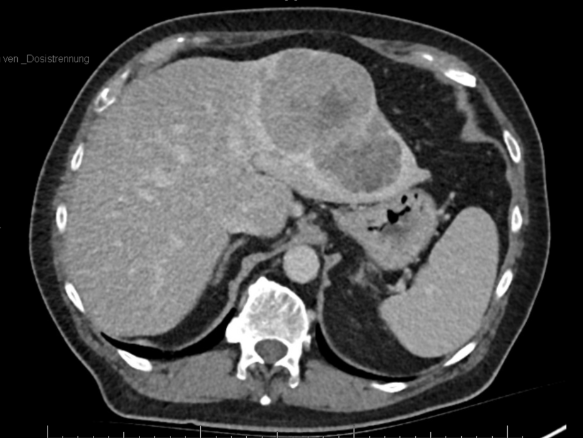} 
  \end{minipage}%

\textbf{Model Response:} \\

& Given Mr. Taylor's \colorbox{lightblue}{advanced HCC} with portal vein thrombosis (PVT), his preserved liver function (Child-Pugh A5), and his good performance status (ECOG PS 0), a multidisciplinary approach is crucial for optimizing his treatment strategy [1], [5], [8]. The presence of PVT categorizes his HCC as advanced or BCLC stage C, for which systemic therapy is recommended over locoregional therapies [3], [4], [59].\\
& First-line systemic therapy with atezolizumab plus bevacizumab is recommended for patients like Mr. Taylor, who have advanced HCC, no contraindications to bevacizumab, and preserved liver function [60], [93]. This recommendation is based on the IMbrave150 trial, which demonstrated a significant survival benefit of atezolizumab-bevacizumab over sorafenib [93]. However, given the presence of esophageal varices, it is imperative to ensure that these have been appropriately managed to minimize the risk of variceal bleeding, a known complication associated with bevacizumab due to its antiangiogenic effects [70], [71].\\
& Considering the progression of his disease despite previous TACE sessions, further locoregional therapies such as additional TACE or radioembolization (TARE) are unlikely to provide significant benefit and may not be appropriate given the extent of his PVT [10], [63]. However, if systemic therapy is contraindicated or not tolerated, and Mr. Taylor's PVT is classified as Vp1 or Vp2 (involving segmental or subsegmental branches of the portal vein), locoregional therapies could be reconsidered in a highly selective manner [7], [63].\\
& Clinical trial enrollment should be explored for Mr. Taylor, given the rapidly evolving landscape of HCC treatment, which may offer access to novel therapeutic agents or combinations not yet available in standard practice [2], [22]. This approach is particularly relevant for advanced HCC, where emerging treatments continue to show promise.\\
& Supportive care measures, including management of ascites and screening for esophageal varices, should be optimized. Therapeutic paracentesis may be required for symptomatic ascites, and beta-blockers or endoscopic variceal ligation (EVL) should be considered for variceal prophylaxis [80], [81], [82]. Given his portal hypertension, as evidenced by splenomegaly and ascites, careful monitoring for complications such as spontaneous bacterial peritonitis (SBP) is warranted [85].\\
& Regular monitoring of Mr. Taylor's liver function, tumor markers, and imaging studies is essential to assess the response to any new treatment and adjust the management plan as necessary [6], [24]. Palliative care consultation should also be involved early to manage symptoms and support quality of life [6].\\

& In summary, the next steps in managing Mr. Taylor's disease should include consideration of systemic therapy with atezolizumab plus bevacizumab, exploration of clinical trial enrollment, and supportive care measures tailored to his specific needs and disease status.\\

& To further refine our approach and ensure the most accurate and personalized treatment recommendations for Mr. Taylor, could you please supply the path to the histologic or histopathology images of Mr. Taylor's liver tumor? This will enable a detailed analysis to identify potential genetic markers critical for his treatment planning. Additionally, if you could list any specific genetic alterations already identified in Mr. Taylor's case, it would greatly enhance our ability to query targeted treatment options and relevant clinical trials. For the most recent imaging studies, specifying the regions of interest would allow for precise measurements and better assessment of the tumor's response to treatment over time. Lastly, providing specific search terms related to Mr. Taylor's condition and any known mutations will facilitate a more focused literature search to support evidence-based decision-making.\\

\bottomrule 

W & Mrs. Williams, born on September 5, 1967, and identified as BC.33-2, is a 57-year-old woman who was diagnosed with colorectal cancer in 2020. Following her diagnosis, she underwent surgical removal of the tumor but experienced a prolonged hospital stay due to complications of postoperative pneumonia and sepsis. She was treated with adjuvant chemotherapy as part of her recovery process.
Up until November of the previous year, Mrs. Williams was under consistent follow-up care. During one of these follow-up visits, imaging revealed the presence of a solitary metastasis in her liver, as detailed in the accompanying radiology report: \\

& \/***\\
& \textbf{Radiology Report, CT Abdomen November 2023}\\
& Mrs Williams, *03/05/1966 \\
& Procedure: CT Abdomen with iv Contrast\\
& Clinical Information: Post colorectal cancer, status post total mesorectal excision. Evaluation for metastasis.\\
& Technique: Multi-phase imaging of the abdomen was performed following the administration of intravenous contrast material.\\
& Findings:\\
& Liver: A single hypodense lesion noted in the left lobe of the liver, measuring approximately 0.5 cm in its greatest dimension (Location: [250, 185, 295, 230]). This finding is suggestive for liver metastasis post colorectal cancer. No additional focal liver lesions are identified.\\
& Biliary Tree: The intrahepatic and extrahepatic bile ducts are not dilated. The gallbladder is unremarkable.\\
& Pancreas: The pancreas is normal in size, contour, and enhancement pattern. No mass, cyst, or ductal dilatation is seen.\\
& Spleen: The spleen is normal in size and attenuation with no focal lesions.\\
& Adrenal Glands: Both adrenal glands are normal in size and morphology.\\
& Kidneys: Both kidneys are normal in size, shape, and function without evidence of mass, stone, or hydronephrosis. Small cyst on the left kidney, bosniak 1.\\
& Bowel: There is no evidence of bowel obstruction or wall thickening. The visualized portions of the colon, including the anastomosis site, show no signs of local recurrence.\\
& Vessels: The abdominal aorta and its major branches are patent without evidence of aneurysm or significant stenosis.\\
& Lymph Nodes: No enlarged abdominal or pelvic lymph nodes are identified.\\
& Peritoneum: There are no signs of peritoneal implants or significant ascites.\\
& Impression:\\
& 1 small hypodense lesion in the left lobe of the liver, suggestive of liver metastasis. Given the clinical history, further evaluation with MRI or follow-up imaging may be warranted for characterization.\\
& No evidence of local recurrence post total mesorectal excision for colorectal cancer.\\
& No other significant abnormality detected in the abdomen.\\
& Notes: Uncomplicated biopsy of a liver lesion and dispatch to pathology.\\
& \/***\\

& Following the discovery of liver metastasis, Mrs. Williams underwent six cycles of chemotherapy using the FOLFOX regimen and Panitumumab. Subsequent imaging (CTA\_02\_2024.png) confirmed the metastatic lesion remained stable in location (coordinates: 250, 185, 295, 230), with these findings documented in our system. Genetic analysis revealed the presence of a BRAF V600E mutation in her tumor.\\

\\
& \textbf{Doctor:} What is the optimal management approach for this patient? Develop a detailed treatment plan. Should we consider local treatment options such as surgical intervention or liver-directed therapies? \\
\\

& \textbf{Appendix:} \\

& \begin{minipage}{4cm}
    \centering
    CT Abdomen, November 2023 \\
    \includegraphics[width=\linewidth]{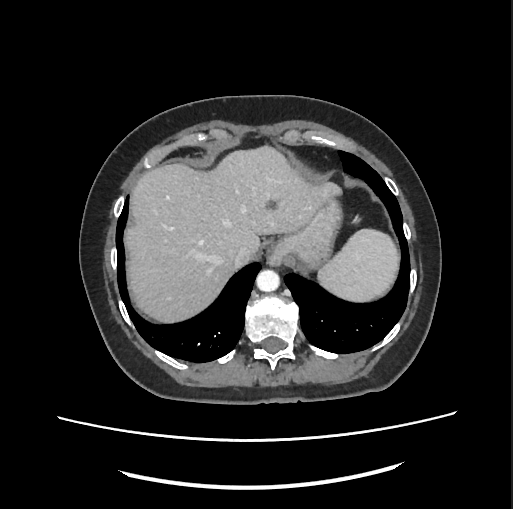} 
  \end{minipage}%
  \hspace{1mm} 
  \begin{minipage}{4cm}
    \centering
    CT Abdomen, February 2024 \\
    \includegraphics[width=\linewidth]{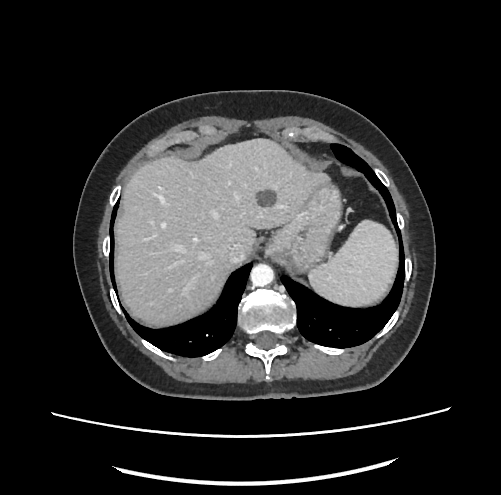} 
  \end{minipage} \\
    \hspace{1mm}
    \\
& \begin{minipage}{6cm}
    \centering
    Histology\footnotemark \\
    \includegraphics[width=\linewidth]{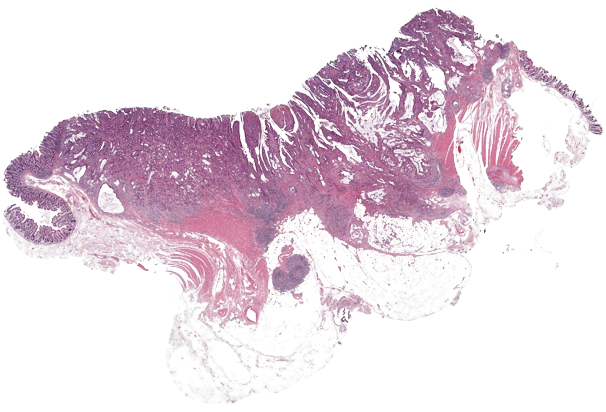} 
  \end{minipage} \\

& \noindent \textbf{Model Response:} \\

& Given Mrs. Williams' progressive disease under the FOLFOX regimen and Panitumumab, as evidenced by the \colorbox{lightblue}{growth of her liver metastasis}, the next steps involve a comprehensive reassessment of her treatment plan based on her genetic profile and current medical guidelines. The \colorbox{lightblue}{increase in the size of the metastasis, with a ratio of approximately 1.799,} \colorbox{lightblue}{clearly indicates progressive disease [Tool].} \\

& \textbf{1. \colorbox{lightblue}{Targeted Therapy for BRAF V600E Mutation:}} Considering the BRAF V600E mutation, the combination of encorafenib and cetuximab is recommended. This combination is FDA-approved for mCRC with a BRAF V600E mutation, after prior therapy [12], [13], [41]. This approach is supported by the efficacy shown in the BEACON CRC trial, which demonstrated a significant improvement in overall survival and response rate with this combination [3], [41]. \\

& \textbf{2. Immunotherapy Consideration:} The \colorbox{lightblue}{MSI-High status of the tumor} suggests a potential benefit from immunotherapy. \colorbox{red}{Pembrolizumab, recommended as first-line therapy} for patients with dMMR mCRC by the ASCO 2022 guideline [20], could be considered given the patient's genetic profile. This option should be discussed with Mrs. Williams, considering the potential for a better response to immunotherapy in MSI-High status tumors [19], [20], [22]. \\

& \textbf{3. Local Treatment Options:} For the solitary liver metastasis, exploring local treatments such as radiofrequency ablation (RFA), stereotactic body radiotherapy (SBRT), or selective internal radiation therapy (SIRT) could be beneficial [46], [47], [48]. These options might offer localized control of the metastasis, especially if systemic therapy options are exhausted or if the metastasis becomes more localized. \\

& \textbf{4. Systemic Therapy Alternatives:} If targeted therapy and immunotherapy are not immediately pursued or if further progression is noted, considering FOLFIRI as a second-line therapy could be an option [76]. However, given the genetic makeup of Mrs. Williams' tumor, targeted therapy options should be prioritized over switching to another chemotherapy regimen [5], [7]. \\

& \textbf{5. Multidisciplinary Team Discussion:} A multidisciplinary discussion involving medical oncologists, surgical oncologists, and radiologists is crucial to tailor the treatment plan to Mrs. Williams' specific condition, genetic makeup, and preferences [56], [57]. This team approach ensures that all potential treatment avenues are explored and that the chosen strategy aligns with the patient's goals and quality of life considerations. \\

& \textbf{6. Patient Consultation:} It is essential to discuss all potential treatment options with Mrs. Williams, including the benefits, risks, and potential side effects of each treatment [6]. Her preferences, quality of life considerations, and any concerns should be central to the decision-making process. \\
& In summary, the immediate next steps involve prioritizing targeted therapy with encorafenib and cetuximab, exploring immunotherapy options due to the MSI-High status, and considering local treatment options for the liver metastasis. A multidisciplinary approach and patient-centered discussion are paramount in deciding the future course of treatment for Mrs. Williams. \\
& To enhance our exploration of second-line therapy options and local treatment possibilities for Mrs. Williams, it would be beneficial if you could provide the exact search terms or phrases you have in mind for investigating the effectiveness of FOLFIRI in BRAF V600E mutated colorectal cancer and the latest advancements in local treatment options for liver metastasis. Additionally, specifying the particular aspects of colorectal cancer treatment you're most interested in could greatly assist in tailoring the search to yield the most relevant and up-to-date guidelines and research findings. This focused approach will ensure we utilize our resources to gather the most pertinent information to aid in Mrs. Williams' treatment planning.\\

\bottomrule 


X & Ms. Xing, a 33-year-old patient, was diagnosed with cholangiocellular carcinoma in early 2023. Following her diagnosis, she underwent a complete surgical resection (R0) and was treated with adjuvant chemotherapy using capecitabine for six months. Subsequent imaging showed no evidence of disease recurrence until September of the same year, when MRI imaging ("September2023.png") revealed a new, solitary metastasis.
  In response, a treatment regimen of gemcitabine and cisplatin was initiated. Considering Ms. Xing's young age, a liver biopsy was performed for comprehensive panel diagnostics. Both imaging results and panel findings are made available for review:\\
& \/***\\
& \textbf{Radiology Report, MRI Abdomen September 2023} \\ 
& Mrs Xing, *02/03/1987\\
& Procedure: MRI Abdomen with iv Contrast \\
& Clinical Question: Post Cholangiocellular carcinoma, new metastasis? Recurrence? Staging. \\
& Technique: MRI Abdomen \\
& Findings: \\
& Liver: There is one single hypointense lesion in the left lobe of the liver, measuring approximately 0.4 cm in its greatest dimension (Location: [475, 250, 490, 275]). Highly suggestive for recurrence of the known cholangiocellular carcinoma. \\
& Biliary Tree: The intrahepatic and extrahepatic bile ducts are not dilated.  \\
& Pancreas: The pancreas appears regular in size, shape, and enhancement, showing no abnormalities such as masses, cysts, or dilation of ducts. \\
& Spleen: The spleen is within normal limits regarding size and density, with no detected lesions. \\
& Adrenal Glands: Both adrenal glands present a normal appearance in terms of size and structure. \\
& Kidneys: The kidneys are normal in all aspects of size, contour, and functionality, with no signs of masses, stones, or hydronephrosis. \\
& Bowel: No obstructions or abnormal thickening is observed in the bowel.  \\
& Vessels: The abdominal aorta and its main branches are unobstructed, with no aneurysms or significant narrowing observed. \\
& Lymph Nodes: There is no enlargement of the abdominal or pelvic lymph nodes. \\
& Peritoneum: No evidence of peritoneal implants or significant fluid accumulation is present. \\
& Impression: \\
& Strong suspicion of a recurrence of the known CCC due to a solitary metastasis in the left liver.\\

& \textbf{Molecular Report Summary:}\\
& Microsatellite-Instability High, Mutation in BRAF V600E and CD74-ROS1 variant fusion.\\
& No alterations in NTRK or FGFR2.\\
& \/***\\

& Currently, the patient experiences recurrent ascites with notable abdominal tension.\\
& Also, a new MRI-scan was performed yesterday (“February2024.png”) by our in house radiologists.\\
& According to the latest radiology report, the MRI scan performed on the patient with the system ID X-09.22 and documented as "February2024.png" revealed a single metastatic lesion in the left lobe of the liver, delineated by the coordinates 455, 270, 505, 320. The report emphasizes the importance of a comprehensive review by comparing these findings with the images from a previous examination in September 2023 during the forthcoming tumor board meeting for a detailed analysis. Also, the imaging findings suggest the presence of peritoneal carcinomatosis. Her system ID is X-09.22.\\

\\
& \textbf{Doctor:} Please investigate the current state of the disease in detail. What does the new MRI scan show? Did the metastasis grow? In case of progress, what targeted or chemotherapy options could we suggest for the tumor board tomorrow? What treatment options does the patient still have according to the official guidelines for cholangiocellular cancer given her medical history. Be aware that the patient is extremely young and we need all options we could find. Also check any options you can find on google or pubmed.\\

\\
\\
\\
\\
\\
\\
\\
\\
\\
\\
\\
& \textbf{Appendix:} \\

& \begin{minipage}{4cm}
    \centering
    MRI Abdomen, September 2023 \\
    \includegraphics[width=\linewidth]{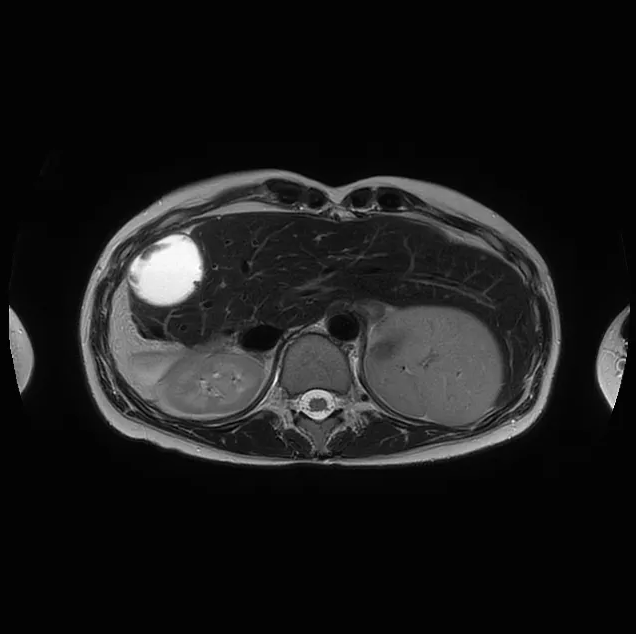} 
  \end{minipage}%
  \hspace{1mm} 
  \begin{minipage}{4cm}
    \centering
    MRI Abdomen,   February 2024 \\
    \includegraphics[width=\linewidth]{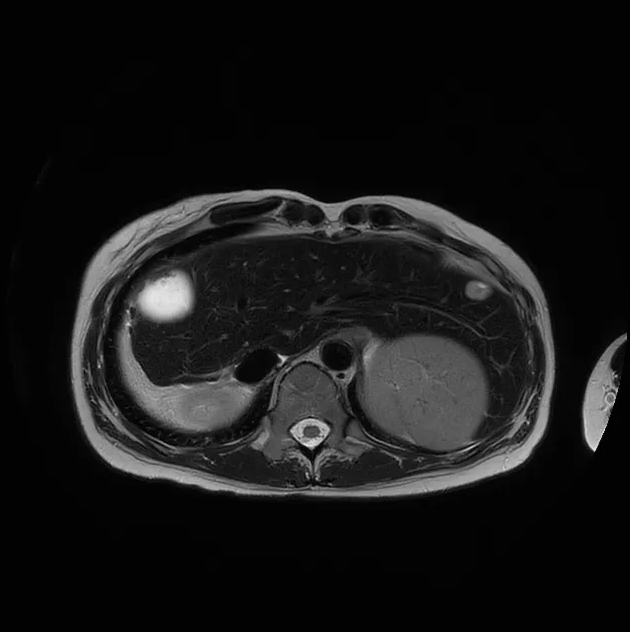} 
  \end{minipage} \\
    \hspace{1mm}
    \\
& \begin{minipage}{6cm}
    \centering
    Histology \\
    \includegraphics[width=\linewidth]{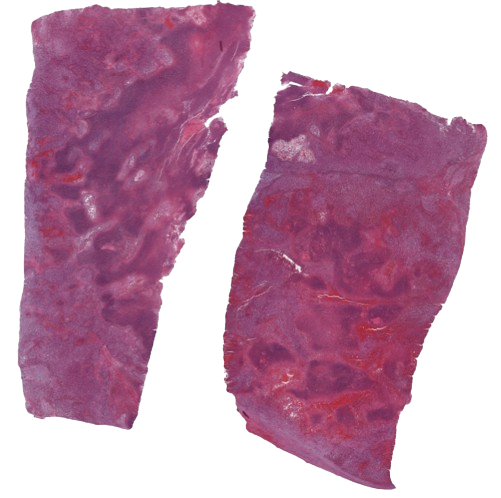} 
  \end{minipage} \\
    
& \noindent \textbf{Model Response:} \\

& The latest MRI scan from February 2024 reveals that the \colorbox{lightblue}{metastatic lesion in Ms. Xing's} \colorbox{lightblue}{left liver lobe has notably enlarged, growing from an area of 133 in September 2023 to 518,} \colorbox{lightblue}{signifying a progression in her condition. This significant increase highlights the critical} \colorbox{lightblue}{need to consider further treatment avenues for Ms. Xing.} \\

& Given Ms. Xing's medical history and the molecular characteristics of her tumor, including the BRAF V600E mutation and CD74-ROS1 variant fusion, several targeted therapy options emerge: \\

& \colorbox{lightblue}{\textbf{1. Dabrafenib and Trametinib:}} This combination is recommended for BRAF V600E-mutated tumors, including cholangiocellular carcinoma, based on their effectiveness in other cancers with similar genetic alterations \colorbox{lightblue}{[Tool]}. The combination has FDA approval for BRAF V600E mutant unresectable or metastatic melanoma and other solid tumors excluding colorectal cancer, with noted effectiveness in biliary tract cancer \colorbox{lightblue}{[Tool].} \\

& \colorbox{lightblue}{\textbf{2. Crizotinib, Entrectinib, and Repotrectinib:}} These drugs are identified treatments for ROS1-positive tumors, primarily in NSCLC. Given the CD74-ROS1 fusion in Ms. Xing's tumor, these options could be explored for their potential applicability, despite the primary association with NSCLC \colorbox{lightblue}{[Tool].} \\

& \colorbox{lightblue}{\textbf{3. Encorafenib and Binimetinib:}} This combination, highlighted through a Google search, could be considered based on their effectiveness in cancers with BRAF V600E mutations \colorbox{lightblue}{[Tool]}. Although not specifically indicated for cholangiocellular carcinoma, their use in similar genetic contexts suggests potential applicability. \\

& Regarding chemotherapy options, the current guidelines for cholangiocellular carcinoma recommend systemic chemotherapy as the primary treatment for advanced disease. Given Ms. Xing's prior treatment with gemcitabine and cisplatin, alternative regimens could include: \\

& - \textbf{FOLFOX} (fluorouracil, leucovorin, and oxaliplatin) as a second-line treatment option, especially for patients who have progressed on a gemcitabine-based regimen [55]. \\
& For Ms. Xing, considering her young age and the urgent need for effective management strategies, participation in clinical trials exploring novel agents or combinations should be strongly considered. This includes trials for targeted therapies based on her tumor's molecular profile or emerging immunotherapy options for cholangiocellular carcinoma. \\

& Given the conflicting information on microsatellite status, further clarification is needed. The \colorbox{lightblue}{'check\_mutations' tool indicated Microsatellite Stability (MSS)}, however, the patient's molecular report specified that there was Microsatellite Instability (MSI). If MSI-High status is confirmed, immunotherapy with pembrolizumab could be considered, as it has shown effectiveness in MSI-High tumors across various cancer types [57]. \\

& In summary, for the tumor board discussion, the recommended approach includes exploring targeted therapy options such as Dabrafenib and Trametinib for the BRAF V600E mutation and potentially Crizotinib or Entrectinib for the CD74-ROS1 fusion. Additionally, considering second-line chemotherapy options and the possibility of clinical trial participation are crucial steps. The significant progression of Ms. Xing's disease necessitates a prompt and comprehensive evaluation of these treatment strategies to optimize her care. \\
& To refine our approach and ensure the most accurate and personalized treatment recommendations for Ms. Xing, could you please provide the latest comprehensive genomic profiling of her tumor? This includes any recent sequencing data that might reveal additional actionable mutations or variants not previously identified. Additionally, detailed information regarding her response to past treatments, including any side effects experienced and the duration of response, would greatly enhance our ability to tailor future therapy options. This data is crucial for leveraging our resources to their fullest potential in guiding Ms. Xing's treatment plan. \\

\bottomrule 

Z & Mr. Zu, a 55-year-old individual, has been diagnosed with pancreatic adenocarcinoma based on histological evidence. Unfortunately, we don't have access to the imaging that led to this diagnosis (classified as T2N1MX). Just yesterday, he underwent comprehensive staging, which included an IV-contrast-enhanced CT scan of the abdomen. While the radiology report is still pending, the images are available for review. Clinically, Mr. Zu is in robust health and is deemed a suitable candidate for a range of therapeutic interventions. \\

\\
& \textbf{Doctor:} Considering Mr. Zu's diagnosis and the current stage of his pancreatic adenocarcinoma, what specific treatment recommendations should we prepare for discussion at tomorrow's tumor board meeting? \\

& \textbf{Appendix:} \\

& \begin{minipage}{4cm}
    \centering
    CT Abdomen, February 2024\footnotemark \\
    \includegraphics[width=\linewidth]{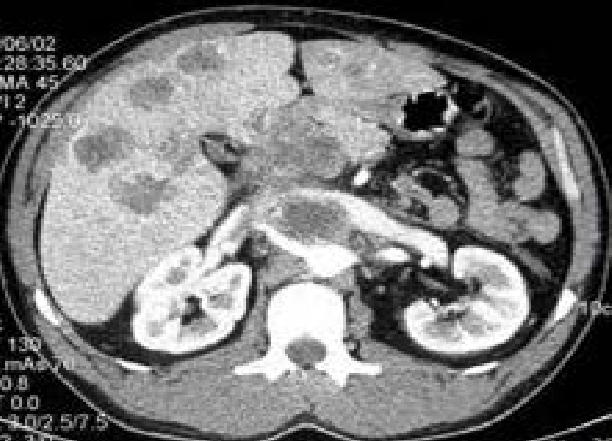} 
  \end{minipage}%

\textbf{Model Response:} \\

& Given Mr. Zu's diagnosis of pancreatic adenocarcinoma with potential \colorbox{lightblue}{liver metastases} \colorbox{lightblue}{as indicated by the presence of multiple hypodense areas within the liver [Tool]}, a comprehensive and multidisciplinary approach is warranted. The treatment strategy should balance aggressive management of metastatic disease with considerations for Mr. Zu's quality of life and preferences. Here's a detailed treatment plan based on current medical guidelines and the specific patient context at hand: \\
& \textbf{1. Confirm Metastatic Status:} Before finalizing the treatment plan, it's crucial to confirm whether the hypodense liver areas represent metastatic disease. This can be achieved through further imaging, such as MRI with liver-specific contrast or PET-CT, and potentially a liver biopsy for histological confirmation [Tool]. The outcome of these investigations will significantly influence the treatment pathway. \\
& \textbf{2. Genetic Profiling:} Although the \colorbox{lightblue}{mutation check encountered an error [Tool]}, pursuing comprehensive genetic testing is essential. This includes testing for MSI, BRAF, KRAS, and potentially other relevant markers like BRCA mutations, given the pancreatic origin of the cancer. The results could open up options for targeted therapies [21], [43]. \\
& \textbf{3. Systemic Chemotherapy:} If metastases are confirmed, systemic chemotherapy becomes the primary treatment approach. For a patient with a good performance status like Mr. Zu, a regimen like FOLFIRINOX (if tolerated) or gemcitabine plus nab-paclitaxel could be considered, as these are standard options for metastatic pancreatic adenocarcinoma [13], [19]. The choice between these regimens should consider Mr. Zu's overall health, comorbidity profile, and personal preferences. \\
& \textbf{4. Local Control Options:} If the liver metastases are limited in number and the primary tumor is controlled, local therapies such as stereotactic body radiotherapy (SBRT) for liver metastases could be discussed. This approach could potentially improve quality of life and overall survival [23], [26]. \\
& \textbf{5. Supportive Care:} Integrating palliative care early in the treatment plan is crucial to manage symptoms and improve quality of life. This includes pain management, nutritional support, and addressing any psychological needs [111], [112]. \\
& \textbf{6. Clinical Trials:} Exploring the availability of clinical trials that Mr. Zu may be eligible for is important. This could include trials focusing on novel therapies or combinations that could offer benefit beyond standard care [154]. \\
& \textbf{7. Multidisciplinary Evaluation:} Ensure that Mr. Zu's treatment plan is discussed in a multidisciplinary tumor board meeting, incorporating input from medical oncology, surgical oncology, radiation oncology, radiology, and palliative care to formulate a comprehensive and individualized treatment plan [150]. \\
& \textbf{8. Follow-Up and Monitoring:} Plan for regular follow-up to assess treatment response, manage side effects, and adjust the treatment plan as necessary. This includes monitoring tumor markers and imaging studies [78].\\
& Given the complexity of Mr. Zu's case, especially with the potential presence of liver metastases, it's imperative that the treatment plan is flexible and responsive to the evolving nature of his condition. Further, the \colorbox{lightblue}{limitations encountered in genetic profiling necessitate a reattempt or alternative methods} \colorbox{lightblue}{to ensure a comprehensive understanding of the tumor's genetic landscape, which could} \colorbox{lightblue}{significantly impact treatment decisions [Tool].}\\
& To further refine our approach and ensure we're leveraging the most accurate and personalized treatment options for Mr. Zu, I would kindly ask for the re-submission of histologic or histopathology images. Specifically, high-resolution images of the affected tissue would be invaluable. These images are crucial for a detailed genetic modeling analysis, aiming to identify MSI, BRAF, and KRAS mutations with precision. Additionally, if you could provide any available clinical data regarding Mr. Zu's health status, including any previous treatments and their outcomes, it would greatly assist in tailoring the query to the OncoKB database. This information will enable us to pinpoint targeted treatment options that are most likely to benefit Mr. Zu, based on his unique genetic profile and clinical scenario.\\

\bottomrule

\footnotetext[1]{Note that the Appendix is only for illustration purposes here. Any data shown is not directly provided to the
agent, however is stored on the filesystem and can be accessed autonomously on demand through retrieving the
necessary information where to find it from the main text.}

\footnotetext[2]{It should be noted that the outputs generated by the GPT-4 model are presented in their original format and structure, with only markdown elements such as bold text and subsection titles being translated into LaTeX, and adding color highlighting.}

\footnotetext[3]{We are aware that the actual images are MRI images, which is correctly identified by the model. Additionally, the model asks the user about the mentioned CT images.}

\footnotetext[4]{We acknowledge that the image in question depicts material from a primary tumor site, rather than from a metastatic liver site as proposed in this hypothetical scenario. This distinction has no bearing on the outcomes or conclusions of our study.}

\footnotetext[5]{This image is obtained from the following publication: Souguir, A. et al. CT-guided celiac plexus neurolysis in the management of pain related to unresectable pancreatic cancer: About 2 cases. Open J. Gastroenterol. 05, 37–41 (2015) under the Creative Commons Attribution 4.0 International License (CC BY 4.0).}

\end{longtable}
\setcounter{footnote}{0}
 
\textbf{Supplementary Table 3. Model Outputs.}
This table presents the unaltered inputs to the model, encompassing
patient contexts, queries (instructions) and the employed data. The
appendix illustrates additional data, which were not directly fed into
the model. Minor adjustments have been applied to the model's outputs
for formatting purposes, specifically converting from markdown to LaTeX
and we mark tool usage in purple, errors in red and potentially adverse
responses in orange.
\clearpage

\begin{longtable}{@{} l >{\raggedright\arraybackslash}p{12cm} @{}}

\caption*{{\Large Supplementary Table 4: Tool Results}}\\
\\ 

\toprule 
\textbf{No.} & \textbf{Tool Results} \\
\midrule 
\endfirsthead 

\toprule
\textbf{No.} & \textbf{Tool Results} \\
\midrule
\endhead 

A & \vspace{-\baselineskip}
\begin{wrapfigure}{l}{0.45\textwidth} 
    \centering
    \includegraphics[width=0.2\textwidth]{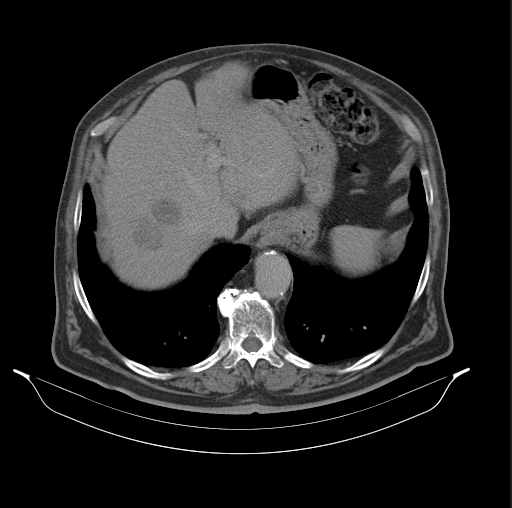} 
    \hspace{1pt} 
    \includegraphics[width=0.2\textwidth]{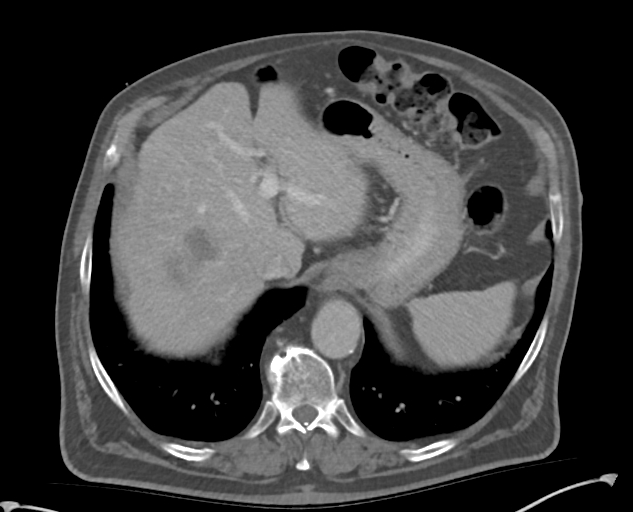} 
\end{wrapfigure}
\textbf{1. Radiology Report:} 
\textbf{CTA\_11\_11\_2023.png}: The initial imaging from November 11, 2023, showed a large area of low attenuation in the liver, which could suggest a \colorbox{mygreen}{liver lesion} or cyst, and evidence of \colorbox{myred}{free fluid} in the abdomen, indicating conditions such as ascites or acute pathology. 

\textbf{CTA\_02\_24\_2024.png}: The latest imaging from February 24, 2024, revealed several hypodense areas in the liver, \colorbox{mygreen}{highly suspicious for metastatic lesions}, particularly in the \colorbox{mygreen}{right lobe}. The liver parenchyma surrounding these lesions appeared normal. The spleen, \colorbox{myred}{pancreas, adrenal glands, and kidneys were unremarkable}, with no signs of bowel obstruction or acute abdomen. The comparison between the two images suggested \colorbox{mygreen}{little to no change in the liver} \colorbox{mygreen}{lesion's size and location but a } \colorbox{myred}{reduction in the amount of free fluid} in the abdomen, which could indicate a response to treatment or a change in the patient's condition. 
\textbf{2. Genetic Predictions:}

- \textbf{Microsatellite Instability (MSI):} The prediction is \colorbox{mygreen}{Microsatellite Stable (MSS)} with a high probability of 0.97, indicating that the cancer is not likely to have high levels of MSI, which is associated with a better response to certain types of immunotherapy.

- \textbf{BRAF Mutation:} The prediction is \colorbox{mygreen}{wild-type} with a very low probability of 0.03, suggesting that there is no BRAF mutation present.

- \textbf{KRAS Mutation:} The prediction is \colorbox{mygreen}{wild-type} with a probability of 0.48, indicating that there is no KRAS mutation present.
\textbf{Summary:}
The radiology reports and genetic predictions provide a comprehensive overview of Mrs. Adams' current condition. The imaging studies show \colorbox{mygreen}{multiple liver lesions consistent with metastases} and a \colorbox{myred}{reduction in abdominal free fluid}, suggesting a potential response to the chemotherapy regimen. The genetic analysis indicates that the cancer is \colorbox{mygreen}{MSS and lacks mutations in BRAF and KRAS}, which could influence the choice of targeted therapies. This information is crucial for planning the next steps in Mrs. Adams' cancer treatment, potentially guiding the selection of targeted therapies based on the genetic makeup of her cancer.\\

\bottomrule 

B & \vspace{-\baselineskip}
\begin{wrapfigure}{l}{0.45\textwidth} 
    \centering
    \includegraphics[width=0.2\textwidth]{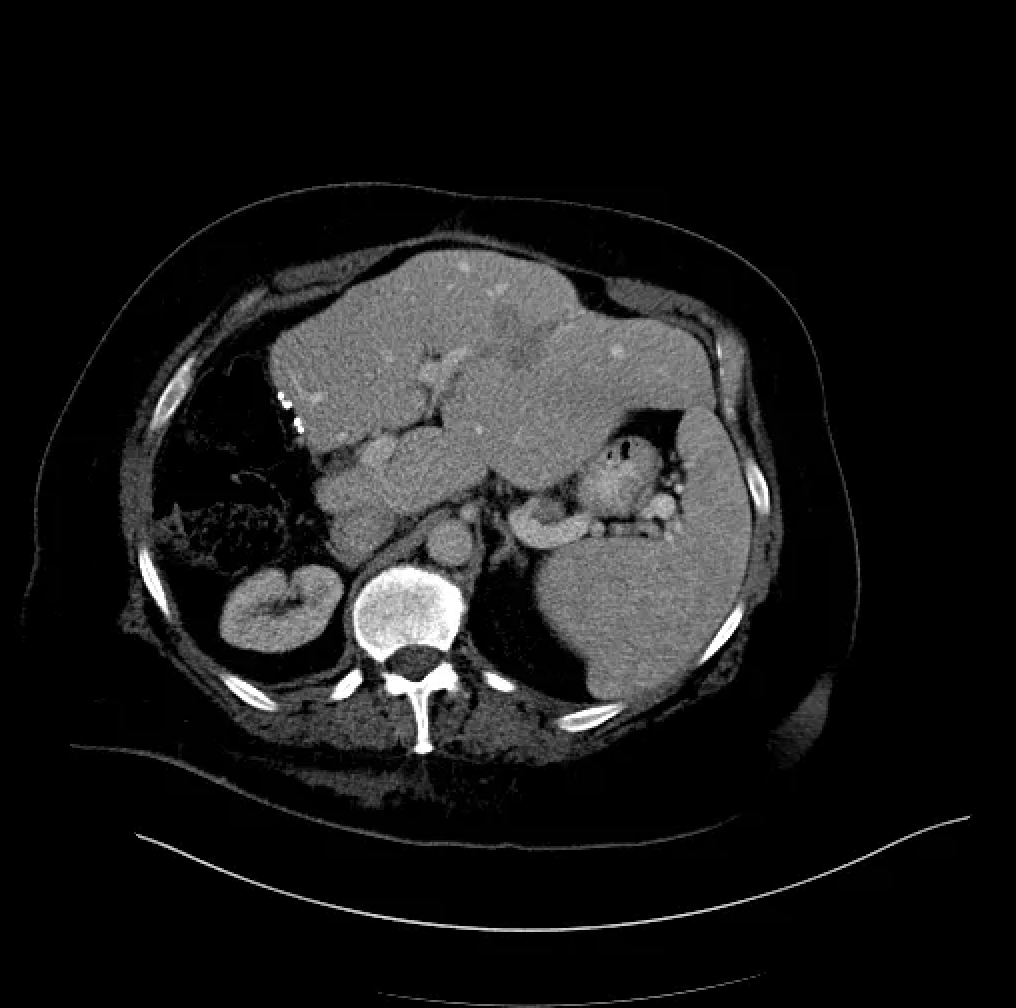} 
    \hspace{1pt} 
    \includegraphics[width=0.2\textwidth]{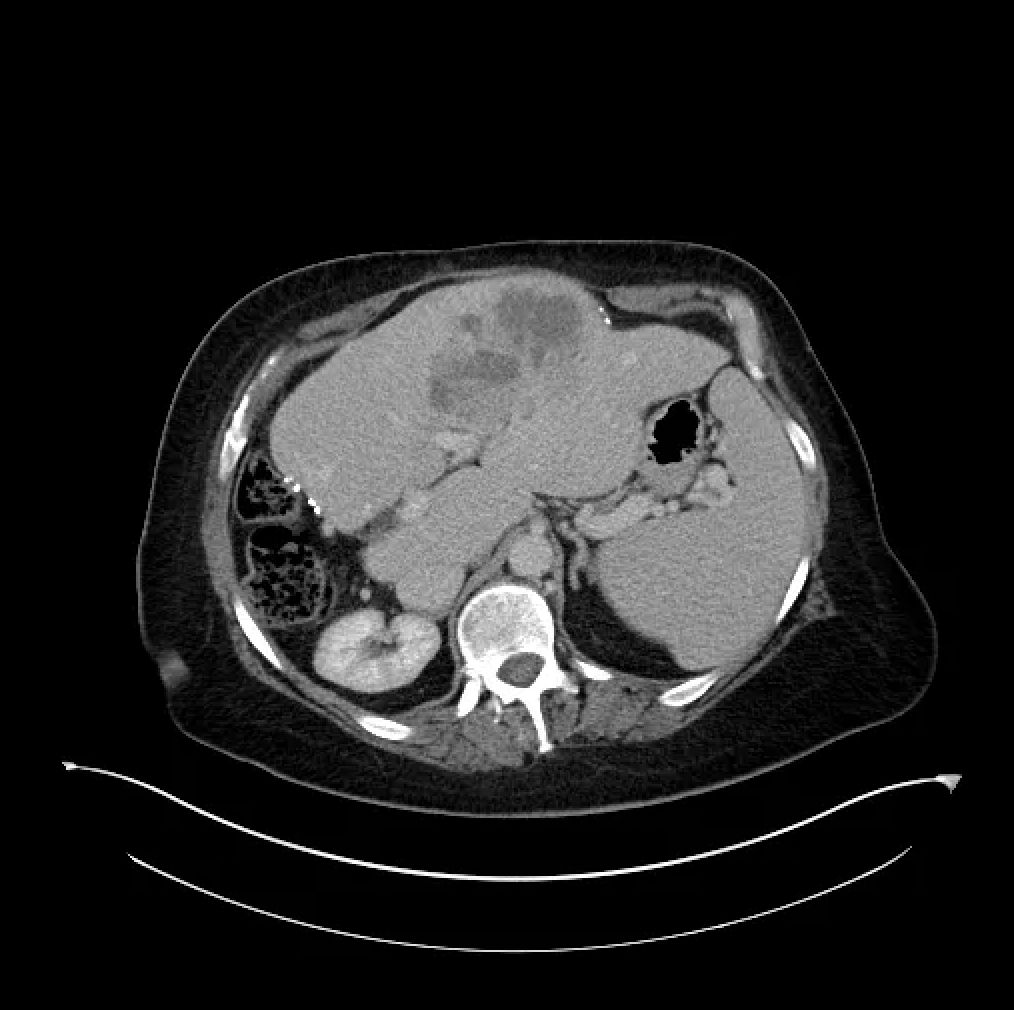} 
    
\end{wrapfigure}

Based on the tools used and the information received, here is a detailed summary of the new information regarding Mrs. Brown's condition and potential implications for her oncologic treatment:

\textbf{1. Radiology Report Analysis:}

- Two CT images were analyzed: CTA\_10\_02\_24.png and CTA\_01\_11\_23.png.

- \textbf{CTA\_10\_02\_24.png}: This image shows a cross-sectional view of the human abdomen, particularly focusing on the liver and kidneys. A notable difference in density within the liver suggests the \colorbox{mygreen}{presence of a lesion or abnormality}. The kidneys and other anatomical structures are visible, with the \colorbox{myred}{right} kidney partially obscured by the liver.

- \textbf{CTA\_01\_11\_23.png}: Similar to the first image, this scan shows the liver, kidneys, spine, and abdominal wall. However, the liver appears to have a more uniform density \colorbox{mygreen}{without the prominent lesion} or abnormality observed in the first image.

- \textbf{Comparison}: The most significant finding when comparing the two images is the appearance of a \colorbox{mygreen}{heterogeneous area within the liver in}  \colorbox{mygreen}{CTA\_10\_02\_24.png, which is not present in CTA\_01\_11\_23.png}. This could indicate a pathological change, possibly the development of a \colorbox{mygreen}{new liver metastasis}.

\textbf{2. Genetic Predictions from Histopathology Images:}

- \textbf{MSI (Microsatellite Instability)}: The prediction is \colorbox{mygreen}{MSS (Microsatellite} \colorbox{mygreen}{ Stable)} with a high probability of 0.97. This suggests that the cancer is not characterized by high levels of microsatellite instability, which can influence the response to certain types of immunotherapy.

- \textbf{BRAF Mutation}: The prediction is \colorbox{mygreen}{wild-type} with a very low probability of 0.03, indicating that a BRAF mutation is unlikely. This has implications for targeted therapy options, as BRAF mutations can sometimes be targeted with specific drugs.

- \textbf{KRAS Mutation}: The prediction is \colorbox{mygreen}{mutated} with a probability of 0.68. This suggests the \colorbox{mygreen}{presence of a KRAS mutation}, which is significant as it can impact the effectiveness of certain treatments, including EGFR inhibitors like Panitumumab.\\
\\
\\
\\
\\
\\
& \textbf{Implications for Oncologic Treatment:}
\\
& - The presence of a \colorbox{mygreen}{new lesion or abnormality in the liver}, as indicated by the radiology report, suggests that the disease may have \colorbox{mygreen}{progressed}. This needs to be considered when planning further treatment.\\

& - The genetic analysis indicates that the cancer is \colorbox{mygreen}{MSS and KRAS mutated,} \colorbox{mygreen}{with no BRAF mutation}. This genetic profile can influence the choice of chemotherapy and targeted therapy options. Specifically, the presence of a KRAS mutation may limit the effectiveness of EGFR inhibitors like Panitumumab.\\

& - Given these findings, a continuation of chemotherapy plus Panitumumab may need to be re-evaluated in light of the KRAS mutation. Other treatment options, possibly including other targeted therapies or immunotherapies, may need to be considered depending on the full clinical context and additional factors such as the patient's overall health, preferences, and other genetic markers not analyzed here. 
\\

& This summary provides a comprehensive overview of the new information obtained from the tools used, highlighting key findings and their potential implications for Mrs. Brown's oncologic treatment. \\

\bottomrule 

D & \vspace{-\baselineskip}
\begin{wrapfigure}{l}{0.45\textwidth} 
    \centering
    \includegraphics[width=0.2\textwidth]{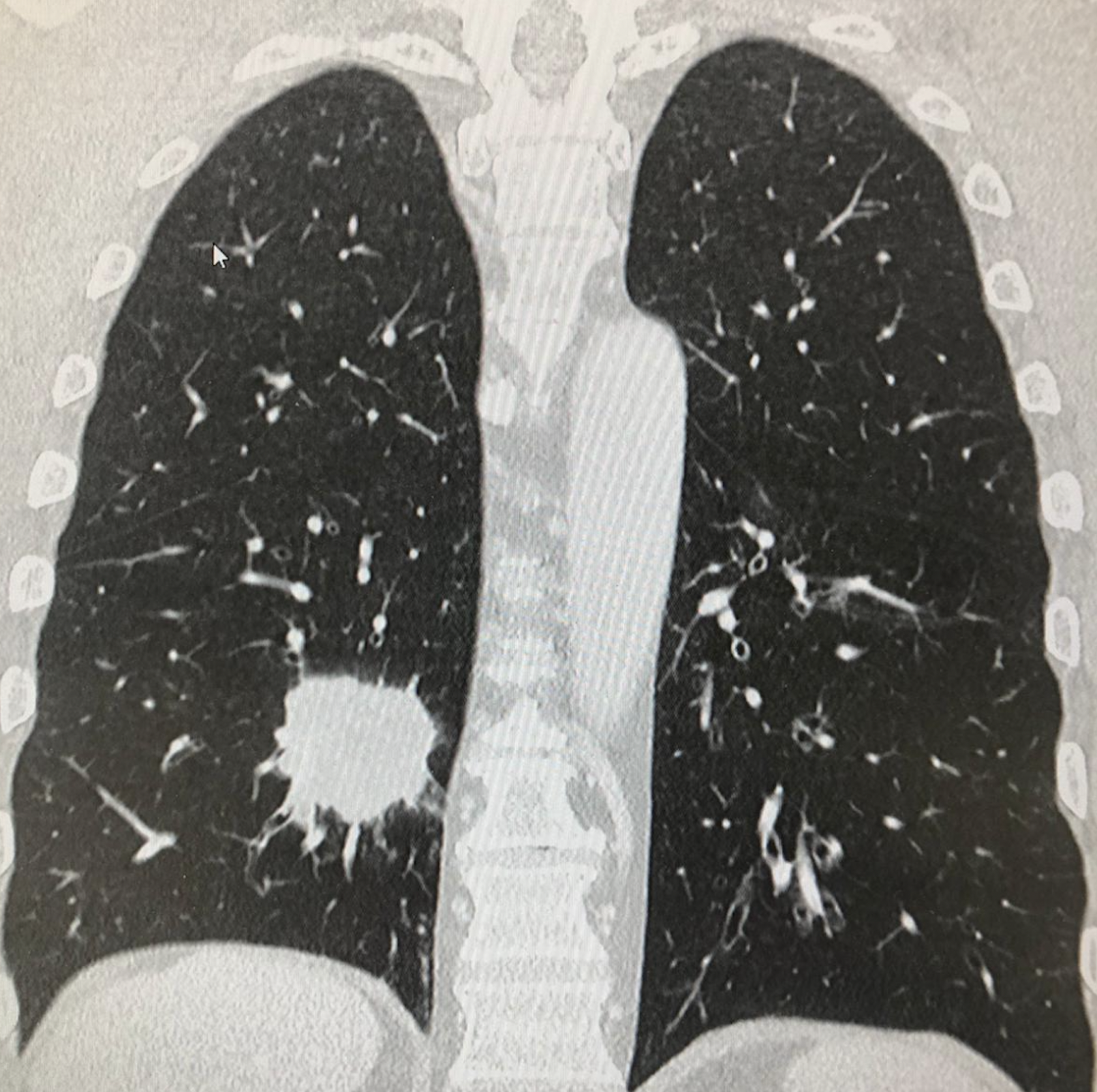} 
    \hspace{1pt} 
    \includegraphics[width=0.2\textwidth]{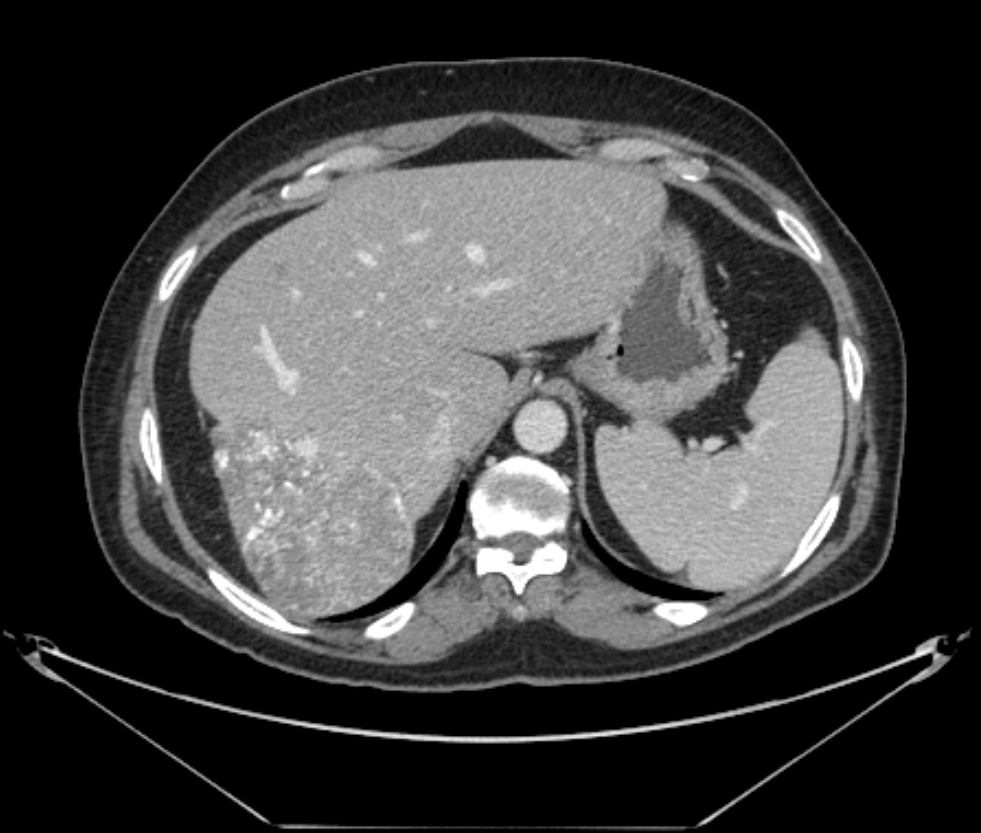} 
\end{wrapfigure}

The strategy to determine the suitability of Mr. Doe for surgery or systemic therapy, and to explore options for targeted therapies, involved the use of two tools: \textbf{gen\_radiology\_report} and \textbf{check\_mutations}. Here's a summary of the new information received from these tools and what they mean:

\textbf{Radiology Report Analysis:}

Two CT images, \textbf{CT\_Lung.png} and \textbf{CT\_Liver.png}, were analyzed to assess the extent of Mr. Doe's colorectal cancer.

\textbf{CT\_Lung.png} findings:

- \colorbox{myred}{Multiple} well-defined nodules scattered throughout both lung fields, suggestive of \colorbox{mygreen}{metastatic disease}.

- No significant consolidation, pleural effusion, or lymphadenopathy.

- The nodules are hyperdense, non-calcified, and without cavitation.

- The report concludes that these findings raise concern for \colorbox{mygreen}{metastatic disease} and recommends further evaluation with a tissue biopsy.

- \textbf{CT\_Liver.png} findings:

- \colorbox{myred}{Multiple} hypodense lesions in both lobes of the liver, consistent with hepatic metastases.

- The \colorbox{mygreen}{largest lesion in the right lobe has irregular margins}, indicating an aggressive growth pattern.

- No signs of cirrhosis, fatty infiltration, thrombosis, or encasement of vascular structures.

- No evidence of ascites or visible lymphadenopathy in the provided slice.

- The report concludes that the findings are concerning for metastatic spread of colorectal cancer and recommends a \colorbox{mygreen}{full review of the entire CT scan series} for a complete assessment.

\textbf{Genetic Predictions from Histopathology Images}
The \textbf{check\_mutations} tool provided predictions on three genetic targets: MSI, BRAF, and KRAS.

- \textbf{MSI (Microsatellite Instability)}: The prediction is \colorbox{mygreen}{MSI-High} with a probability of 0.95. This indicates a high likelihood of the cancer being MSI-High, which is relevant for determining suitability for certain immunotherapies.

- \textbf{BRAF}: The prediction is \colorbox{mygreen}{wild-type} with a probability of 0.48. This suggests that, \colorbox{mygreen}{contrary to Mr. Doe's recollection, the BRAF V600E} \colorbox{mygreen}{mutation might not be present.} \\

& - \textbf{KRAS}: The prediction is \colorbox{mygreen}{mutated} with a probability of 0.52. This indicates a slight likelihood of a KRAS mutation, which has implications for targeted therapy options.\\

& \textbf{Targeted Treatment Options for BRAF V600E} \\
& The \textbf{onko\_kb} tool provided results on BRAF V600E. \\

& - We received a comprehensive \colorbox{mygreen}{report on the BRAF V600E mutation}, highlighting its oncogenic nature, known effect as a gain-of-function mutation, and its role in activating the downstream MAPK pathway independent of RAS. It also provided information on the sensitivity of this mutation to RAF monomer inhibitors such as vemurafenib and dabrafenib. The report detailed various treatments, including dabrafenib alone or in combination with trametinib, and their FDA-approved indications for different cancer types. Notably, it mentioned the combination of encorafenib and cetuximab as FDA-approved for the treatment of adult patients with metastatic colorectal cancer with a BRAF V600E mutation after prior therapy, based on the Phase III BEACON study. \\

& \textbf{\colorbox{mygreen}{Web Search} for BRAF V600E treatment options} \\
& - The tool returned information on several drugs and treatment regimens available for colorectal cancer with the BRAF V600E mutation. These included Encorafenib in combination with Cetuximab, a targeted drug trio (Encorafenib, Cetuximab, and Binimetinib), Pembrolizumab for MSI mCRC harboring BRAFV600E mutation, and the feasibility of BRAF inhibitors in combination with MEK inhibitors to overcome drug resistance. It also mentioned ongoing trials and recent approvals for anti-BRAF/EGFR therapy in metastatic BRAFV600E colorectal cancer. \\

& \textbf{Implications} \\
& - The presence of metastatic disease in both the lungs and liver significantly impacts the treatment strategy, potentially making Mr. Doe less suitable for surgery and more reliant on systemic therapy. \\
& - The genetic predictions suggest that Mr. Doe's cancer is \colorbox{mygreen}{MSI-High and} \colorbox{mygreen}{ has a KRAS mutation}, which opens up possibilities for targeted therapies and immunotherapies, particularly given his refusal to undergo chemotherapy. \\
& - The discrepancy regarding the BRAF mutation status highlights the importance of confirming genetic alterations through reliable molecular testing, as they directly influence the selection of targeted therapies. \\

& This comprehensive analysis combining radiological and genetic data provides a clearer picture of Mr. Doe's disease status and the potential therapeutic avenues, emphasizing the need for a multidisciplinary approach to his treatment planning. \\

\bottomrule 

G & \vspace{-\baselineskip}
\begin{wrapfigure}{l}{0.45\textwidth} 
    \centering
    \includegraphics[width=0.2\textwidth]{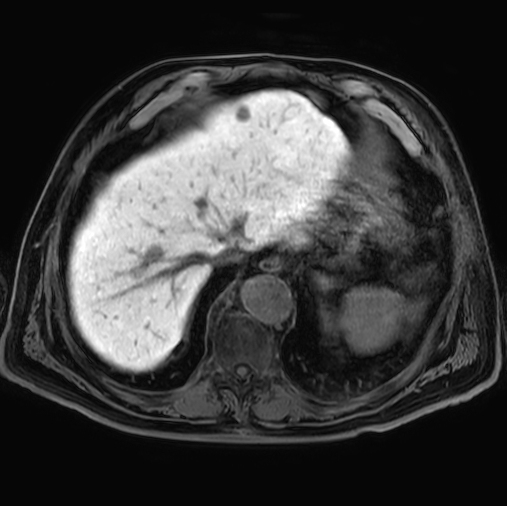} 
    \hspace{1pt} 
    \includegraphics[width=0.2\textwidth]{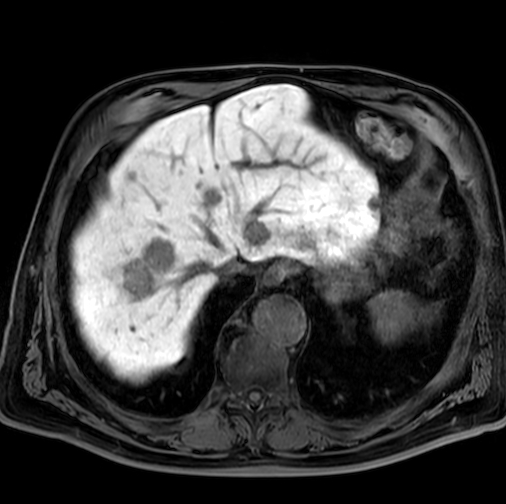} 
\end{wrapfigure}

To develop a strategy for the best treatment approach for Mr. Garcia, who has \colorbox{mygreen}{metastatic colorectal cancer}, the following steps were taken:

\textbf{1. Segmentation of Hepatic Lesions in Imaging Data:} Utilizing the `segment\_image` function, the areas of hepatic lesions identified in the MRI scans were measured for both the initial scan taken on November 8, 2023 ('08\_11\_23.png'), and the recent control imaging conducted on February 10, 2024 ('02\_10\_24.png'). The bounding box coordinates for the lesions were provided based on the radiology reports.

- For the initial scan ('08\_11\_23.png'), the \colorbox{mygreen}{overall area of the regions} \colorbox{mygreen}{of interest (three lesions) was calculated to be \textbf{1315} square units.}

- For the recent control scan ('02\_10\_24.png'), the \colorbox{mygreen}{overall area of the regions} \colorbox{mygreen}{of interest (four lesions) was calculated to be \textbf{2809} square units.}

\textbf{2. Calculation of Lesion Growth Ratio:} Using the `calculate` function, the \colorbox{mygreen}{ratio of the lesion areas between the recent control imaging and the initial scan} \colorbox{mygreen}{was determined to be approximately \textbf{2.14}}. This indicates that the overall area of the hepatic lesions has more than doubled over the period between the two scans.

\textbf{Summary of New Information Received}:

- The segmentation of hepatic lesions from the MRI scans provided precise measurements of the lesion areas at two different time points.

- The calculation of the lesion growth ratio revealed a significant \colorbox{mygreen}{increase in the overall area of the hepatic lesions}, suggesting progression of the metastatic disease.

This information is crucial for understanding the progression of Mr. Garcia's metastatic colorectal cancer and will inform the next steps in developing a treatment strategy, including the consideration of additional diagnostic tests, genetic profiling, and potential treatment options. \\

\bottomrule 

Le & \vspace{-\baselineskip}
\begin{wrapfigure}{l}{0.45\textwidth} 
    \centering
    \includegraphics[width=0.2\textwidth]{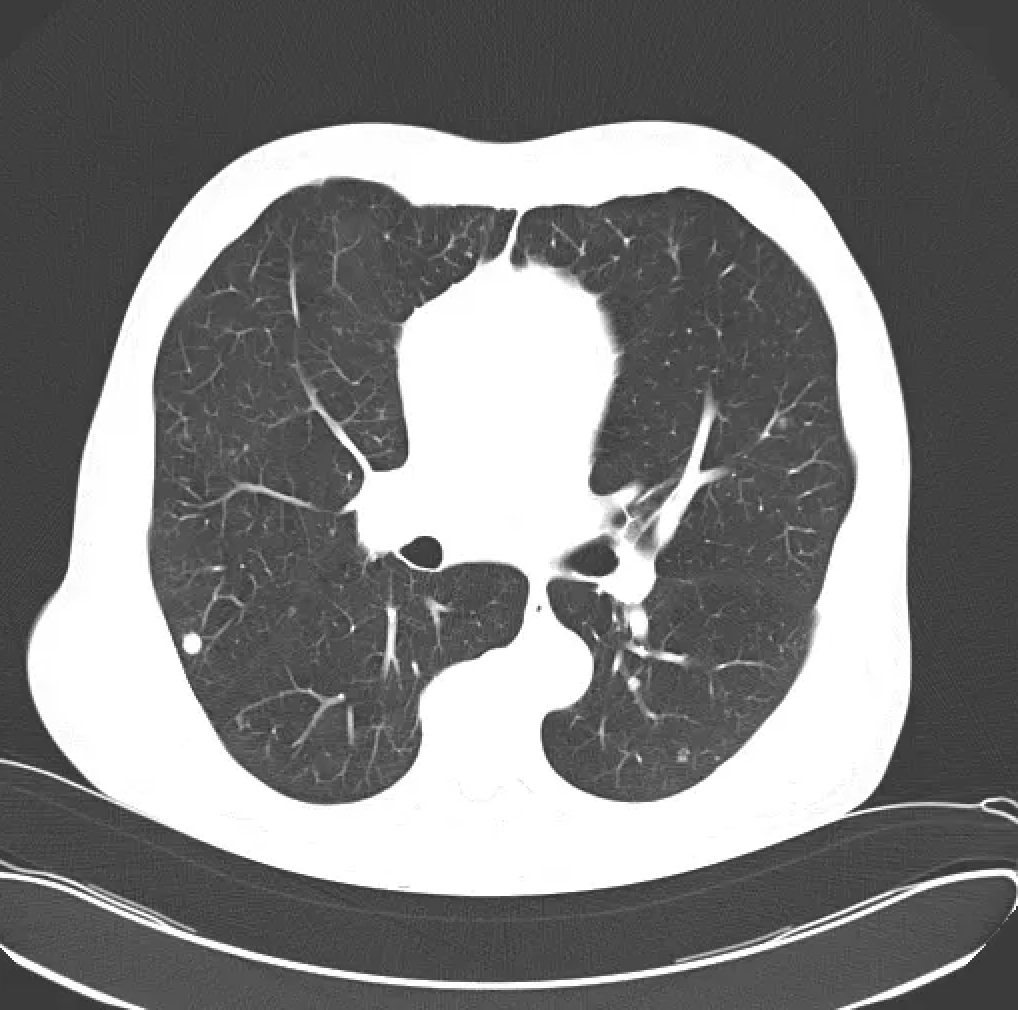} 
    \hspace{1pt} 
    \includegraphics[width=0.2\textwidth]{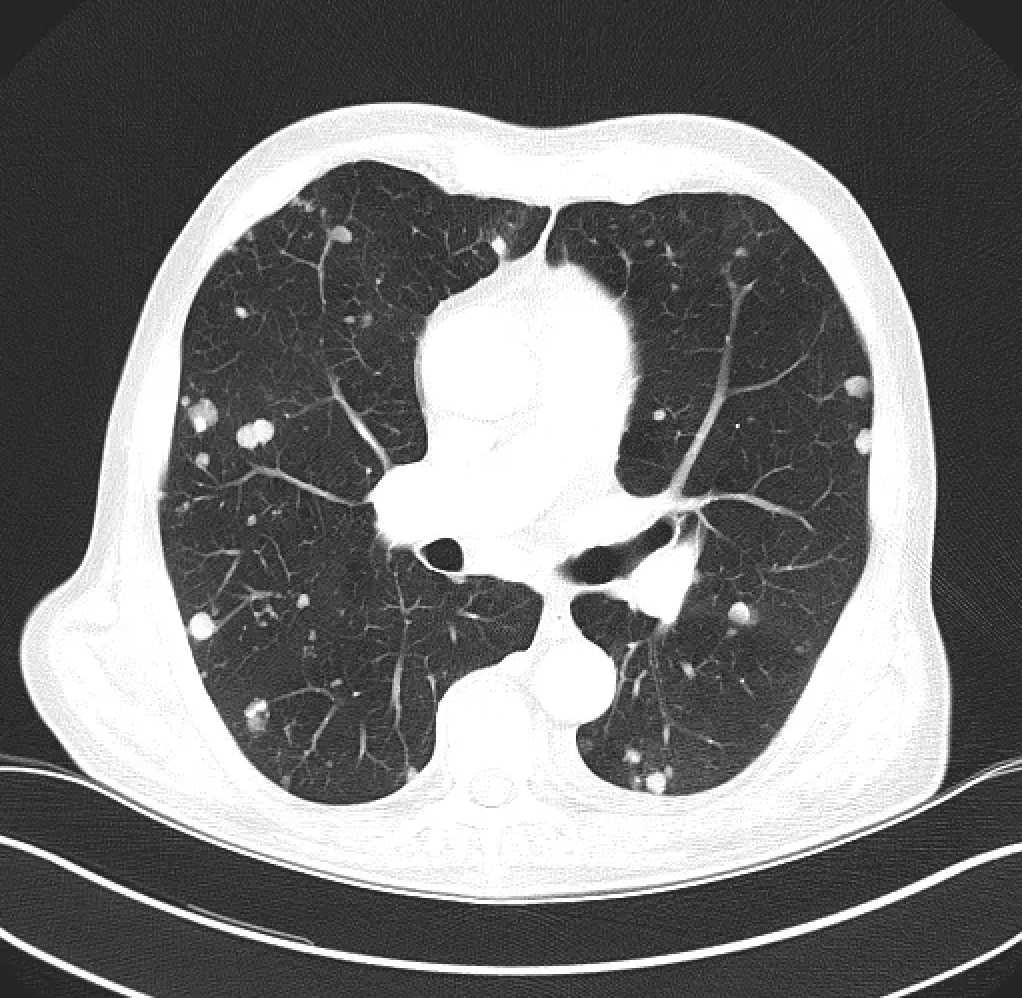} 
\end{wrapfigure}

Based on the tools used and the information received, here is a detailed summary of the new information:


\textbf{1. Radiology Report Analysis:}

- Two CT scans were evaluated: 'CT\_Lung\_Feb\_2024.png' and 'CT\_Lung\_Nov2023.png'.

- The report for 'CT\_Lung\_Nov2023.png' described normal thoracic anatomy \colorbox{myred}{without obvious abnormalities in the lung fields}, mediastinum, or bony structures.

- The comparison between the two scans showed that the \colorbox{mygreen}{'CT\_Lung\_Feb\_2024.png' image had more and denser opacities throughout the} \colorbox{mygreen}{lung fields compared to the 'CT\_Lung\_Nov2023.png' image}, suggesting a more advanced or different pathological process in the February 2024 scan.

- The findings indicate a \colorbox{mygreen}{progression in the lung nodules from November} \colorbox{mygreen}{2023 to February 2024}, which could be consistent with a recurrence or progression of the pancreatic cancer.

\textbf{2. \colorbox{mygreen}{Google Search on KRAS G12C Mutation} Treatment in 2023:}
- The search revealed significant advancements in the treatment of KRAS G12C mutation in cancer, with the development and approval of targeted therapies.

- Key findings include:

- \textbf{\colorbox{mygreen}{Sotorasib}} has been compared to Docetaxel in patients with non-small-cell lung cancer (NSCLC) harboring the KRAS G12C mutation, showing promising results.

- \textbf{\colorbox{mygreen}{Divarasib (GDC-6036)}} reported durable clinical responses across KRAS G12C-positive tumors with mostly low-grade adverse events.

- \textbf{\colorbox{mygreen}{Adagrasib}}, alone or in combination with Cetuximab, evaluated in colorectal cancer patients with the KRAS G12C mutation, showing efficacy.

- These studies and publications highlight the progress in the development of targeted therapies for cancers with the KRAS G12C mutation, offering new hope for patients with this genetic alteration.

\textbf{Conclusion:}
The radiology report comparison suggests a \colorbox{mygreen}{progression} in Mr. Lee's \colorbox{mygreen}{lung nodules}, which could be indicative of a recurrence or progression of his pancreatic cancer. The Google search on KRAS G12C mutation treatment in 2023 provided valuable insights into the latest advancements in targeted therapies for this mutation, including \colorbox{mygreen}{Sotorasib, Divarasib, and Adagrasib}, which could be considered in the treatment strategy for Mr. Lee's condition.

\\ \bottomrule 

Lo & \vspace{-\baselineskip}
\begin{wrapfigure}{l}{0.2\textwidth} 
    \centering
    \includegraphics[width=0.2\textwidth]{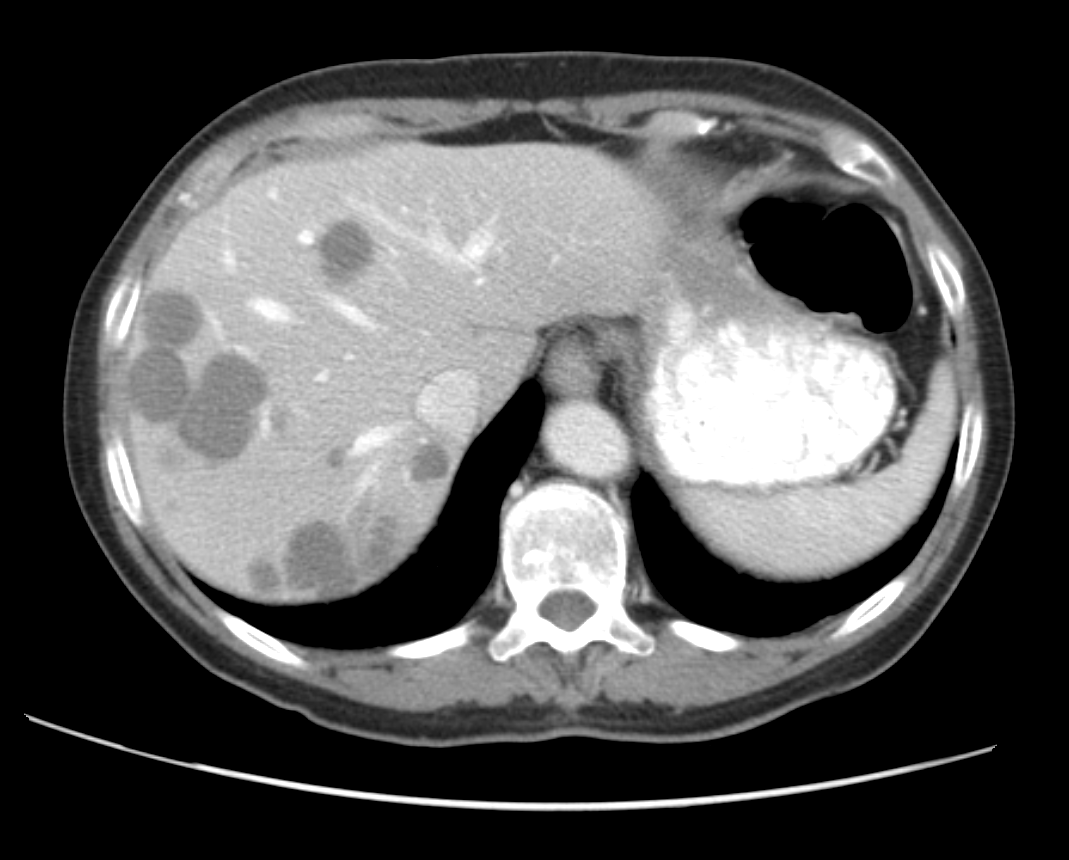} 
\end{wrapfigure}

Based on the tools used and the information received, we have gathered significant insights into Mr. Lopez's condition:
\textbf{1. Radiology Report Analysis (CT Scan):}

- \textbf{Tool Used}: `functions.gen\_radiology\_report`

- \textbf{Input}: Path to the CT scan folder "Imaging/Lopez" with a query focusing on the extent of the primary tumor, lymph node involvement, and signs of metastasis, especially in the liver.

- \textbf{Output}: The CT scan (staging\_CTA.png) showed \colorbox{mygreen}{multiple hypodense} \colorbox{mygreen}{areas in the liver}, which could indicate cysts, \colorbox{mygreen}{lesions}, or areas of lower density compared to the normal liver parenchyma. Other organs such as the spleen, \colorbox{myred}{kidneys}, stomach, spine, abdominal aorta, and bowel loops were visualized and appeared normal. The presence of multiple hypodense areas in the liver is of clinical significance and warrants further investigation.

- \textbf{Interpretation}: The findings \colorbox{mygreen}{suggest potential liver involvement,} \colorbox{mygreen}{which could indicate metastasis}. This is crucial for staging the cancer and determining the appropriate treatment strategy.

\textbf{2. Genetic Mutation Analysis}:

- \textbf{Tool Used}: `functions.check\_mutations`

- \textbf{Input}: Patient ID "AA-21-C" with targets for MSI, BRAF, and KRAS mutations.

- \textbf{Output}: The genetic analysis revealed that the patient is \colorbox{mygreen}{MSS (Microsatellite Stable)} with a high probability of 0.97, indicating that the cancer is not MSI-high which often has implications for certain treatment options. Both \colorbox{mygreen}{BRAF and KRAS genes are wild-type} with probabilities of 0.02 and 0.27, respectively.

- \textbf{Interpretation}: The absence of mutations in BRAF and KRAS genes suggests that treatments specifically targeting these mutations may not be applicable. The MSS status also influences the treatment approach, as MSI-high status can sometimes make patients eligible for immunotherapy.

\textbf{Summary}:
The combination of the CT scan findings and genetic mutation analysis provides a comprehensive overview of Mr. Lopez's condition. The presence of \colorbox{mygreen}{multiple hypodense areas in the liver suggests} \colorbox{mygreen}{potential metastasis}, which is critical for staging the cancer and could influence the decision between local therapies and systemic treatments. The genetic analysis indicates that the cancer is \colorbox{mygreen}{MSS and lacks mutations in BRAF and KRAS}, which guides us away from certain targeted therapies. This information is essential for developing a tailored treatment strategy for Mr. Lopez, considering both the extent of the disease and the molecular characteristics of the tumor. Further investigations and multidisciplinary discussions are necessary to finalize the treatment plan.

\\ \bottomrule 

S & \vspace{-\baselineskip}
\begin{wrapfigure}{l}{0.45\textwidth} 
    \centering
    \includegraphics[width=0.2\textwidth]{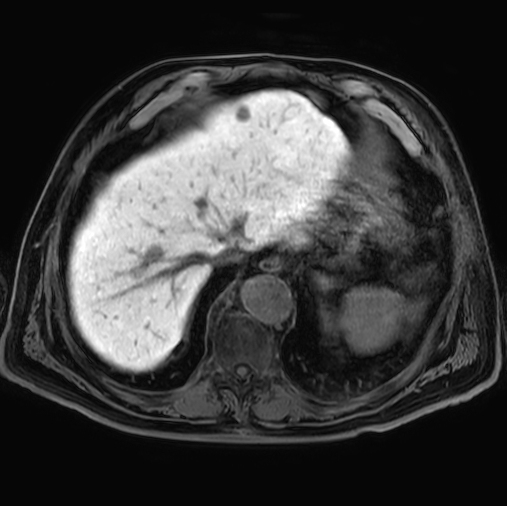} 
    \hspace{1pt} 
    \includegraphics[width=0.2\textwidth]{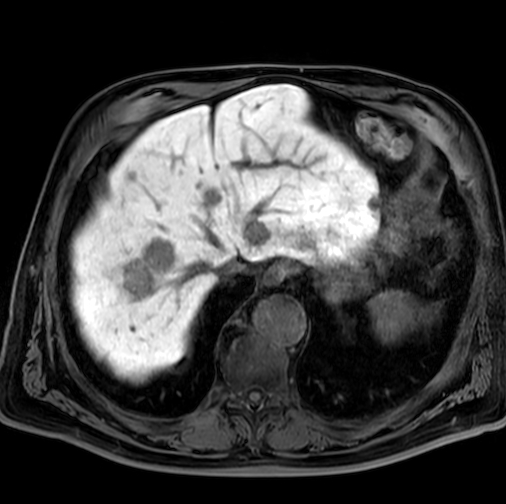} 
\end{wrapfigure}

Based on the utilization of multiple tools, we have gathered new information regarding Mrs. Smith's current health status, particularly focusing on her liver metastases and genetic mutations.
\textbf{1. Radiology Report Analysis:}

- The comparison between two MRI scans, one from February 2024 (02\_10\_24.png) and the other from November 2023 (08\_11\_23.png), revealed the following:

- Both images show a similar cross-section of the human abdomen.

- A \colorbox{myred}{notable bright area on the left side of the images, indicative of} \colorbox{myred}{a region of high signal intensity}, was observed. This area was \colorbox{myred}{more pronounced in the February 2024 image compared to the November} \colorbox{myred}{2023 image.}

- The contrast between different tissues was \colorbox{myred}{slightly more distinct in the} \colorbox{myred}{November 2023 image}.

- The decrease in intensity of the bright area from February 2024 to November 2023 might suggest a change in the liver metastases, but without further clinical context, a definitive assessment cannot be made.

\textbf{2. Genetic Mutation Analysis:}

- The genetic predictions from the histopathology images of patient B13-22A (Mrs. Smith) indicated:

- \textbf{KRAS Mutation}: Detected with a probability of 0.52. This suggests a \colorbox{mygreen}{mutation in the KRAS gene}, which is a significant finding given that Mrs. Smith was previously reported as KRAS wild type in May 2023.

- \textbf{BRAF Status}: Confirmed as \colorbox{mygreen}{wild type} with a high confidence (probability of 0.03), consistent with the initial genetic profiling.

- \textbf{MSI Status}: Predicted as \colorbox{mygreen}{MSS (Microsatellite Stable)} with a high probability of 0.96, aligning with the initial diagnosis.

\textbf{Summary}:

The radiology report comparison suggests a \colorbox{mygreen}{potential change in the liver} \colorbox{mygreen}{metastases,} \colorbox{myred}{indicated by the variation in signal intensity between the two} \colorbox{myred}{MRI scans}. However, without additional clinical context, the significance of this change remains uncertain. The genetic mutation analysis revealed a \colorbox{mygreen}{new finding of a KRAS mutation}, which differs from the initial genetic profiling that identified Mrs. Smith as KRAS wild type. The \colorbox{mygreen}{BRAF wild type and MSS status} remain consistent with the initial diagnosis. This new information, particularly the KRAS mutation, could have implications for Mrs. Smith's treatment plan and may warrant a reevaluation of the current regimen of 5-Fluorouracil and Panitumumab.\\

\bottomrule 

T & \vspace{-\baselineskip}
\begin{wrapfigure}{l}{0.25\textwidth} 
    \centering
    \includegraphics[width=0.2\textwidth]{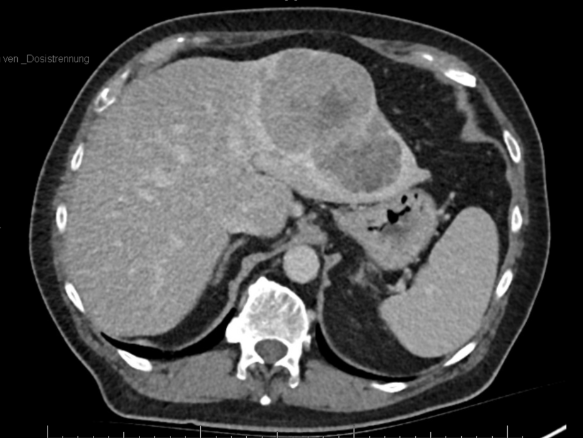} 
\end{wrapfigure}

The radiology report generated from the CT scan image labeled "Staging\_03\_07\_2024.png" for Mr. Taylor provides several key findings relevant to his current clinical status and the \colorbox{mygreen}{progression} of his hepatocellular carcinoma (HCC) in the context of portal vein thrombosis (PVT) and underlying liver cirrhosis:

\textbf{1. Liver Morphology and Lesions:} The liver shows irregular contours and heterogeneous density, consistent with cirrhosis and previous interventions. \colorbox{myred}{Multiple} \colorbox{mygreen}{hypodense areas} within the liver parenchyma are observed, suggestive of multifocal HCC. The report emphasizes the need for comparison with previous imaging to assess for progression.

\textbf{2. Portal Vein Thrombosis:} \colorbox{myred}{A filling defect within the portal vein} is noted, aligning with the known PVT. The report suggests further evaluation of the thrombosis's extent and its impact on portal circulation.

\textbf{3. Vascular Structures}: The aorta and inferior vena cava appear patent without obvious abnormalities.

\textbf{4. Adjacent Organs and Ascites:} The \colorbox{myred}{spleen is enlarged}, likely secondary to portal hypertension. \colorbox{myred}{Ascites} is also present, indicating a common complication of liver cirrhosis and portal hypertension.

\textbf{5. Conclusion:} The CT image shows features consistent with cirrhotic liver, \colorbox{myred}{multifocal} \colorbox{mygreen}{HCC}, \colorbox{myred}{portal vein thrombosis, ascites, and splenomegaly} due to portal hypertension. The report recommends a comprehensive comparison with previous imaging to determine the progression of HCC and the status of portal vein thrombosis, as well as to identify any new findings that could influence the management of Mr. Taylor's disease.

This detailed radiology report provides crucial information for reassessing therapeutic strategies for Mr. Taylor, highlighting the importance of comparing current findings with previous imaging to understand the disease's progression and adjust the management plan accordingly.

\\ \bottomrule 

W & \vspace{-\baselineskip}
\begin{wrapfigure}{l}{0.45\textwidth} 
    \centering
    \includegraphics[width=0.2\textwidth]{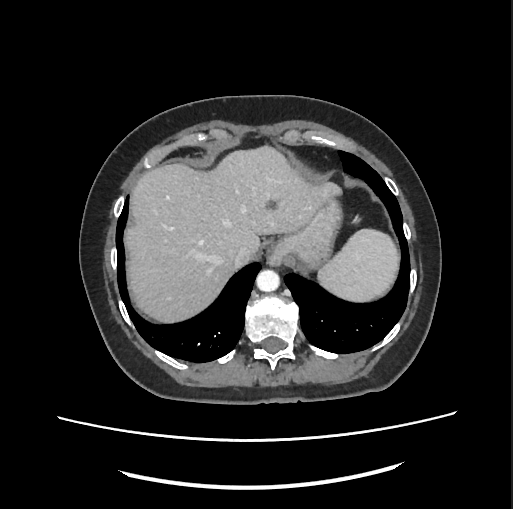} 
    \hspace{1pt} 
    \includegraphics[width=0.2\textwidth]{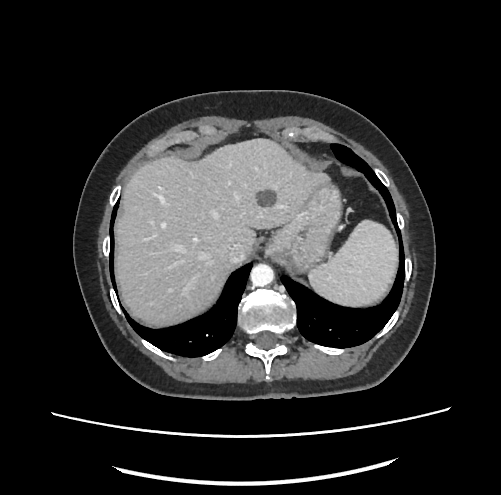} 
\end{wrapfigure}

Here's a summary of the new information we received from the tools used:
\textbf{1. Segmentation of Liver Metastasis:}

- The area of the liver metastasis in the November 2023 CT scan (Imaging/Williams/CTA\_11\_2023.png) was measured to be \colorbox{mygreen}{204 square units.}
   
- The area of the liver metastasis in the February 2024 CT scan (Imaging/Williams/CTA\_02\_2024.png) was measured to be \colorbox{mygreen}{367 square units.}
   
- The \colorbox{mygreen}{ratio between the February 2024 and November 2023 areas is } \colorbox{mygreen}{approximately 1.799}, indicating an increase in the size of the metastasis.
\textbf{2. Genetic Predictions:}

- The genetic test revealed the tumor is \colorbox{mygreen}{MSI-High} with a probability of 0.68.
   
- \colorbox{mygreen}{BRAF mutation was confirmed} with a probability of 0.7.
   
- The tumor is \colorbox{mygreen}{KRAS wild-type} with a probability of 0.44.

\textbf{3. \colorbox{mygreen}{OncoKB Database Information:}}

- For the BRAF V600E mutation, it is described as oncogenic with a known effect of "Gain-of-function". It activates the downstream MAPK pathway independent of RAS and retains sensitivity to RAF monomer inhibitors such as \colorbox{mygreen}{vemurafenib and dabrafenib}. There are several FDA-approved treatments for BRAF V600E mutations in different cancers, including dabrafenib and \colorbox{mygreen}{trametinib} for melanoma, lung cancer, and others. \colorbox{mygreen}{Encorafenib and cetuximab} are FDA-approved for metastatic colorectal cancer with a BRAF V600E mutation after prior therapy.
\textbf{Interpretation:}

- The increase in the size of the liver metastasis under chemotherapy indicates progressive disease rather than a response to treatment.

- The described BRAF mutation was confirmed.

- The presence of a BRAF V600E mutation and MSI-High status suggests potential targeted therapy options. Specifically, the BRAF V600E mutation opens up several treatment avenues with FDA-approved drugs such as dabrafenib, trametinib, encorafenib, and cetuximab.

- The MSI-High status, although not yielding specific treatment options from OncoKB\footnotemark, is generally associated with a better response to immunotherapy in colorectal cancer.
\\
\\
\\
\\
& \textbf{Next Steps:} \\

& - Considering the progressive disease and the genetic makeup of the tumor, a discussion about switching to a targeted therapy regimen that includes BRAF inhibitors (e.g., dabrafenib or encorafenib) and possibly combining with an EGFR inhibitor (e.g., cetuximab) or a MEK inhibitor (e.g., trametinib) could be warranted. \\

& - Further exploration of immunotherapy options, given the MSI-High status, could also be considered. \\

\\ \bottomrule 

X & \vspace{-\baselineskip}
\begin{wrapfigure}{l}{0.45\textwidth} 
    \centering
    \includegraphics[width=0.2\textwidth]{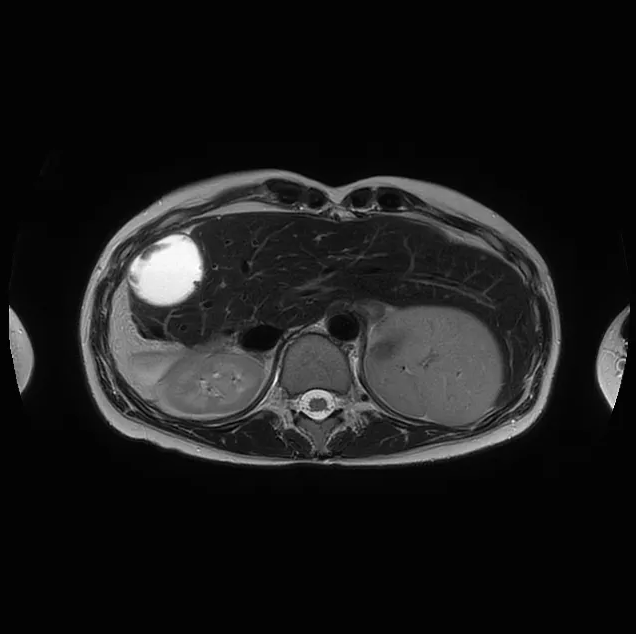} 
    \hspace{1pt} 
    \includegraphics[width=0.2\textwidth]{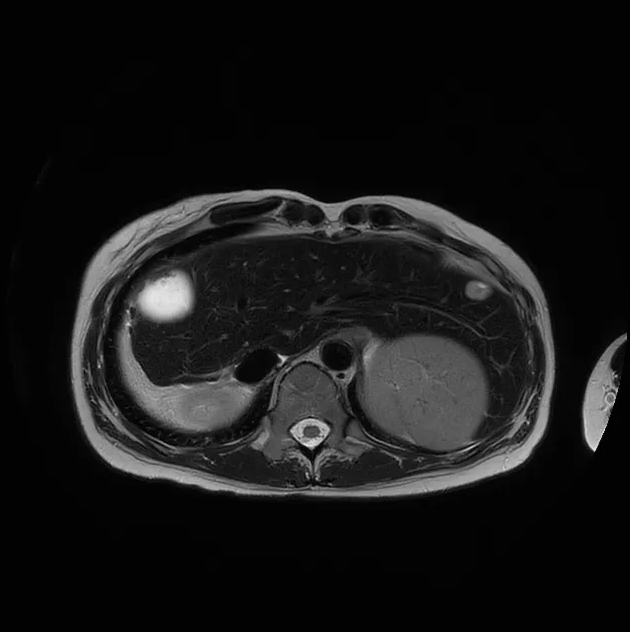} 
\end{wrapfigure}

The strategy to investigate the current state of Ms. Xing's disease and explore targeted treatment options involved multiple steps, utilizing various tools to gather comprehensive information. Here's a summary of the new information received from the tools used, along with their implications:
\textbf{1. Segmentation of Metastasis in Imaging:}

   - \textbf{Tool Used}: `segment\_image` for both September 2023 and February 2024 images.
   
   - \textbf{Inputs}: September 2023 image with coordinates [475, 250, 490, 275] and February 2024 image with coordinates [455, 270, 505, 320].
   
   - \textbf{Outputs}: The \colorbox{mygreen}{area of the metastasis in September 2023 was 133, and in} \colorbox{mygreen}{February 2024, it was 518.}
   
   - \textbf{Implication}: The \colorbox{mygreen}{significant increase in the area of the metastasis} \colorbox{mygreen}{(approximately 3.89 times larger)} indicates disease progression.

\textbf{2. Investigation of Targeted Treatment Options for BRAF V600E Mutation:}

   - \textbf{Tool Used}: `onco\_kb`.
   
   - \textbf{Inputs}: Query for BRAF V600E mutation.
   
   - \textbf{Outputs}: Identified treatments including \colorbox{mygreen}{Dabrafenib and Trametinib}, with FDA approval for BRAF V600E mutant unresectable or metastatic melanoma, and other solid tumors excluding colorectal cancer. Also noted was the \colorbox{mygreen}{effectiveness of these drugs in biliary tract cancer.}
   
   - \textbf{Implication}: Given Ms. Xing's BRAF V600E mutation, Dabrafenib and Trametinib could be considered for her treatment, noting their approval and effectiveness in similar cases.

\textbf{3. Investigation of Targeted Treatment Options for CD74-ROS1 Variant Fusion:}

   - \textbf{Tool Used}: \colorbox{mygreen}{`onco\_kb`.}
   
   - \textbf{Inputs}: Query for CD74-ROS1 variant fusion.
   
   - \textbf{Outputs}: Identified \colorbox{mygreen}{treatments for ROS1-positive tumors,} \colorbox{mygreen}{including Crizotinib, Entrectinib, and Repotrectinib}, with FDA approval for metastatic non-small cell lung cancer (NSCLC) whose tumors are ROS1-positive. \colorbox{mygreen}{Ceritinib and Lorlatinib} were also mentioned as treatment options.
   
   - \textbf{Implication}: Although the CD74-ROS1 fusion is primarily associated with NSCLC, the identified treatments could be explored for their potential applicability in Ms. Xing's case, given the oncogenic nature of this fusion.
\\
& \textbf{4. Google Search for Targeted Treatment Options:}  \\

   & - \textbf{Tool Used}: \colorbox{mygreen}{`google\_search`}  \\
   
   & - \textbf{Inputs}: Query for targeted treatment options for cholangiocellular cancer with BRAF V600E mutation and CD74-ROS1 fusion\\
   
   & - \textbf{Output}: The Google search provided insights into potential targeted treatment options for cholangiocellular cancer with BRAF V600E mutation and CD74-ROS1 fusion. It highlighted the use of \colorbox{mygreen}{encorafenib and binimetinib, dabrafenib,} \colorbox{mygreen}{and the combination of dabrafenib and trametinib} as potential options based on their effectiveness in other cancers with similar genetic alterations.  \\
\\
& \textbf{5. PubMed Query for Latest Targeted Treatment Options:} \\

& - \textbf{Tool Used:} \textbf{`query\_pubmed`}\\
   
& - \textbf{Inputs:} Query for 'targeted treatment options for cholangiocellular cancer with BRAF V600E mutation and CD74-ROS1 fusion?'\\
   
& - \textbf{Outputs:} The \colorbox{myred}{PubMed search did not yield specific results} for targeted treatment options for cholangiocellular cancer with BRAF V600E mutation and CD74-ROS1 fusion. However, it emphasized the importance of targeting specific gene fusions in cancer treatment and mentioned the potential of NTRK gene fusions as novel targets, suggesting the relevance of molecular assays in identifying specific gene fusions for targeted therapy. \\
\\
& \textbf{6. Genetic Predictions from Histopathology Images:} \\

& - \textbf{Tool Used}: `check\_mutations` \\

& - \textbf{Inputs}: Patient ID "X-09.22" and targets for MSI, KRAS, BRAF. \\
   
& - \textbf{Microsatellite Instability (MSI)}: The prediction indicated \colorbox{mygreen}{Microsatellite Stability (MSS)} with a high probability of 0.89, suggesting that the tumor is not characterized by a high level of microsatellite instability. \\
   
& - \textbf{BRAF Mutation}: The analysis predicted a \colorbox{mygreen}{BRAF mutation} with a probability of 0.74. This is consistent with the molecular report summary provided, which identified a mutation in BRAF V600E. \\
   
& - \textbf{KRAS Mutation}: The prediction indicated a \colorbox{mygreen}{wild-type KRAS} with a low probability of 0.29, suggesting that no KRAS mutation is present in the tumor. \\

& This comprehensive approach, utilizing segmentation to quantify disease progression and querying the OncoKB database for targeted treatment options, provides a solid foundation for discussing Ms. Xing's treatment in the molecular tumor board. The \colorbox{mygreen}{significant growth of the metastasis} underscores the urgency of considering advanced targeted therapies, specifically those targeting the BRAF V600E mutation and CD74-ROS1 variant fusion, to potentially manage the disease progression as also shown by the provided web search results. The genetic predictions confirmed the presence of a \colorbox{mygreen}{BRAF mutation and showed the absence of a KRAS mutation} \colorbox{mygreen}{and MSS status}, which is \colorbox{mygreen}{contradictory to the described finding of} \colorbox{mygreen}{microsatellite instability.} \\

\bottomrule 

Z & \vspace{-\baselineskip}
\begin{wrapfigure}{l}{0.25\textwidth} 
    \centering \footnotemark
    \includegraphics[width=0.2\textwidth]{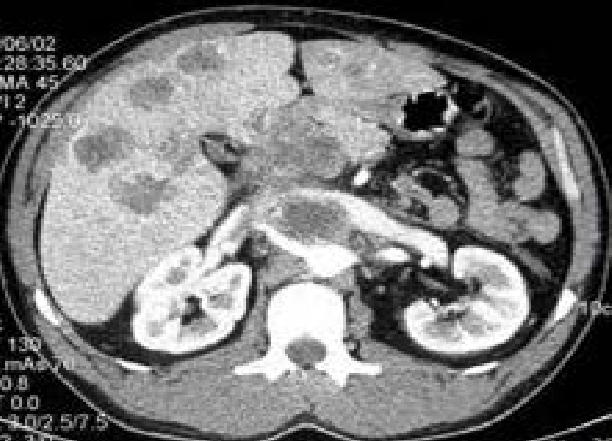} 
\end{wrapfigure}

Based on the tools utilized and the information received, here is a detailed summary of the new information:
\textbf{1. Radiology Report Generation:}

- \textbf{Input}: Requested a detailed radiology report for Mr. Zu, a 55-year-old patient with histologically confirmed pancreatic adenocarcinoma (T2N1MX), who underwent an iv-contrast enhanced CT-scan of the abdomen for staging. The query specifically asked for the size of the primary tumor, presence of any metastases, and condition of surrounding organs.

- \textbf{Output}: Radiology Report for staging\_Feb\_24.png: The image provided appears to be a CT scan of the abdominal region of a patient. The scan is an axial slice which shows various organs including the liver, kidneys, spleen, and portions of the gastrointestinal tract.

- The \colorbox{mygreen}{liver has multiple hypodense areas which could indicate the presence of} \colorbox{mygreen}{ lesions} or cysts. The exact nature of these would require clinical correlation and possibly further imaging or biopsy.

- The kidneys are visible on both sides of the spine and seem to be of normal size and shape. There does not appear to be any obvious abnormality with the kidneys from this single slice.

- The \colorbox{myred}{spleen} is visible on the left side of the image (which corresponds to the patient's left side) and appears to be of normal size and density.

- There are no obvious signs of free air or fluid in the abdominal cavity that would suggest perforation or acute abdominal pathology.

- The aorta and other vascular structures appear to be normal without evidence of aneurysm or dissection in this slice.

- Bowel loops are visible and appear to be normal in caliber without evidence of obstruction or significant wall thickening.

- The vertebral body and spinal canal appear normal without evidence of fracture or pathology.

- \textbf{Conclusion}: The most significant finding in this CT scan is the \colorbox{mygreen}{presence of multiple hypodense areas within the liver}. These findings warrant further investigation to determine their nature. The rest of the visible abdominal organs and structures appear to be within normal limits based on this single image. However, it is important to note that a full evaluation would require a review of the entire CT scan series, as well as clinical information and laboratory results for accurate diagnosis and staging.

\textbf{2. Mutation Check:}

- \textbf{Input}: Requested a check for specific genetic mutations (MSI, BRAF, KRAS) for patient Mr. Zu.

- \textbf{Output}: The \colorbox{myred}{tool encountered an error} and did not provide any results for the mutation check.

\\
& \textbf{Summary:} The radiology report suggests the presence of \colorbox{mygreen}{multiple lesions}\colorbox{mygreen}{in the liver}, which could potentially indicate metastases, given the patient's diagnosis of pancreatic adenocarcinoma. However, the report also highlights the \colorbox{mygreen}{limitation of making a definitive diagnosis based on a single CT slice} \colorbox{mygreen}{and recommends a full review of the entire CT scan series for a comprehensive} \colorbox{mygreen}{assessment}. Unfortunately, the attempt to check for specific genetic mutations encountered an error, and no information regarding the genetic profile of the tumor was obtained. This information would have been valuable for determining targeted therapy options. \\

\bottomrule

\footnotetext[1]{For our experiments we used the free, non-authorized version of the OncoKB API which is restricted to BRAF, TP53 and ROS1 as targets, thus causing a lack of results.}

\footnotetext[2]{This image is obtained from the following publication: Souguir, A. et al. CT-guided celiac plexus neurolysis in the management of pain related to unresectable pancreatic cancer: About 2 cases. Open J. Gastroenterol. 05, 37–41 (2015) under the Creative Commons Attribution 4.0 International License (CC BY 4.0).}

\end{longtable}
\setcounter{footnote}{0}
  
\textbf{Supplementary Table 4. Tool Results.}
This table displays the tool results obtained by the agent for each
patient case prior to generating the final answer from Suppl. Table 3.
Results are color-coded to indicate accuracy, with correct outcomes in
green and incorrect ones in red.
\clearpage

\begin{longtable}{l p{10cm} l l l l r} 
\caption*{{\Large Supplementary Table 5: Model Correctness - Human Evaluation}}\\
\toprule
\textbf{No.} & \textbf{Model Outputs} & \textbf{1} & \textbf{2} & \textbf{3} & \textbf{4} & \textbf{$\mathbb{M}$} \\\midrule
\endfirsthead
\toprule
\textbf{No.} & \textbf{Model Outputs} & \textbf{1} & \textbf{2} & \textbf{3} & \textbf{4} & \textbf{$\mathbb{M}$} \\
\midrule
\endhead
\midrule
\multicolumn{7}{r}{Continued on next page} \\
\midrule
\endfoot
\bottomrule
\endlastfoot
A & Given Mrs. Adams' stable liver metastases as evidenced by the comparison of CTA\_11\_11\_2023.png and CTA\_02\_24\_2024.png [Tool], alongside her genetic profile indicating MSS, and wild-type status for both BRAF and KRAS [Tool], her disease appears to be in a controlled state under the current regimen of FOLFOX and bevacizumab. & \cmark & \cmark & \cmark & \cmark & \cmark \\
& The stability of liver lesions suggests that the chemotherapy, in conjunction with bevacizumab, is effective in managing her metastatic colorectal cancer (mCRC) [Tool]. & \cmark & \cmark & \cmark & \cmark & \cmark \\
& The development of hypertension, a known side effect of bevacizumab [11][12][13], necessitates a multidisciplinary approach to manage this condition without compromising her cancer treatment efficacy.  & \cmark & \cmark & \cmark & \cmark & \cmark \\
& Given the stable disease and the absence of BRAF and KRAS mutations [Tool], Mrs. Adams remains a candidate for continued anti-EGFR therapy, considering her left-sided mCRC, which is typically more responsive to such treatments [2][8]. & \cmark & \cmark & \cmark & \cmark & \cmark \\
& **Next Steps in Treatment Plan:** 1. **Management of Hypertension**: Immediate consultation with a cardiologist is recommended to manage her newly developed hypertension, potentially related to bevacizumab [22][23][26][27].  & \cmark & \cmark & \cmark & \cmark & \cmark \\
& Adjustments to her antihypertensive therapy may be necessary, with continuous monitoring of blood pressure to prevent cardiovascular complications [68][69][70][71]. & \cmark & \cmark & \cmark & \cmark & \cmark \\
& 2. **Ongoing Cancer Treatment**: Given the stable disease, continuation of the current regimen is advisable. However, the patient's tolerance, especially concerning the development of hypertension, should be closely monitored [1][4].  & \cmark & \cmark & \cmark & \cmark & \cmark \\
& If hypertension cannot be controlled, a dose adjustment or a switch to an alternative antiangiogenic agent with a potentially different side effect profile, such as aflibercept or ramucirumab, could be considered [9][86]. & \lmark & \xmark & \cmark & \lmark & \lmark \\
& 3. **Re-evaluation for Surgical Intervention**: Despite the stable metastatic disease, re-evaluation for the possibility of liver metastasectomy or ablation should be considered, as these could offer a chance for prolonged survival or even cure in selected cases [33][34][35]. & \cmark & \cmark & \cmark & \cmark & \cmark \\
& 4. **Maintenance Therapy**: If surgery is not feasible and once a maximum response to the current regimen is achieved, transitioning to maintenance therapy with a fluoropyrimidine (e.g., capecitabine) with or without bevacizumab could be considered to reduce toxicity while maintaining disease control [4][85]. & \cmark & \cmark & \cmark & \cmark & \cmark \\
& 5. **Genetic and Molecular Marker Re-evaluation**: Although initial genetic predictions indicate MSS, and wild-type BRAF and KRAS, a broader panel re-evaluation might uncover other actionable mutations or markers that could open up additional targeted therapy options [33][42][46]. & \cmark & \cmark & \cmark & \cmark & \cmark \\
& 6. **Clinical Trials and Novel Therapies**: Exploration of clinical trials focusing on novel therapies or combinations that could be beneficial for her specific genetic and molecular profile is recommended [6][33]. & \cmark & \cmark & \cmark & \cmark & \cmark \\
& 7. **Palliative Care Integration**: Early integration of palliative care to manage symptoms, improve quality of life, and support Mrs. Adams and her family through treatment decisions is crucial [126][127][129]. & \cmark & \cmark & \cmark & \cmark & \cmark \\
& 8. **Regular Monitoring**: Continuous assessment of her disease with imaging studies and tumor markers is essential to evaluate the response to treatment and adjust the treatment plan as necessary [38][80]. & \cmark & \cmark & \cmark & \cmark & \cmark \\
& This comprehensive approach, tailored to Mrs. Adams' specific medical history, current condition, and potential treatment pathways, aims to optimize her quality of life while effectively managing her mCRC. Regular reassessment of her condition and treatment tolerance will guide any necessary adjustments to her treatment plan. & \cmark & \cmark & \cmark & \cmark & \cmark \\
& To further refine Mrs. Adams' treatment plan, it would be helpful to know the specifics of her blood pressure readings over time, to assess the severity and progression of her hypertension. Utilizing my array of resources, I can further assist by identifying optimal blood pressure management strategies that align with her ongoing cancer treatment. Additionally, detailed information regarding her response to previous treatments and any side effects experienced would be invaluable. This would enable a more personalized approach to considering the integration of anti-EGFR therapy, ensuring it complements her current regimen while minimizing potential adverse effects. & \cmark & \cmark & \cmark & \cmark & \cmark \\
\midrule

B & Given Mrs. Brown's current clinical scenario, including the recent CT findings and genetic profile of her tumor, a strategic re-evaluation of her treatment plan is warranted. The presence of a new liver lesion suggests disease progression, and the genetic analysis revealing a KRAS mutation significantly impacts the choice of targeted therapy.  & \cmark & \cmark & \cmark & \cmark & \cmark \\
& Specifically, the efficacy of Panitumumab, an EGFR inhibitor, is compromised in the context of KRAS mutations [25], [26]. Therefore, continuing chemotherapy plus Panitumumab is not recommended based on the current evidence. & \cmark & \cmark & \cmark & \cmark & \cmark \\
&  1. **Discontinue Panitumumab**: Given the KRAS mutation, discontinuing Panitumumab is advised. KRAS mutations are known to confer resistance to EGFR inhibitors, rendering treatments like Panitumumab ineffective [25], [26]. & \cmark & \cmark & \cmark & \cmark & \cmark \\
& 2. **Adjust Chemotherapy Regimen**: Considering the disease progression, adjusting the chemotherapy regimen might be necessary. Options include switching to a different chemotherapy backbone such as FOLFIRI (folinic acid, fluorouracil, and irinotecan) if not previously used, or exploring the addition of bevacizumab, an anti-VEGF therapy, which is not affected by KRAS mutation status and could offer a benefit in this setting [10], [11], [12]. & \cmark & \cmark & \cmark & \cmark & \cmark \\
  & 3. **Explore Additional Targeted Therapies**: Given the KRAS mutation, targeted therapies that are effective against KRAS-mutated tumors should be considered. While direct KRAS inhibitors like sotorasib are currently approved for lung cancer with KRAS G12C mutations, ongoing trials are exploring their efficacy in colorectal cancer. They may offer potential benefits, although their applicability depends on the specific KRAS mutation present. Participation in a clinical trial targeting KRAS mutations could be an option [39], [40]. & \cmark & \cmark & \cmark & \cmark & \cmark \\
  & 4. **Consider Local Treatments for Liver Metastasis**: Depending on the size, location, and number of liver metastases, local treatments such as radiofrequency ablation (RFA) or stereotactic body radiation therapy (SBRT) could be considered to manage the liver disease [6], [16]. & \cmark & \cmark & \cmark & \cmark & \cmark \\
  &  5. **Clinical Trials**: Enrolling Mrs. Brown in clinical trials focusing on novel therapies for KRAS-mutated mCRC or new combinations of existing drugs could provide access to cutting-edge treatments and potentially improve outcomes [4]. & \xmark & \cmark & \cmark & \xmark & \xmark \\
  &  6. **Supportive Care and Monitoring**: Regardless of the treatment adjustments, ensuring supportive care to manage symptoms and side effects, along with regular monitoring of disease status through imaging and biomarkers, is crucial [6]. & \cmark & \cmark & \cmark & \cmark & \cmark \\
  &  7. **Patient Education and Involvement**: Discussing the rationale behind changing the treatment plan, potential benefits and side effects of the new regimen, and the importance of close monitoring with Mrs. Brown is essential. Her preferences and concerns should be taken into account in shared decision-making [7]. & \cmark & \cmark & \cmark & \cmark & \cmark \\
  & In summary, the treatment plan for Mrs. Brown should involve discontinuing Panitumumab due to the KRAS mutation, considering an alternative chemotherapy regimen possibly combined with bevacizumab, exploring eligibility for clinical trials, and evaluating local treatment options for liver metastasis. Regular follow-up and supportive care remain integral components of her care plan. & \cmark & \cmark & \cmark & \cmark & \cmark \\
  & To refine our approach based on Mrs. Brown's unique genetic profile and the progression of her disease, it would be beneficial to gather more detailed information on any additional genetic alterations that have not yet been explored. Utilizing my array of resources, I can further assist by identifying the most promising treatment avenues tailored to her specific genetic landscape. This includes pinpointing novel or underutilized targeted therapies that may bypass the resistance mechanisms posed by the KRAS mutation. Additionally, understanding the full scope of her disease through comprehensive genetic profiling could unveil new opportunities for clinical trial participation, potentially offering access to innovative treatments not yet available in standard care. & \cmark & \cmark & \cmark & \cmark & \cmark \\

D & Given Mr. Doe's advanced metastatic colorectal cancer with evidence of metastases in both the lungs and liver, as indicated by the **gen\_radiology\_report** tool analysis of **CT\_Lung.png** and **CT\_Liver.png** [Tool], he is more suitable for systemic therapy rather than surgery.  & \cmark & \cmark & \cmark & \cmark & \cmark \\
  & The presence of multiple well-defined nodules in both lung fields and multiple hypodense lesions in both lobes of the liver strongly suggests a widespread metastatic disease, making surgical intervention less feasible and unlikely to provide curative outcomes [Tool]. & \xmark & \cmark & \cmark & \xmark & \xmark \\
  & The **check\_mutations** tool's prediction of MSI-High status with a probability of 0.95 [Tool] opens up possibilities for immunotherapy, particularly given Mr. Doe's refusal to undergo chemotherapy. Pembrolizumab, an immune checkpoint inhibitor, is recommended for patients with MSI-High or dMMR metastatic colorectal cancer (mCRC) as a first-line treatment [25][30].  & \cmark & \cmark & \cmark & \cmark & \cmark \\
  & This aligns with Mr. Doe's treatment preferences and the current medical guidelines suggesting pembrolizumab for MSI-High mCRC [81][84]. & \cmark & \cmark & \cmark & \cmark & \cmark \\
  & The discrepancy regarding the BRAF V600E mutation status, with Mr. Doe's recollection of a positive result and the **check\_mutations** tool predicting wild-type BRAF with a probability of 0.48 [Tool], necessitates a confirmatory molecular test. If the BRAF V600E mutation is confirmed, targeted therapy with a combination of encorafenib and cetuximab is recommended in the second line, based on the BEACON study results and FDA approval for mCRC with BRAF V600E mutation after prior therapy [45][49]. & \cmark & \cmark & \cmark & \cmark & \cmark \\
  & This combination has shown efficacy in increasing overall survival compared with standard chemotherapy [72][73]. & \cmark & \cmark & \cmark & \cmark & \cmark \\
  & Furthermore, the slight likelihood of a KRAS mutation, as indicated by the **check\_mutations** tool with a probability of 0.52 [Tool], suggests that anti-EGFR therapies such as cetuximab or panitumumab, typically used in RAS wild-type mCRC, may not be as effective [42][43]. This further emphasizes the importance of confirming the BRAF mutation status, as the presence of a KRAS mutation limits the use of certain targeted therapies. & \cmark & \cmark & \cmark & \cmark & \cmark \\
  & Given the advanced nature of Mr. Doe's disease and his preference against chemotherapy, exploring clinical trials and newer treatment modalities is also a prudent step. This could include trials focusing on novel immunotherapies, targeted therapies based on other actionable mutations, or combination therapies that have not yet been standardly approved but show promise in early-phase trials. & \cmark & \cmark & \cmark & \cmark & \cmark \\
  & In summary, Mr. Doe's treatment strategy should involve: 1. Confirming the BRAF mutation status through a reliable molecular testing method. If BRAF V600E mutation is confirmed, proceed with targeted therapy using encorafenib plus cetuximab in future lines of treatment [45][49]. & \cmark & \cmark & \cmark & \cmark & \cmark \\
  & 2. Considering immunotherapy with pembrolizumab due to the MSI-High status, which has shown efficacy in MSI-High or dMMR mCRC as a first-line treatment [25][30]. & \cmark & \cmark & \cmark & \cmark & \cmark \\
  & 3. Exploring clinical trials and newer treatment modalities, given the advanced nature of Mr. Doe's disease and his preference against chemotherapy. & \cmark & \cmark & \cmark & \cmark & \cmark \\
  & 4. Involving a multidisciplinary team including medical oncologists, genetic counselors, and possibly palliative care specialists to address Mr. Doe's treatment preferences, manage symptoms, and ensure the best possible quality of life.
 & \cmark & \cmark & \cmark & \cmark & \cmark \\
  & Further investigations should include a repeat or confirmatory genetic test to resolve the discrepancy regarding the BRAF mutation status and comprehensive molecular profiling to identify any other actionable mutations or biomarkers that could guide therapy selection. This approach ensures a comprehensive and patient-centered care plan that aligns with the latest clinical evidence and Mr. Doe's treatment preferences.
 & \cmark & \cmark & \cmark & \cmark & \cmark \\
  & To further refine our analysis and recommendations, could you please provide the coordinates of the nodules and lesions highlighted in the CT scans? This information is crucial for a detailed segmentation analysis, which will help in accurately assessing the extent of metastatic spread. Additionally, if you could supply the measurements of these lesions over time, we can perform a progression analysis to better understand the disease's trajectory. Lastly, sharing specific search terms or questions you have regarding colorectal cancer, particularly those focusing on KRAS mutations and targeted therapy options, would enable a targeted literature review, ensuring we gather the most relevant and up-to-date information to guide treatment planning. & \cmark & \cmark & \cmark & \cmark & \cmark \\
\midrule

G & Given Mr. Garcia's progression of metastatic colorectal cancer, as evidenced by the significant increase in the overall area of hepatic lesions from 1,315 square units to 2,809 square units [Tool], which indicates a growth by a factor of 2.14 [Tool], a comprehensive and multidisciplinary approach is warranted. The strategy for Mr. Garcia should include the following steps, aligned with the structured approach and medical guidelines: & \cmark & \cmark & \cmark & \cmark & \cmark \\
  & 1. **Re-evaluation of the Molecular Profile**: Given the progression of disease, it's crucial to reassess the molecular characteristics of Mr. Garcia's tumor. This includes testing for RAS, BRAF mutations, and MSI status, as these can significantly influence the choice of targeted therapies [1], [2].  & \cmark & \cmark & \cmark & \cmark & \cmark \\
  & If not previously done, testing for HER2 amplification and NTRK fusions should also be considered due to their potential actionability [10], [124]. & \cmark & \cmark & \cmark & \cmark & \cmark \\
  & 2. **Assessment for Second-Line Chemotherapy Options**: Based on the molecular profile and prior treatment with FOLFOX, if Mr. Garcia's tumor is RAS wild-type and without BRAF V600E mutation, options may include anti-EGFR therapy (cetuximab or panitumumab) combined with irinotecan-based chemotherapy [2], [11].  & \cmark & \cmark & \cmark & \cmark & \cmark \\
  & For RAS or BRAF mutant tumors, or if anti-EGFR therapy is not suitable, consideration of FOLFIRI with bevacizumab or aflibercept could be appropriate [2], [7]. & \cmark & \cmark & \cmark & \cmark & \cmark \\
  & 3. **Consideration of Immunotherapy**: For tumors that are MSI-H or dMMR, immunotherapy with pembrolizumab or nivolumab, with or without ipilimumab, should be considered [3], [16], [26], [27]. This is based on the significant efficacy of immune checkpoint inhibitors in this subset of patients. & \xmark & \cmark & \cmark & \xmark & \xmark \\
  & 4. **Liver-Directed Therapies**: Given the liver-dominant nature of Mr. Garcia's metastatic disease, exploring liver-directed therapies such as radioembolization (Y-90), transarterial chemoembolization (TACE), or ablative therapies could provide local control and potentially palliate symptoms [38], [40], [48], [49]. These options should be discussed in a multidisciplinary team (MDT) setting, considering Mr. Garcia's overall health status and liver function. & \cmark & \cmark & \cmark & \cmark & \cmark \\
  & 5. **MDT Discussion and Supportive Care Optimization**: An MDT discussion involving medical oncology, surgical oncology, radiology, and palliative care is essential to integrate the above considerations into a personalized treatment plan for Mr. Garcia [Strategy]. Concurrently, optimization of supportive care measures to manage symptoms and maintain quality of life is paramount [121]. & \cmark & \cmark & \cmark & \cmark & \cmark \\
  & 6. **Clinical Trial Enrollment**: Finally, given the refractory nature of Mr. Garcia's disease, exploration of eligibility for clinical trials testing novel agents or combinations should be a priority. This could offer access to emerging therapies and contribute to the advancement of treatment for metastatic colorectal cancer [136]. & \cmark & \cmark & \cmark & \cmark & \cmark \\
  & In summary, Mr. Garcia's treatment plan should be personalized based on the latest molecular profiling, consider second-line systemic therapy options tailored to his tumor's characteristics, evaluate the feasibility of liver-directed therapies, and ensure supportive care needs are met, all within the context of an MDT discussion. Participation in clinical trials should also be actively pursued. & \cmark & \cmark & \cmark & \cmark & \cmark \\
  & To further refine our approach in managing Mr. Garcia's condition, I would need to analyze histologic images for genetic mutations that could guide targeted therapy selection. Could you provide the path to the folder containing Mr. Garcia's histologic images? This will enable me to perform genetic modeling to check for MSI, BRAF, and KRAS mutations. Additionally, knowing any specific genetic alterations identified would allow me to consult relevant databases for targeted treatment options, enhancing the precision of Mr. Garcia's treatment plan. Lastly, detailed information regarding Mr. Garcia's eligibility criteria for clinical trials would allow for a targeted search for novel therapeutic opportunities, potentially offering access to groundbreaking treatments. & \cmark & \cmark & \cmark & \cmark & \cmark \\
\midrule

Le & Given Mr. Lee's history of pancreatic cancer, post-Whipple procedure status, and completion of five cycles of gemcitabine chemotherapy, along with the recent CT scan findings indicating progression of lung nodules, a comprehensive approach is warranted.  & \cmark & \cmark & \cmark & \cmark & \cmark \\
  & The presence of a KRAS G12C mutation in his tumor adds a significant layer to his treatment planning, as this mutation opens the door to targeted therapy options that have shown promise in recent studies [Tool]. & \cmark & \cmark & \cmark & \cmark & \cmark \\
  & \#\#\# Standard Chemotherapy Protocols For patients like Mr. Lee, who have undergone surgical resection followed by adjuvant chemotherapy but show signs of disease recurrence or progression, the choice of further treatment depends on several factors including performance status, extent of disease, and molecular tumor profile.  & \cmark & \cmark & \cmark & \cmark & \cmark \\
  & In the absence of specific contraindications, and considering Mr. Lee's good performance status and active lifestyle, a more aggressive approach could be considered. & \cmark & \cmark & \cmark & \cmark & \cmark \\
  & 1. **Modified FOLFIRINOX**: This regimen, consisting of oxaliplatin, irinotecan, fluorouracil (5-FU), and leucovorin, has been recommended for patients with a good performance status in the metastatic setting and could be considered here, despite its intensive nature [4][5]. The regimen's efficacy in the adjuvant setting post-surgery suggests its potential utility in cases of recurrence [13][20]. & \lmark & \cmark & \cmark & \lmark & \lmark \\
  &  2. **Gemcitabine and Nab-Paclitaxel**: This combination has shown efficacy in the metastatic pancreatic cancer setting and could be an option for Mr. Lee, given his previous tolerance of gemcitabine [16][18]. This regimen might offer a balance between efficacy and tolerability. & \cmark & \cmark & \cmark & \cmark & \cmark \\
  & \#\#\# Targeted Therapy for KRAS G12C Mutation\\
  & The discovery of the KRAS G12C mutation in Mr. Lee's tumor is pivotal. Recent advancements have led to the development of targeted therapies for this specific mutation, which is traditionally considered challenging to treat [Tool].
  & \cmark & \cmark & \cmark & \cmark & \cmark \\
& 1. **Sotorasib**: As the first targeted agent with regulatory approval for KRAS G12C-mutated NSCLC, sotorasib represents a novel treatment option. Although primarily studied in lung cancer, its efficacy in KRAS G12C-positive tumors suggests potential applicability in Mr. Lee's case, given the lung nodules' progression [26][29][33]. The reported durable clinical responses and manageable safety profile make it a compelling option [Tool]. & \cmark & \cmark & \cmark & \cmark & \cmark \\
  & 2. **Divarasib (GDC-6036)**: This agent has shown durable responses across KRAS G12C-positive tumors, including low-grade adverse events, which could be advantageous for maintaining Mr. Lee's quality of life [Tool]. & \cmark & \cmark & \cmark & \cmark & \cmark \\
  & 3. **Adagrasib**: Similar to sotorasib, adagrasib has shown efficacy in KRAS G12C-mutated tumors. Its use, alone or in combination with cetuximab, has been evaluated in colorectal cancer patients with this mutation, showing promising efficacy [Tool]. Given the cross-tumor activity of KRAS G12C inhibitors, adagrasib could be considered for Mr. Lee, especially in light of the lung nodules' progression. & \cmark & \cmark & \cmark & \cmark & \cmark \\
  & \#\#\# Conclusion and Next Steps For Mr. Lee, a dual approach incorporating systemic chemotherapy with consideration for targeted therapy against the KRAS G12C mutation appears warranted.  & \cmark & \cmark & \cmark & \cmark & \cmark \\
  & Given the progression in lung nodules and the presence of the KRAS G12C mutation, exploring the use of sotorasib or adagrasib, in consultation with a multidisciplinary team, could provide a targeted approach to his treatment.  & \cmark & \cmark & \cmark & \cmark & \cmark \\
  & Additionally, enrollment in clinical trials investigating these or new agents for KRAS G12C mutations should be considered, offering Mr. Lee access to cutting-edge treatments and contributing to the growing body of knowledge on managing this genetic alteration [Tool]. & \cmark & \cmark & \cmark & \cmark & \cmark \\
  & Regular monitoring and supportive care measures are essential to maintain Mr. Lee's quality of life during treatment, including managing any side effects from the targeted therapy and providing psychological support as needed. This approach aligns with current medical guidelines and reflects a personalized treatment strategy based on Mr. Lee's unique medical history, current condition, and molecular tumor profile [4][5][13][20][26][29][33]. & \cmark & \cmark & \cmark & \cmark & \cmark \\
  & To further refine our approach and ensure the most accurate and personalized treatment recommendations for Mr. Lee, I would need to delve deeper into specific aspects of his condition. If you could provide more details on the lung nodules, this would help in assessing the dynamics of the disease more accurately. Additionally, if histopathology images are available, providing the unique identifier for Mr. Lee along with specifying interest in the KRAS mutation would enable a precise genetic modeling to confirm the KRAS G12C mutation status. This detailed information is crucial for tailoring the most effective treatment strategy. & \cmark & \cmark & \cmark & \cmark & \cmark \\
\midrule

Lo & Given Mr. Lopez's diagnosis of left-sided colorectal cancer with potential liver metastases, MSS status, and wild-type BRAF and KRAS genes [Patient], [Tool], the optimal management approach should be multi-pronged, incorporating systemic chemotherapy, evaluation for surgical resection or liver-directed therapies, and supportive care measures.  & \cmark & \cmark & \cmark & \cmark & \cmark \\
  & 1. **Systemic Chemotherapy Initiation**: Based on the NCCN guidelines [1], for Mr. Lopez, who has MSS, left-sided colorectal cancer without BRAF or KRAS mutations, a combination chemotherapy regimen such as FOLFOX (oxaliplatin plus fluorouracil and leucovorin) or FOLFIRI (irinotecan plus fluorouracil and leucovorin) should be considered as the first line of treatment.  & \cmark & \cmark & \cmark & \cmark & \cmark \\
  & Given the absence of RAS mutations, the addition of an anti-EGFR agent, specifically cetuximab or panitumumab, is recommended to enhance treatment efficacy [1], [74]. & \cmark & \cmark & \cmark & \cmark & \cmark \\
  & 2. **Further Imaging and Assessment for Liver Metastases**: Before deciding on local treatments, further characterization of the liver lesions with MRI or PET-CT is advised to confirm metastasis and assess resectability [7]. This step is crucial as it influences the decision regarding local versus systemic therapy and the potential for curative intervention [4]. & \cmark & \cmark & \cmark & \cmark & \cmark \\
  & 3. **Evaluation for Surgical Resection or Liver-Directed Therapies**: If the liver metastases are confirmed and deemed limited and potentially resectable, Mr. Lopez should be evaluated by a multidisciplinary team including a surgical oncologist [4], [6].  & \cmark & \cmark & \cmark & \cmark & \cmark \\
  & Options such as radiofrequency ablation (RFA), microwave ablation (MWA), or stereotactic body radiotherapy (SBRT) could be considered if surgical resection is not feasible due to the location or number of lesions, or if Mr. Lopez's health status does not permit surgery [21], [23], [84]. & \cmark & \cmark & \cmark & \cmark & \cmark \\
  & 4. **Maintenance Therapy and Regular Monitoring**: Following the initial treatment phase, transitioning Mr. Lopez to maintenance therapy with a less intensive regimen, such as fluoropyrimidine alone or in combination with bevacizumab, could be considered to manage the disease while minimizing side effects [32].  & \cmark & \cmark & \cmark & \cmark & \cmark \\
  & Regular monitoring through imaging studies and tumor markers will be crucial to adjust the treatment plan based on the disease's response and progression. & \cmark & \cmark & \cmark & \cmark & \cmark \\
  & 5. **Comprehensive Supportive Care**: Throughout the treatment process, it is essential to provide supportive care measures aimed at managing symptoms, improving quality of life, and addressing any treatment-related side effects. This includes nutritional support, pain management, psychological support, and management of chemotherapy or targeted therapy-related toxicities [5]. & \cmark & \cmark & \cmark & \cmark & \cmark \\
  & In summary, Mr. Lopez's treatment plan should start with systemic chemotherapy combined with an anti-EGFR agent, followed by a detailed assessment of liver metastases for potential resectability or suitability for liver-directed therapies. The treatment strategy should be flexible, with adjustments based on response to therapy and Mr. Lopez's overall health status. A multidisciplinary approach involving oncologists, radiologists, surgical oncologists, and supportive care specialists is essential for optimal management. & \cmark & \cmark & \cmark & \cmark & \cmark \\
  & To further refine Mr. Lopez's treatment strategy with precision and up-to-date research findings, could you provide the exact genetic mutations beyond BRAF and KRAS wild-type status? This will enable a detailed OncoKB analysis to identify any additional targeted therapy options. Additionally, for a comprehensive assessment of the liver metastases, please supply the file path to the most recent CT scan images along with the coordinates for each visible lesion. This will assist in accurately determining the extent of liver involvement and the feasibility of local treatments. & \cmark & \cmark & \cmark & \cmark & \cmark \\
\midrule

S & Given the comprehensive details provided about Mrs. Smith's medical history and the recent diagnostic findings, a thorough analysis reveals critical insights into her current health status, particularly concerning her liver metastases and the implications of the newly detected KRAS mutation on her treatment regimen. & \cmark & \cmark & \cmark & \cmark & \cmark \\
  & **Current Health Status and Liver Metastases:** \\
  & The radiology report analysis comparing MRI scans from February 2024 and November 2023 indicates a notable change in the liver metastases.  & \cmark & \cmark & \cmark & \cmark & \cmark \\
  & The increased brightness in the February 2024 scan compared to the November 2023 scan suggests a potential progression or alteration in the liver metastases [Tool].  & \xmark & \xmark & \cmark & \xmark & \xmark \\
  & However, without additional clinical context, such as liver function tests or symptoms, the clinical significance of this radiological change remains uncertain. It's crucial to correlate these imaging findings with clinical assessments and laboratory results including tumor markers to accurately determine the current state of the liver metastases. & \cmark & \xmark & \cmark & \xmark & \xmark \\
  & **Implications of KRAS Mutation:** \\
 & The detection of a KRAS mutation (with a probability of 0.52) represents a significant change from the initial genetic profiling, which identified Mrs. Smith as KRAS wild type [Tool]. This finding is crucial as it impacts the efficacy of the current treatment regimen. Anti-EGFR therapies, such as Panitumumab, are known to be ineffective in the presence of KRAS mutations [117][118]. & \cmark & \cmark & \cmark & \cmark & \cmark \\
 &  The presence of a KRAS mutation suggests that Mrs. Smith's cancer has acquired resistance to the anti-EGFR monoclonal antibody, thereby reducing the effectiveness of her current treatment regimen of 5-Fluorouracil and Panitumumab. & \cmark & \cmark & \cmark & \cmark & \cmark \\
  & **Treatment Recommendations:** \\
 & Given the new genetic mutation analysis revealing a KRAS mutation, continuing the regimen of 5-Fluorouracil and Panitumumab is not advisable [117][118].  & \cmark & \cmark & \cmark & \cmark & \cmark \\
 & The detection of a KRAS mutation necessitates a reevaluation of Mrs. Smith's treatment plan. Alternative treatment options should be considered, focusing on regimens that do not rely on the efficacy of anti-EGFR therapies. Options may include chemotherapy combinations without anti-EGFR monoclonal antibodies or considering enrollment in clinical trials exploring novel agents targeting KRAS mutations or other pathways relevant to her tumor's genetic profile [120][126]. & \cmark & \cmark & \cmark & \cmark & \cmark \\
  & **Further Actions:** \\ 
  & 1. **Clinical Correlation**: A multidisciplinary team discussion is recommended to integrate the new radiological and genetic findings with Mrs. Smith's clinical status, including any new symptoms or changes in her overall health. & \cmark & \cmark & \cmark & \cmark & \cmark \\
  & 2. **Additional Testing**: Further genetic testing to confirm the KRAS mutation subtype could provide insights into targeted therapy options, including KRAS G12C inhibitors if applicable [120]. & \cmark & \cmark & \cmark & \cmark & \cmark \\
  & 3. **Treatment Adjustment**: Based on the KRAS mutation, switching to a chemotherapy regimen not reliant on EGFR inhibition, such as FOLFIRI (irinotecan, fluorouracil, and leucovorin), possibly in combination with bevacizumab if not contraindicated, should be considered [19][20]. & \cmark & \cmark & \cmark & \cmark & \cmark \\
  & 4. **Clinical Trials**: Exploration of clinical trials targeting KRAS mutations or other novel therapies suitable for Mrs. Smith's genetic profile and disease state is highly recommended [126]. & \cmark & \cmark & \cmark & \cmark & \cmark \\
  & In conclusion, the new findings necessitate a significant adjustment in Mrs. Smith's treatment strategy, moving away from anti-EGFR therapies due to the detected KRAS mutation and considering alternative systemic therapies or clinical trial enrollment for novel treatments. & \cmark & \cmark & \cmark & \cmark & \cmark \\
  & To enhance our understanding of the progression of Mrs. Smith's liver metastases and refine her treatment options, could you please supply the exact dates of the CT scans you're referring to? This will allow for a precise comparison of changes over time. Additionally, knowing the precise KRAS mutation subtype (e.g., G12C, G12D) will enable a more targeted search for applicable therapies and clinical trials. Lastly, for a comprehensive literature review on the efficacy of her current treatment in the context of her specific KRAS mutation, please provide any keywords or phrases that you believe are crucial for this search, such as "KRAS mutation colorectal cancer resistance." These details are vital for leveraging our resources to their fullest potential, ensuring a tailored approach to Mrs. Smith's care. & \cmark & \cmark & \cmark & \cmark & \cmark \\
\midrule

T & Given Mr. Taylor's advanced HCC with portal vein thrombosis (PVT), his preserved liver function (Child-Pugh A5), and his good performance status (ECOG PS 0), a multidisciplinary approach is crucial for optimizing his treatment strategy [1], [5], [8].  & \cmark & \cmark & \cmark & \cmark & \cmark \\
  & The presence of PVT categorizes his HCC as advanced or BCLC stage C, for which systemic therapy is recommended over locoregional therapies [3], [4], [59]. & \cmark & \xmark & \cmark & \cmark & \cmark \\
  &  First-line systemic therapy with atezolizumab plus bevacizumab is recommended for patients like Mr. Taylor, who have advanced HCC, no contraindications to bevacizumab, and preserved liver function [60], [93].  & \cmark & \cmark & \cmark & \cmark & \cmark \\
  & This recommendation is based on the IMbrave150 trial, which demonstrated a significant survival benefit of atezolizumab-bevacizumab over sorafenib [93].  & \cmark & \cmark & \cmark & \cmark & \cmark \\
  & However, given the presence of esophageal varices, it is imperative to ensure that these have been appropriately managed to minimize the risk of variceal bleeding, a known complication associated with bevacizumab due to its antiangiogenic effects [70], [71]. & \cmark & \cmark & \cmark & \cmark & \cmark \\
  & Considering the progression of his disease despite previous TACE sessions, further locoregional therapies such as additional TACE or radioembolization (TARE) are unlikely to provide significant benefit and may not be appropriate given the extent of his PVT [10], [63].  & \cmark & \cmark & \cmark & \cmark & \cmark \\
  & However, if systemic therapy is contraindicated or not tolerated, and Mr. Taylor's PVT is classified as Vp1 or Vp2 (involving segmental or subsegmental branches of the portal vein), locoregional therapies could be reconsidered in a highly selective manner [7], [63]. & \cmark & \cmark & \cmark & \cmark & \cmark \\
  & Clinical trial enrollment should be explored for Mr. Taylor, given the rapidly evolving landscape of HCC treatment, which may offer access to novel therapeutic agents or combinations not yet available in standard practice [2], [22]. This approach is particularly relevant for advanced HCC, where emerging treatments continue to show promise. & \cmark & \cmark & \cmark & \cmark & \cmark \\
  & Supportive care measures, including management of ascites and screening for esophageal varices, should be optimized. Therapeutic paracentesis may be required for symptomatic ascites, and beta-blockers or endoscopic variceal ligation (EVL) should be considered for variceal prophylaxis [80], [81], [82]. & \cmark & \cmark & \cmark & \cmark & \cmark \\
  &  Given his portal hypertension, as evidenced by splenomegaly and ascites, careful monitoring for complications such as spontaneous bacterial peritonitis (SBP) is warranted [85]. & \cmark & \cmark & \cmark & \cmark & \cmark \\
  & Regular monitoring of Mr. Taylor's liver function, tumor markers, and imaging studies is essential to assess the response to any new treatment and adjust the management plan as necessary [6], [24].  & \cmark & \cmark & \cmark & \cmark & \cmark \\
  & Palliative care consultation should also be involved early to manage symptoms and support quality of life [6]. & \cmark & \cmark & \cmark & \cmark & \cmark \\
  &  In summary, the next steps in managing Mr. Taylor's disease should include consideration of systemic therapy with atezolizumab plus bevacizumab, exploration of clinical trial enrollment, and supportive care measures tailored to his specific needs and disease status.
 & \cmark & \cmark & \cmark & \cmark & \cmark \\
  & To further refine our approach and ensure the most accurate and personalized treatment recommendations for Mr. Taylor, could you please supply the path to the histologic or histopathology images of Mr. Taylor's liver tumor? This will enable a detailed analysis to identify potential genetic markers critical for his treatment planning. Additionally, if you could list any specific genetic alterations already identified in Mr. Taylor's case, it would greatly enhance our ability to query targeted treatment options and relevant clinical trials. For the most recent imaging studies, specifying the regions of interest would allow for precise measurements and better assessment of the tumor's response to treatment over time. Lastly, providing specific search terms related to Mr. Taylor's condition and any known mutations will facilitate a more focused literature search to support evidence-based decision-making. & \cmark & \cmark & \cmark & \cmark & \cmark \\
\midrule

W & Given Mrs. Williams' progressive disease under the FOLFOX regimen and Panitumumab, as evidenced by the growth of her liver metastasis, the next steps involve a comprehensive reassessment of her treatment plan based on her genetic profile and current medical guidelines. The increase in the size of the metastasis, with a ratio of approximately 1.799, clearly indicates progressive disease [Tool]. & \cmark & \cmark & \cmark & \cmark & \cmark \\
  &  1. **Targeted Therapy for BRAF V600E Mutation**: Considering the BRAF V600E mutation, the combination of encorafenib and cetuximab is recommended. This combination is FDA-approved for mCRC with a BRAF V600E mutation, after prior therapy [12], [13], [41].  & \cmark & \cmark & \cmark & \cmark & \cmark \\
  & This approach is supported by the efficacy shown in the BEACON CRC trial, which demonstrated a significant improvement in overall survival and response rate with this combination [3], [41]. & \cmark & \cmark & \cmark & \cmark & \cmark \\
  &  2. **Immunotherapy Consideration**: The MSI-High status of the tumor suggests a potential benefit from immunotherapy. Pembrolizumab, recommended as first-line therapy for patients with dMMR mCRC by the ASCO 2022 guideline [20], could be considered given the patient's genetic profile.  & \xmark & \cmark & \cmark & \xmark & \xmark \\
  & This option should be discussed with Mrs. Williams, considering the potential for a better response to immunotherapy in MSI-High status tumors [19], [20], [22]. & \cmark & \cmark & \cmark & \cmark & \cmark \\
  &  3. **Local Treatment Options**: For the solitary liver metastasis, exploring local treatments such as radiofrequency ablation (RFA), stereotactic body radiotherapy (SBRT), or selective internal radiation therapy (SIRT) could be beneficial [46], [47], [48]. These options might offer localized control of the metastasis, especially if systemic therapy options are exhausted or if the metastasis becomes more localized. & \cmark & \cmark & \cmark & \cmark & \cmark \\
  &  4. **Systemic Therapy Alternatives**: If targeted therapy and immunotherapy are not immediately pursued or if further progression is noted, considering FOLFIRI as a second-line therapy could be an option [76].  & \cmark & \cmark & \cmark & \cmark & \cmark \\
  & However, given the genetic makeup of Mrs. Williams' tumor, targeted therapy options should be prioritized over switching to another chemotherapy regimen [5], [7]. & \cmark & \cmark & \cmark & \cmark & \cmark \\
  &  5. **Multidisciplinary Team Discussion**: A multidisciplinary discussion involving medical oncologists, surgical oncologists, and radiologists is crucial to tailor the treatment plan to Mrs. Williams' specific condition, genetic makeup, and preferences [56], [57]. This team approach ensures that all potential treatment avenues are explored and that the chosen strategy aligns with the patient's goals and quality of life considerations.
 & \cmark & \cmark & \cmark & \cmark & \cmark \\
  &  6. **Patient Consultation**: It is essential to discuss all potential treatment options with Mrs. Williams, including the benefits, risks, and potential side effects of each treatment [6]. Her preferences, quality of life considerations, and any concerns should be central to the decision-making process.
 & \cmark & \cmark & \cmark & \cmark & \cmark \\
  &  In summary, the immediate next steps involve prioritizing targeted therapy with encorafenib and cetuximab, exploring immunotherapy options due to the MSI-High status, and considering local treatment options for the liver metastasis. A multidisciplinary approach and patient-centered discussion are paramount in deciding the future course of treatment for Mrs. Williams.
 & \cmark & \cmark & \cmark & \cmark & \cmark \\
  & To enhance our exploration of second-line therapy options and local treatment possibilities for Mrs. Williams, it would be beneficial if you could provide the exact search terms or phrases you have in mind for investigating the effectiveness of FOLFIRI in BRAF V600E mutated colorectal cancer and the latest advancements in local treatment options for liver metastasis. Additionally, specifying the particular aspects of colorectal cancer treatment you're most interested in could greatly assist in tailoring the search to yield the most relevant and up-to-date guidelines and research findings. This focused approach will ensure we utilize our resources to gather the most pertinent information to aid in Mrs. Williams' treatment planning. & \cmark & \cmark & \cmark & \cmark & \cmark \\
\midrule

X & The latest MRI scan from February 2024 reveals that the metastatic lesion in Ms. Xing's left liver lobe has notably enlarged, growing from an area of 133 in September 2023 to 518, signifying a progression in her condition. This significant increase highlights the critical need to consider further treatment avenues for Ms. Xing. <& \cmark & \cmark & \cmark & \cmark & \cmark \\
  & Given Ms. Xing's medical history and the molecular characteristics of her tumor, including the BRAF V600E mutation and CD74-ROS1 variant fusion, several targeted therapy options emerge: & \cmark & \cmark & \cmark & \cmark & \cmark \\
  & 1. **Dabrafenib and Trametinib**: This combination is recommended for BRAF V600E-mutated tumors, including cholangiocellular carcinoma, based on their effectiveness in other cancers with similar genetic alterations [Tool]. The combination has FDA approval for BRAF V600E mutant unresectable or metastatic melanoma and other solid tumors excluding colorectal cancer, with noted effectiveness in biliary tract cancer [Tool]. & \cmark & \cmark & \cmark & \cmark & \cmark \\
  & 2. **Crizotinib, Entrectinib, and Repotrectinib**: These drugs are identified treatments for ROS1-positive tumors, primarily in NSCLC. Given the CD74-ROS1 fusion in Ms. Xing's tumor, these options could be explored for their potential applicability, despite the primary association with NSCLC [Tool]. & \cmark & \cmark & \cmark & \cmark & \cmark \\
  & 3. **Encorafenib and Binimetinib**: This combination, highlighted through a Google search, could be considered based on their effectiveness in cancers with BRAF V600E mutations [Tool]. Although not specifically indicated for cholangiocellular carcinoma, their use in similar genetic contexts suggests potential applicability. & \cmark & \cmark & \cmark & \cmark & \cmark \\
  & Regarding chemotherapy options, the current guidelines for cholangiocellular carcinoma recommend systemic chemotherapy as the primary treatment for advanced disease. Given Ms. Xing's prior treatment with gemcitabine and cisplatin, alternative regimens could include:
   - **FOLFOX** (fluorouracil, leucovorin, and oxaliplatin) as a second-line treatment option, especially for patients who have progressed on a gemcitabine-based regimen [55]. & \cmark & \lmark & \cmark & \cmark & \cmark \\
  & For Ms. Xing, considering her young age and the urgent need for effective management strategies, participation in clinical trials exploring novel agents or combinations should be strongly considered. This includes trials for targeted therapies based on her tumor's molecular profile or emerging immunotherapy options for cholangiocellular carcinoma. & \cmark & \cmark & \cmark & \cmark & \cmark \\
  & Given the conflicting information on microsatellite status, further clarification is needed. The 'check\_mutations' tool indicated Microsatellite Stability (MSS), however, the patient's molecular report specified that there was Microsatellite Instability (MSI). & \cmark & \cmark & \cmark & \cmark & \cmark \\
  &  If MSI-High status is confirmed, immunotherapy with pembrolizumab could be considered, as it has shown effectiveness in MSI-High tumors across various cancer types [57]. & \cmark & \cmark & \cmark & \cmark & \cmark \\
  & In summary, for the tumor board discussion, the recommended approach includes exploring targeted therapy options such as Dabrafenib and Trametinib for the BRAF V600E mutation and potentially Crizotinib or Entrectinib for the CD74-ROS1 fusion. Additionally, considering second-line chemotherapy options and the possibility of clinical trial participation are crucial steps. The significant progression of Ms. Xing's disease necessitates a prompt and comprehensive evaluation of these treatment strategies to optimize her care. & \cmark & \cmark & \cmark & \cmark & \cmark \\
  & To refine our approach and ensure the most accurate and personalized treatment recommendations for Ms. Xing, could you please provide the latest comprehensive genomic profiling of her tumor? This includes any recent sequencing data that might reveal additional actionable mutations or variants not previously identified. Additionally, detailed information regarding her response to past treatments, including any side effects experienced and the duration of response, would greatly enhance our ability to tailor future therapy options. This data is crucial for leveraging our resources to their fullest potential in guiding Ms. Xing's treatment plan. & \cmark & \cmark & \cmark & \cmark & \cmark \\
\midrule

Z & Given Mr. Zu's diagnosis of pancreatic adenocarcinoma with potential liver metastases as indicated by the presence of multiple hypodense areas within the liver [Tool], a comprehensive and multidisciplinary approach is warranted. The treatment strategy should balance aggressive management of metastatic disease with considerations for Mr. Zu's quality of life and preferences. Here's a detailed treatment plan based on current medical guidelines and the specific patient context at hand: & \cmark & \cmark & \cmark & \cmark & \cmark \\
  & 1. **Confirm Metastatic Status**: Before finalizing the treatment plan, it's crucial to confirm whether the hypodense liver areas represent metastatic disease. This can be achieved through further imaging, such as MRI with liver-specific contrast or PET-CT, and potentially a liver biopsy for histological confirmation [Tool]. The outcome of these investigations will significantly influence the treatment pathway. & \cmark & \cmark & \cmark & \cmark & \cmark \\
  & 2. **Genetic Profiling**: Although the mutation check encountered an error [Tool], pursuing comprehensive genetic testing is essential. This includes testing for MSI, BRAF, KRAS, and potentially other relevant markers like BRCA mutations, given the pancreatic origin of the cancer. The results could open up options for targeted therapies [21], [43]. & \cmark & \cmark & \cmark & \cmark & \cmark \\
  & 3. **Systemic Chemotherapy**: If metastases are confirmed, systemic chemotherapy becomes the primary treatment approach. For a patient with a good performance status like Mr. Zu, a regimen like FOLFIRINOX (if tolerated) or gemcitabine plus nab-paclitaxel could be considered, as these are standard options for metastatic pancreatic adenocarcinoma [13], [19]. The choice between these regimens should consider Mr. Zu's overall health, comorbidity profile, and personal preferences. & \cmark & \cmark & \cmark & \cmark & \cmark \\
  & 4. **Local Control Options**: If the liver metastases are limited in number and the primary tumor is controlled, local therapies such as stereotactic body radiotherapy (SBRT) for liver metastases could be discussed. This approach could potentially improve quality of life and overall survival [23], [26]. & \cmark & \cmark & \cmark & \cmark & \cmark \\
  & 5. **Supportive Care**: Integrating palliative care early in the treatment plan is crucial to manage symptoms and improve quality of life. This includes pain management, nutritional support, and addressing any psychological needs [111], [112].
 & \xmark & \cmark & \cmark & \cmark & \cmark \\
  & 6. **Clinical Trials**: Exploring the availability of clinical trials that Mr. Zu may be eligible for is important. This could include trials focusing on novel therapies or combinations that could offer benefit beyond standard care [154].
 & \cmark & \cmark & \cmark & \cmark & \cmark \\
  & 7. **Multidisciplinary Evaluation**: Ensure that Mr. Zu's treatment plan is discussed in a multidisciplinary tumor board meeting, incorporating input from medical oncology, surgical oncology, radiation oncology, radiology, and palliative care to formulate a comprehensive and individualized treatment plan [150]. & \cmark & \cmark & \cmark & \cmark & \cmark \\
  & 8. **Follow-Up and Monitoring**: Plan for regular follow-up to assess treatment response, manage side effects, and adjust the treatment plan as necessary. This includes monitoring tumor markers and imaging studies [78]. & \cmark & \cmark & \cmark & \cmark & \cmark \\
  & Given the complexity of Mr. Zu's case, especially with the potential presence of liver metastases, it's imperative that the treatment plan is flexible and responsive to the evolving nature of his condition. Further, the limitations encountered in genetic profiling necessitate a reattempt or alternative methods to ensure a comprehensive understanding of the tumor's genetic landscape, which could significantly impact treatment decisions [Tool]. & \cmark & \cmark & \cmark & \cmark & \cmark \\
  & To further refine our approach and ensure we're leveraging the most accurate and personalized treatment options for Mr. Zu, I would kindly ask for the re-submission of histologic or histopathology images. Specifically, high-resolution images of the affected tissue would be invaluable. These images are crucial for a detailed genetic modeling analysis, aiming to identify MSI, BRAF, and KRAS mutations with precision. Additionally, if you could provide any available clinical data regarding Mr. Zu's health status, including any previous treatments and their outcomes, it would greatly assist in tailoring the query to the OncoKB database. This information will enable us to pinpoint targeted treatment options that are most likely to benefit Mr. Zu, based on his unique genetic profile and clinical scenario.
 & \cmark & \cmark & \cmark & \cmark & \cmark \\

\end{longtable}
\setcounter{footnote}{0}

\textbf{Supplementary Table 5. Model Correctness.}
This table presents the human evaluators' ratings for each statement
generated by the model across all patient scenarios. A checkmark
signifies a statement deemed accurate, a cross represents a statement
assessed as incorrect, and an orange lightning bolt highlights a
response considered harmful. \emph{M} denotes the consensus reached
through a majority vote among the evaluators. For further information,
please consult the Material and Methods section under Human Results
Evaluation.
\clearpage

\begin{longtable}{l p{10cm} l l l l r} 
\caption*{{\Large Supplementary Table 6: Model Completeness - Human Evaluation}}\\
\toprule
\textbf{No.} & \textbf{Keywords} & \textbf{1} & \textbf{2} & \textbf{3} & \textbf{4} & \textbf{$\mathbb{M}$} \\\midrule
\endfirsthead
\toprule
\textbf{No.} & \textbf{Keywords} & \textbf{1} & \textbf{2} & \textbf{3} & \textbf{4} & \textbf{$\mathbb{M}$} \\
\midrule
\endhead
\midrule
\multicolumn{7}{r}{Continued on next page} \\
\midrule
\endfoot
\bottomrule
\endlastfoot
A & Imaging assessments from 11/11/2023 to 02/24/2024 indicate no progression of disease. & \cmark & \cmark & \cmark & \cmark & \cmark \\
  & Genetic testing reveals no mutations in BRAF and KRAS; microsatellite status is stable. & \cmark & \cmark & \cmark & \cmark & \cmark \\
  & Address hypertension as a potential adverse effect of bevacizumab treatment. & \cmark & \cmark & \cmark & \cmark & \cmark \\
  & Propose a management approach for the elevated blood pressure. & \cmark & \cmark & \cmark & \cmark & \cmark \\
  & Given the wild-type KRAS status, consider targeting EGFR with Panitumumab as an alternative herapy. & \cmark & \cmark & \cmark & \cmark & \cmark \\
  & Continue FOLFOX regimen to maximize response, then assess for surgical or liver-directed therapies, noting no evidence of extrahepatic metastasis. & \cmark & \cmark & \cmark & \cmark & \cmark \\
  & In cases of sustained stable disease without viable local treatment options, consider treatment de-escalation and maintenance with 5FU and monoclonal antibody therapy. & \cmark & \cmark & \cmark & \cmark & \cmark \\
\midrule
B & Imaging evaluations reveal progressive disease characterized by increasing metastatic involvement of the liver. & \cmark & \cmark & \cmark & \cmark & \cmark \\
  & Mutational analysis confirms wild-type BRAF and mutated KRAS status, with microsatellite stability maintained. & \xmark & \cmark & \cmark & \cmark & \cmark \\
  & Due to the presence of a KRAS mutation, cessation of anti-EGFR therapy is recommended. & \cmark & \cmark & \cmark & \cmark & \cmark \\
  & Recommend further mutational analyses, including a comprehensive panel diagnostic, to delineate KRAS subtypes and identify additional actionable mutations. & \xmark & \cmark & \cmark & \cmark & \cmark \\
  & Transition the therapeutic strategy to a second-line regimen, specifically FOLFIRI combined with Bevacizumab, to address disease progression. & \cmark & \cmark & \cmark & \cmark & \cmark \\
  & Given the presence of a solitary liver metastasis, a thorough assessment for surgical resection or localized radiotherapy options is strongly advised to achieve optimal disease control. & \cmark & \cmark & \cmark & \cmark & \cmark \\
\midrule
D & Confirm the presence of metastases in both the lungs and liver. & \cmark & \cmark & \cmark & \cmark & \cmark \\
  & Given the pulmonary and hepatic metastatic burden, a systemic therapeutic approach is favored over surgical interventions.  & \cmark & \cmark & \cmark & \cmark & \cmark \\
  & Disucss the necessity of presenting the patient case in a tumor board with oncologists, radiologists and surgeons to define the therapeutic concept once and the potential of local resection. & \cmark & \cmark & \cmark & \cmark & \cmark \\
  & Genetic profiling indicates wild-type BRAF and mutant KRAS, with microsatellite instability (MSI). & \cmark & \cmark & \cmark & \xmark & \cmark \\
  & A discrepancy between the patient's report of a BRAF mutation and the predictive tool's identification of BRAF wild-type necessitates further clarification through additional documentation or testing. & \cmark & \cmark & \cmark & \cmark & \cmark \\
  & Should the presence of a BRAF V600E mutation be confirmed, a therapeutic strategy involving Encorafenib and Cetuximab should be remembered as an option in further lines of therapy. & \cmark & \cmark & \cmark & \cmark & \cmark \\
  & In the case of high microsatellite instability (MSIH), Pembrolizumab is recommended as the number one priority, aligning with the patient's preference to avoid chemotherapy. & \cmark & \cmark & \cmark & \cmark & \cmark \\
\midrule
G & Progression of disease is indicated by the observed increase in the size and number of liver metastases over time. & \cmark & \cmark & \cmark & \cmark & \cmark \\
  & The current treatment regimen of FOLFOX is deemed ineffective, necessitating discontinuation and transition to a second-line therapy, likely FOLFIRI in combination with a targeted therapeutic agent. & \cmark & \cmark & \cmark & \cmark & \cmark \\
  & Effective selection of targeted therapy requires molecular profiling, including assessments for microsatellite instability (MSI), and KRAS and BRAF mutations. & \cmark & \cmark & \cmark & \cmark & \cmark \\
  & For cases presenting with microsatellite instability, a second-line immunotherapy regimen excluding Pembrolizumab, specifically Ipilimumab in conjunction with Nivolumab, is recommended. & \xmark & \cmark & \cmark & \cmark & \cmark \\
  & In scenarios where microsatellite stability (MSS) is observed alongside wild-type KRAS and BRAF, integration of an anti-EGFR agent with FOLFIRI is advised. & \cmark & \cmark & \cmark & \cmark & \cmark \\
  & For patients with MSS and mutations in KRAS or BRAF, the addition of an anti-VEGFR agent (such as Bevacizumab) to FOLFIRI is suggested. & \cmark & \cmark & \cmark & \cmark & \cmark \\
  & The consideration of local treatment modalities should be noted in case of a positive response to systemic therapy in the future. & \cmark & \cmark & \cmark & \cmark & \cmark \\
\midrule
Le & Document progression in lung metastases over time as evidenced by an increased number of lung nodules. & \cmark & \cmark & \cmark & \cmark & \cmark \\
  & For patients in good clinical condition, identify chemotherapy as the first line therapy approach. & \cmark & \cmark & \cmark & \cmark & \cmark \\
  & Chemotherapy options are either FOLFIRINOX or the combination of nab-Paclitaxel with Gemcitabine, following current clinical guidelines. & \xmark & \cmark & \cmark & \cmark & \cmark \\
  & Incorporate targeted therapy options to be considered, with Sotorasib identified as a viable option for patients with KRAS G12C mutations. This recommendation is based on evidence from clinical trials demonstrating the efficacy and safety profile of Sotorasib in KRAS G12C-mutated pancreatic cancer after a previous line of therapy (https://pubmed.ncbi.nlm.nih.gov/36546651/). & \cmark & \cmark & \cmark & \cmark & \cmark \\
  & In cases with the KRAS G12C mutation, explore enrollment in clinical trials for future treatments. & \cmark & \cmark & \cmark & \cmark & \cmark \\
\midrule
Lo & CT imaging of the abdomen confirms the presence of widespread liver metastases. & \cmark & \cmark & \cmark & \cmark & \cmark \\
  & Microscopic examination indicates microsatellite stability (MSS) with no mutations detected in either BRAF or KRAS genes (wild-type status for both). & \cmark & \cmark & \cmark & \cmark & \cmark \\
  & For a young, physically robust patient with an ECOG performance status of 1 and left-sided colorectal cancer (CRC) featuring wild-type KRAS and BRAF, the recommended first-line chemotherapy includes either the FOLFOX or FOLFIRI regimen, supplemented with an anti-EGFR antibody. & \cmark & \cmark & \cmark & \cmark & \cmark \\
  & Given the advanced stage, evaluate the feasibility of surgical resection or local therapies such as Radiofrequency Ablation (RFA) or Selective Internal Radiation Therapy (SIRT), especially if systemic therapy leads to a significant reduction in tumor burden. Acknowledge that this usually  goes beyond the capabilities of a single expert / the vision model and suggest forwarding the issue to a multidisciplinary tumor board. & \cmark & \cmark & \cmark & \cmark & \cmark \\
  & Optionally consider FOLFOXIRI plus bevacizumab if patient is in good conditions & \xmark & \xmark & \xmark & \xmark & \xmark \\
\midrule
S & Detect progression of disease within the liver through as evidence in the CT Abdomen scans over time. & \cmark & \cmark & \cmark & \cmark & \cmark \\
  & Molecular profiling indicates a mutation in KRAS, microsatellite stability (MSS), and wild-type status for BRAF. & \cmark & \cmark & \xmark & \cmark & \cmark \\
  & Highlight the discrepancy in KRAS status, transitioning from wild-type to mutated, which necessitates therapeutic reevaluation. & \cmark & \cmark & \cmark & \cmark & \cmark \\
  & Due to disease progression under the regimen of 5-FU plus Panitumumab and the emergence of KRAS mutation, recommend discontinuing Panitumumab. & \cmark & \cmark & \cmark & \cmark & \cmark \\
  & Propose transitioning to a regimen of 5-FU combined with Bevacizumab, or alternatively, FOLFIRI with Bevacizumab, as a response to the current treatment's ineffectiveness. & \cmark & \cmark & \cmark & \xmark & \cmark \\
  & Also consider Aflibercept or Ramucirumab in combination with FOLFIRI as viable alternatives to Bevacizumab with FOLFIRI, particularly due to the patient's progression on an oxaliplatin-based chemotherapy regimen. & \xmark & \xmark & \xmark & \cmark & \xmark \\
  & Advocate for comprehensive panel testing to identify additional mutations and subtypes (including the kras mutation subtype), which may inform further personalized treatment strategies. & \cmark & \cmark & \cmark & \cmark & \cmark \\
\midrule
T & Confirm the presence of a hepatic tumor via the provided CT image. & \cmark & \cmark & \cmark & \cmark & \cmark \\
  & Acknowledge the recurrence of the disease despite multiple Transarterial Chemoembolization (TACE) treatments, coupled with the existing condition of portal vein thrombosis (PVT), necessitating a shift to systemic therapy. & \cmark & \cmark & \cmark & \cmark & \cmark \\
  & Assess therapeutic strategies considering the severity and extent of the portal vein thrombosis. & \cmark & \cmark & \cmark & \cmark & \cmark \\
  & In line with findings from the IMBRAVE 150 trial, recommend Atezolizumab in combination with Bevacizumab as the preferred first-line systemic treatment, offering a superior option to Sorafenib for this clinical scenario. & \cmark & \cmark & \cmark & \cmark & \cmark \\
  & Address associated complications such as varices and portal hypertension by exploring suitable treatment modalities, including but not limited to endoscopic interventions and beta-blockers & \cmark & \cmark & \cmark & \cmark & \cmark \\
\midrule
W & Recognize disease progression characterized by the growth of the metastatic lesion. & \cmark & \cmark & \cmark & \cmark & \cmark \\
  & Molecular analysis reveals a BRAF mutation, wild-type KRAS, and microsatellite instability (MSI). & \cmark & \cmark & \xmark & \cmark & \cmark \\
  & Reassess the necessity of chemotherapy given the presence of a solitary liver metastasis, contemplating less aggressive alternatives. & \cmark & \cmark & \cmark & \cmark & \cmark \\
  & Advocate for the consideration of surgical resection as a viable option for localized disease control. & \xmark & \cmark & \cmark & \cmark & \cmark \\
  & Propose the use of immunotherapy, specifically the combination of Ipilimumab and Nivolumab, acknowledging that the patient is beyond first-line treatment scenarios (whereas in the first line Pembrolizumab could have been an option). & \xmark & \cmark & \xmark & \cmark & \xmark \\
  & Maintain consideration for future treatment strategies targeting BRAF V600E mutations, allowing for adaptability in the treatment plan based on evolving clinical and molecular insights. & \cmark & \cmark & \cmark & \cmark & \cmark \\
\midrule
X & Define progressive disease by increased tumor size, peritoneal carcinomatosis, and the development of ascites on imaging. & \cmark & \xmark & \cmark & \cmark & \cmark \\
  & Detect a BRAF mutation, wild-type KRAS, and microsatellite stability (MSS). & \cmark & \cmark & \xmark & \cmark & \cmark \\
  & Address the discrepancy between reports of microsatellite instability (MSI) from the text and MSS as per ground truth that shall be the prediction of the tool result. & \cmark & \cmark & \cmark & \cmark & \cmark \\
  & For MSI-high cases, consider Pembrolizumab as an immunotherapy option, in line with current guidelines. & \cmark & \cmark & \cmark & \xmark & \cmark \\
  & For BRAF V600E mutations, identify at least Dabrafenib and Trametinib as targeted therapy options, according to current guidelines. & \cmark & \cmark & \cmark & \cmark & \cmark \\
  & Explore options for targeted therapies for ROS CD74 mutations based on available online resources. & \cmark & \cmark & \cmark & \cmark & \cmark \\
  & Raise the question regarding the absence of Durvalumab in the initial treatment regimen which is part of the ESMO standard first line treatment (Cisplatin+Gemcitabine+Durvalumab). & \xmark & \xmark & \xmark & \cmark & \xmark \\
  & Mention FOLFOX as a second-line treatment chemotherapy option (as this eas explicitely asked) in cholangiocarcinoma following Cisplatin+Gemcitabine(+Durvalumab), with alternatives including 5-FU+/-Irinotecan or other targeted therapies. & \cmark & \cmark & \xmark & \cmark & \cmark \\
\midrule
Z & Confirm the presence of liver metastasis as evidenced in the CT image. & \cmark & \cmark & \cmark & \cmark & \cmark \\
  & Conclude the necessity of systemic therapy. & \cmark & \cmark & \cmark & \cmark & \cmark \\
  & For first-line chemotherapy, recommend FOLFIRINOX for patients in favorable clinical condition. For those in less optimal health, consider Gemcitabine plus Nab-Paclitaxel (GN) or Gemcitabine alone as alternatives. & \cmark & \cmark & \cmark & \cmark & \cmark \\
  & Prioritize testing for KRAS mutations; in cases of KRAS wild-type status, comprehensive panel diagnostics are advised to identify additional molecular targets for therapy. & \cmark & \cmark & \xmark & \cmark & \cmark \\
\end{longtable}

\setcounter{footnote}{0}
  
\textbf{Supplementary Table 6. Model Completeness.}
The table outlines the assessments by four human raters regarding the
presence or completion of specified keywords, with \emph{M} representing
the consensus achieved through a majority vote.
\clearpage

\begin{longtable}{l p{10cm} l l l l r} 
\caption*{{\Large Supplementary Table 7: Model Helpfulness - Human Evaluation}}\\

\toprule
\textbf{No.} & \textbf{Subquestions} & \textbf{1} & \textbf{2} & \textbf{3} & \textbf{4} & \textbf{$\mathbb{M}$} \\
\midrule
\endfirsthead
\toprule
\textbf{No.} & \textbf{Subquestions} & \textbf{1} & \textbf{2} & \textbf{3} & \textbf{4} & \textbf{$\mathbb{M}$} \\\midrule
\endhead
\midrule
\multicolumn{7}{r}{Continued on next page} \\
\midrule
\endfoot
\bottomrule
\endlastfoot
A & How did the disease develop? & \cmark & \cmark & \cmark & \cmark & \cmark \\
  & What are the next steps in her cancer treatment plan? & \cmark & \cmark & \cmark & \cmark & \cmark \\
\midrule
B & How do we best proceed with the oncologic treatment of the patient?  & \cmark & \cmark & \cmark & \cmark & \cmark \\
  & Is a continuation of chemotherapy plus Panitumumab recommended? & \cmark & \cmark & \cmark & \cmark & \cmark \\
  & What other options do we have? & \cmark & \cmark & \cmark & \cmark & \cmark \\
\midrule
D & Is the patient rather suitable for surgery or systemic therapy? & \cmark & \cmark & \cmark & \cmark & \cmark \\
  & Are there options for targeted therapies?  & \cmark & \cmark & \cmark & \cmark & \cmark \\
  & Check available drugs for the known mutation in databases. & \xmark & \cmark & \cmark & \xmark & \xmark \\
  & Check available drugs for the known mutation online. & \xmark & \xmark & \xmark & \xmark & \xmark \\
\midrule
G & How do we best proceed with the treatment of this patient? & \cmark & \cmark & \cmark & \cmark & \cmark \\
\midrule
Le & Please provide an overview of the standard chemotherapy protocols recommended for situations similar to Mr. Lee's, in alignment with current medical guidelines. 
 & \cmark & \cmark & \cmark & \cmark & \cmark \\
& Additionally, investigate and summarize the most recent online information regarding treatment options for KRAS G12C mutations, specifically for Mr. Lee's case. & \cmark & \cmark & \cmark & \cmark & \cmark \\
\midrule
Lo & What is the optimal management approach for this patient?  & \cmark & \cmark & \cmark & \cmark & \cmark \\
& Develop a detailed treatment plan. & \cmark & \cmark & \cmark & \cmark & \cmark \\
&  Should we consider local treatment options such as surgical intervention ...  & \cmark & \cmark & \cmark & \cmark & \cmark \\
& ... or liver-directed therapies? & \cmark & \cmark & \cmark & \cmark & \cmark \\
\midrule
S & What is the current state of the patient's health, particularly with respect to the liver metastases previously identified? & \cmark & \cmark & \cmark & \cmark & \cmark \\
  & Are these metastases showing signs of stability? & \cmark & \cmark & \cmark & \cmark & \cmark \\
  & Based on the current condition, is it advisable to persist with the regimen of 5-Fluorouracil and Panitumumab? & \cmark & \cmark & \cmark & \cmark & \cmark \\
\midrule
T & Can you suggest the next steps in the management of his disease? & \cmark & \cmark & \cmark & \cmark & \cmark \\
\midrule
W & investigate the development of the pre-existing metastasis  & \cmark & \cmark & \cmark & \cmark & \cmark \\
  & Did it grow or shrink under chemotherapy?  & \cmark & \cmark & \cmark & \cmark & \cmark \\
  & Can you measure how much it changed?  & \cmark & \cmark & \cmark & \cmark & \cmark \\
  & Is it Progressive Disease or did the tumor respond to the therapy?  & \cmark & \cmark & \cmark & \cmark & \cmark \\
  & How do we proceed with the treatment?  & \cmark & \cmark & \cmark & \cmark & \cmark \\
  & If its progressive disease, what is the standard second line therapy according to guidelines?  & \cmark & \cmark & \cmark & \cmark & \cmark \\
  & Shall we switch to FOLFIRI?  & \cmark & \cmark & \cmark & \cmark & \cmark \\
  & Show me all treatments the guidelines recommend in this situation?  & \cmark & \cmark & \cmark & \cmark & \cmark \\
  & Also check her biopsy results for additional targets.  & \xmark & \cmark & \cmark & \xmark &  \xmark \\
  & Find information on targeting BRAF V600E, too.  & \cmark & \cmark & \cmark & \cmark & \cmark \\
  & Are there local treatment options? & \cmark & \cmark & \cmark & \cmark & \cmark \\
\midrule
X & Please investigate the current state of the disease in detail. What does the new CT scan show?  & \cmark & \cmark & \cmark & \cmark & \cmark \\
  & Did the metastasis grow? & \cmark & \cmark & \cmark & \cmark & \cmark \\
  & In case of progress, what targeted or chemotherapy options could we suggest for the tumor board tomorrow?  & \cmark & \cmark & \cmark & \cmark & \cmark \\
  & What treatment options does the patient still have according to the official guidelines for cholangiocellular cancer given her medical history.  & \cmark & \cmark & \cmark & \cmark & \cmark \\
  & Also check any options you can find on google or pubmed. & \cmark & \xmark & \xmark & \xmark & \xmark \\
\midrule
Z & Considering Mr. Zu's diagnosis and the current stage of his pancreatic adenocarcinoma, what specific treatment recommendations should we prepare for discussion at tomorrow's tumor board meeting? & \cmark & \cmark & \cmark & \cmark & \cmark \\
\end{longtable}
\setcounter{footnote}{0}
  
\textbf{Supplementary Table 7. Model Helpfulness.}
This table lists subquestions and instructions for each patient case
alongside their corresponding evaluations, where a checkmark indicates
success and a cross signifies a failure to respond as defined by the
four human evaluators. \emph{M} represents the majority vote outcome.
\clearpage

\end{document}